\documentclass[lettersize,journal]{IEEEtran}
\usepackage{amsmath,amsfonts}
\usepackage{algorithmic}
\usepackage{algorithm}
\usepackage{array}
\usepackage[caption,font=normalsize,labelfont=sf,textfont=sf]{subfig}
\usepackage{textcomp}
\usepackage{stfloats}
\usepackage{url}
\usepackage{verbatim}
\usepackage{graphicx}
\usepackage{epstopdf}
\graphicspath{{figures/}}
\usepackage{cite}
\usepackage{amssymb,amsthm}
\usepackage{multirow}
\usepackage{hyperref}
\usepackage{booktabs,multirow}
\usepackage{booktabs,multirow}
\usepackage{booktabs,multirow}
\usepackage{float}
\usepackage{array}
\usepackage{lipsum}
\usepackage{xcolor}
\usepackage[utf8]{inputenc}
\makeatletter
\newtheorem{lemma}{Lemma}
\newtheorem{theorem}{Theorem}
\newtheorem{definition}{Definition}

\newcommand{\Rmnum}[1]{\expandafter\@slowromancap\romannumeral #1@}
\makeatother

\begin{document}
	
\title{Learning from $M$-Tuple Dominant Positive and Unlabeled Data}
\author{Jiahe Qin, Junpeng Li, \emph{Member}, \emph{IEEE}, Changchun Hua, \emph{Fellow}, \emph{IEEE} and Yana Yang, \emph{Member}, \emph{IEEE} \thanks{J. Qin, J. Li, C. Hua, and Y. Yang are with the Engineering Research Center of the Ministry of Education for Intelligent Control System and Intelligent Equipment, Yanshan University, Qinhuangdao, China (qinjiahe@stumail.ysu.edu.cn; jpl@ysu.edu.cn; cch@ysu.edu.cn; yyn@ysu.edu.cn).}}
\date{}
\maketitle
	
\begin{abstract}
\emph{Label Proportion Learning (LLP)} addresses the classification problem where multiple instances are grouped into bags and each bag contains information about the proportion of each class. However, in practical applications, obtaining precise supervisory information regarding the proportion of instances in a specific class is challenging. To better align with real-world application scenarios and effectively leverage the proportional constraints of instances within tuples, this paper proposes a generalized learning framework \emph{MDPU}. Specifically, we first mathematically model the distribution of instances within tuples of arbitrary size, under the constraint that the number of positive instances is no less than that of negative instances. Then we derive an unbiased risk estimator that satisfies risk consistency based on the empirical risk minimization (ERM) method. To mitigate the inevitable overfitting issue during training, a risk correction method is introduced, leading to the development of a corrected risk estimator. The generalization error bounds of the unbiased risk estimator theoretically demonstrate the consistency of the proposed method. Extensive experiments on multiple datasets and comparisons with other relevant baseline methods comprehensively validate the effectiveness of the proposed learning framework.
\end{abstract}
	

\section{Introduction}
In many high-precision data analysis scenarios, the contradiction between the demand for large-scale, high-quality annotated data and the difficulty of obtaining such annotations has become increasingly prominent. Traditional fully supervised learning relies on precise per-sample annotations, but it faces significant limitations in terms of privacy protection, technical constraints, and manpower expenses. Although deep neural networks can achieve remarkable performance with sufficient labeled data, the high cost and error-proneness of manual annotation severely restrict their practical application.

\begin{figure}[!ht]
	\centering
	\includegraphics[width=0.45\textwidth]{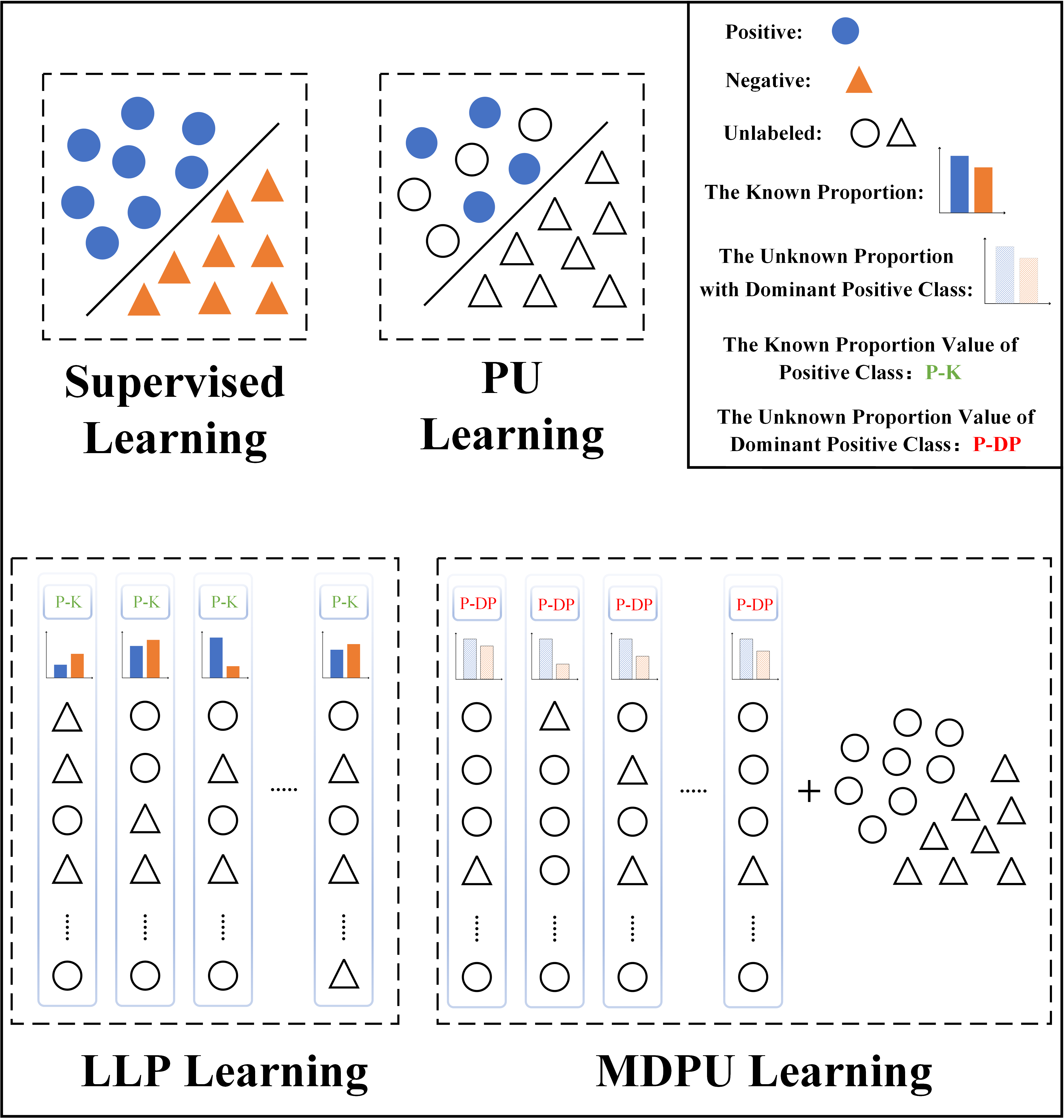}
	\caption{Illustrations of \emph{MDPU learning} and other related learning settings. \emph{MDPU} operates under dominant positive class (P-DP) assumption (positive count $\ge$ negative count in tuples, without exact proportion knowledge)}
	\label{fig:1}
\end{figure}

To address the challenge of scarce labeled data, \emph{Weakly Supervised Learning (WSL)} introduces diverse forms of labels, providing more flexible supervision information for model training. Reference \cite{zhou_brief_2018} systematically summarized various learning paradigms of \emph{WSL}, revealing its theoretical feasibility and broad prospects. Currently, \emph{Weakly Supervised Learning} \cite{Oquab_2015_CVPR, 7243351, zhou_brief_2018, 8735810, 9294086, 10025822} has evolved into diverse paradigms to adapt to different label-constrained scenarios: \emph{Positive-Unlabeled (PU) learning} \cite{elkan_learning_2008, 4796195, zhou2009learning, du2015convex, kiryo2017positive, gong2019large, sansone2019efficient, bekker2020learning, jiang2023positive} trains binary classifier using a small number of positive samples and a large number of pointwise unlabeled samples; \emph{Positive-Confidence (Pconf) learning} \cite{ishida2018binary} trains binary classifier based on the confidence information of positive class in unlabeled samples; \emph{Noisy Label Learning (NLL)} \cite{chen2014ambiguously, patrini2017making, han2018co, 9361098, cheng2020weakly, jiang2021learning, wu2022learning, song2022learning, 10934976, 10689264} designs robust optimization strategies for samples with noisy labels; \emph{Unlabeled-Unlabeled (UU) learning} \cite{lu2018minimal, lu2020mitigating, lu2021binary, wei2024consistent, tang2023multi} extracts potential classification information from unlabeled datasets with known class priors; \emph{Similarity-Unlabeled (SU) learning} \cite{bao2018classification} and \emph{Similarity-Dissimilarity-Unlabeled (SDU) learning} \cite{shimada2021classification} establish binary classifier through similarity constraints between pairwise samples; \emph{Similarity-Confidence (Sconf) learning} \cite{cao2021learning} quantifies the similarity between pairwise samples as confidence information for classification; \emph{Pairwise Confidence Comparison (Pcomp) learning} \cite{feng2021pointwise} trains binary classifier using the class preference relationships between pairwise samples; on the basis of \emph{Pcomp learning}, \emph{Confidence Difference (CD) learning} \cite{wang2023binary} further specifies these class preference relationships as differences in confidence scores, constructing binary classifier based on the confidence score differences. These methods provide diverse solutions for addressing the high cost and low quality of label acquisition in practical applications across different weakly supervised scenarios.

Moreover, the label forms of weakly supervised samples are not limited to incomplete per-sample annotations but can be extended to group-level weak supervision information, such as using aggregated labels of data bags to reflect the distribution characteristics of internal samples. \emph{Learning from Label Proportions (LLP)} \cite{quadrianto2009estimating, yu2014learning, yu2013proptosvm, scott2020learning} is an important branch of weakly supervised learning which has been extensively studied. \emph{LLP} utilize class proportion information within data bags instead of relying on per-sample labels, enabling the training of classification models. \emph{LLP} has demonstrated strong application potential in fields such as social media data analysis and text classification, but its reliance on precise proportion information remains a key challenge for broader practical adoption.

In medical imaging diagnosis, high-resolution scans enable clinicians to estimate positive region distributions, yet limited CT resolution may hinder precise localization of small/localized lesions in individual slices. Consequently, clinicians often assign group-level labels such as ``most slices contain tumors.'' This class-dominance constraint enhances model reliability by preventing deviations from true pathological distributions caused by noisy labels. In contrast, \emph{LLP} requires grouping all training instances into bags with exact positive proportions per bag---a condition impractical for medical scenarios where precise lesion proportion estimation is challenging. Similarly, satellite image analysis employs coarse labels (e.g., ``over 50\% flooded area'') to assess flood extents without \emph{LLP}'s precise proportion requirements.

Additionally, real-world repositories (such as satellite image archives and hospital PACS systems) contain a large amount of pointwise unlabeled data, which encompasses mixed distributions of positive and negative samples. Leveraging pointwise unlabeled data to participate in the training of weakly supervised learning models can enhance the generalization ability of classifiers. 

These practical application scenarios naturally give rise to a form of weakly supervised data tuples, where each tuple contains multiple unlabeled samples with a key prior information: the number of positive samples (e.g., vegetation patches or tumor slices) is not fewer than that of negative samples. However, existing weakly supervised methods fall short in effectively utilizing these statistical constraints along with the abundant unlabeled data.

To address this challenge, we propose a novel weakly supervised learning framework termed \emph{Learning from M-Tuple Dominant Positive and Unlabeled Data (MDPU)}. In this framework, unlabeled data tuples are constrained such that the number of positive samples is not fewer than that of negative samples, and each tuple can contain an arbitrary number of unlabeled samples. Starting with low-dimensional data tuples, we first elaborate on the data generation processes for pairwise and triple \emph{DPU} tuples, then generalize these to derive the mathematical distribution form for tuples with an arbitrary number $ M $ of samples. In addition to leveraging the statistical constraint information on the number of positive samples within tuples, the \emph{MDPU} learning framework also utilizes a large amount of pointwise unlabeled data to enhance classifier performance.

The main contributions of this paper can be concisely summarized as follows:  
\begin{itemize}
\item We propose a generalized learning framework termed \emph{``Learning from M-tuple Dominant Positive and Unlabeled Data (MDPU)''}, which demonstrates strong application value across a wide range of domains.
\item Based on the distribution of \emph{MDPU} data tuples, we derive an unbiased risk estimator that is independent of specific loss functions and optimizers. We derive generalization error bounds for this unbiased risk estimator that achieve the optimal convergence rate of parameters.
\item We introduce a corrected risk function to mitigate overfitting risks inherent in the unbiased risk estimator, thereby significantly improving classifier performance.
\item Extensive experiments on numerous datasets using Pairwise-DPU and Triple-DPU learning validate the rationality and effectiveness of the proposed \emph{MDPU} learning framework.
\end{itemize}

The paper is structured as follows:  
\textbf{Section \uppercase\expandafter{\romannumeral2}} reviews traditional binary classification and \emph{LLP} problem setup. \textbf{Section \uppercase\expandafter{\romannumeral3}} details the data generation for pairwise ($M=2$) and triple ($M=3$) tuples in \emph{MDPU} learning, and generalizes it to arbitrary $M$-tuple data. Additionally, an unbiased risk estimator is derived through \emph{MOVA} data. \textbf{Section \uppercase\expandafter{\romannumeral4}} addresses overfitting in the unbiased risk estimator through risk correction. \textbf{Section \uppercase\expandafter{\romannumeral5}} provides theoretical analysis and proves the consistency of \emph{MDPU}. \textbf{Section \uppercase\expandafter{\romannumeral6}} validates \emph{MDPU} via experiments on several datasets with statistical analyses.  
The proofs of related lemmas and theorems are provided in the appendix.

\section{Preliminaries}\label{Section:2}
%\lipsum[1]%
This section formalizes the ordinary binary classification setup and provide the conceptual basis for Label Proportion Learning.

\subsection{Ordinary binary classification}
Let $\mathcal{X} \subset \mathbb{R}^d$ represent the feature space, and $Y=\left\{ -1,+1 \right\}$ denote the label space comprising two classes. Each instance $\left( x,y \right)$ is sampled from the joint probability distribution characterized by the density $p\left( x,y \right)$. The primary objective is to derive a classifier $g\left( \cdot \right) : \mathcal{X} \rightarrow \mathbb{R} $ capable of minimizing the classification risk:
\begin{equation}
	\label{eq1}
	R\left( g \right) ={\mathbb{E}}_{p\left( x,y \right)}\left[ \ell \left( g\left( x \right) ,y \right) \right] 
\end{equation}
where ${\mathbb{E}}_{p\left( x,y \right)}$ denotes the expectation over $p\left( x,y \right) $, and $
\ell \left( \cdot ,\cdot \right) \,\,: \mathbb{R} \times Y\rightarrow \mathbb{R} ^+
$ is a loss function that assesses the accuracy of the classifier in estimating the true class label. Let $\pi _+=p\left( y=+1 \right)$ ($\pi _-=p\left( y=-1 \right)$) denote the class-prior probability of the positive data (negative data). Then the equivalent expression of classification risk (\ref{eq1}) can be expressed as:
\begin{equation}
	R\left( g \right) \,\,=\,\,\pi _+\mathbb{E} _+\left[ \ell \left( g\left( x \right) ,+1 \right) \right] +\pi _-\mathbb{E} _-\left[ \ell \left( g\left( x \right) ,-1 \right) \right]
\end{equation}
where $\,\,{\mathbb{E}}_+\left[ \cdot \right] $ and $ \,\,{\mathbb{E}}_-\left[ \cdot \right] $ are expectations over class-conditional probability density with $ p_+\left( x \right) =p\left( x|y=+1 \right) $ and $ p_-\left( x \right) =p\left( x|y=-1 \right) $, respectively.

\subsection{Learning from Label Proportions}
\emph{Label Proportions Learning (LLP)} requires obtaining the label proportion information for all classes within each group (or bag) and training a model to predict the true labels of individual samples in the bag using this proportion information. Assuming that the samples in each bag are independently and identically distributed, each bag can be represented as $ \tilde{x} = (x_1, x_2, \cdots, x_k) $, with the corresponding true labels denoted as $ \tilde{y} = (y_1, y_2, \cdots, y_k) $. Based on a large set of training samples $ \{\tilde{x}, f(\tilde{y})\} $, where $ f\left( \tilde{y} \right) =1c/k*\sum_{i=1}^k{\left( \left( y_i+1 \right) /2 \right)}$ represents the label proportion generation function, the empirical proportion risk minimization (EPRM) method is employed to minimize the empirical proportion loss. This results in a classifier $ h $ ($ h \in \mathcal{H} $) that achieves a low prediction error. The empirical proportion loss is expressed as:
\begin{equation}
L\left( h \right) =L\left( \phi _{r}^{f}(h)(\tilde{x}),f(\tilde{y}) \right) 
\end{equation}
where $ \phi _{r}^{f}(h)(\tilde{x})$ represents the predicted label proportion.

\emph{LLP} imposes stringent requirements on label proportion information, which is often not only inaccurate but also entirely inaccessible in scenarios involving personal privacy, data confidentiality or sensitivity. This limitation severely undermines the practical utility of \emph{LLP}. To address this challenge, we propose a novel weakly supervised learning framework that relaxes the dependency on precise label proportions---requiring only the ordinal relationship between class proportions within a tuple---to achieve high-performance classifiers.

\section{Learning From $M$-tuple Dominant Positive and Unlabeled Data}\label{Section:3}
This section first details the generation processes for pairwise, triple and $M$-tuple dominant positive data. Then, an unbiased risk estimator is derived to train a binary classifier by empirical risk minimization (ERM).
\subsection{Data Generation Process of Dominant Positive Data}

\subsubsection{Pairwise Dominant Positive Data}

Arbitrarily sized data tuples are independently sampled, with weakly supervised information (indicative of dominant positive proportions) acquired through crowdsourcing or alternative methods. Specifically, when the tuple contains only two samples, three possible label configurations exist:
\begin{align*}  
\left\{ \left( +1,+1 \right) ,\left( +1,-1 \right) ,\left( -1,+1 \right) \right\} 
\end{align*}  

Let $D_{Pairwise} = \left\{ \left( x_i^1, x_i^2 \right) \right\}_{i=1}^{n_{Pairwise}}$ denote the dataset of binary tuples satisfying the weak supervision condition of positive dominance (positive samples $\geq$ negative samples). This dataset is independently sampled from the distribution characterized in Lemma \ref{lemma-1}.

\begin{lemma}\label{lemma-1}
Data tuples in $D_{Pairwise}$ are independently drawn from:
\begin{equation}
\begin{aligned}
\tilde{p}\left( x^1,x^2 \right) =&\frac{1}{\pi _{+}^{2}+2\pi _+\pi _-}\left[ \pi _{+}^{2}p_+\left( x^1 \right) p_+\left( x^2 \right) \right. \\& \left.    +\pi _+\pi _-p_+\left( x^1 \right) p_-\left( x^2 \right)  \right. \\& \left.+\pi _+\pi _-p_-\left( x^1 \right) p_+\left( x^2 \right) \right] 
\end{aligned}
\end{equation}
\end{lemma}

The pointwise samples $\left\{ x_{i}^{1} \right\} _{i=1}^{n_{Pairwise}}$ in $\tilde{\mathcal{D}}_{P1}$ and $\left\{ x_{i}^{2} \right\} _{i=1}^{n_{Pairwise}}$ in $\tilde{\mathcal{D}}_{P2}$ are independently and identically generated from marginal distributions $\tilde{p}_{P1}\left( x^1 \right)$, $\tilde{p}_{P2}\left( x^2 \right)$, respectively. These distributions are composed of weighted contributions from both positive and negative classes, providing a clearer understanding of how the pointwise data sample is generated.

\begin{lemma} 
The marginal distributions $\tilde{p}_{P1}\left( x^1 \right)$, $\tilde{p}_{P2}\left( x^2 \right)$ can be expressed as:
\begin{align*}
\tilde{p}_{P1}\left( x^1 \right) =\frac{\pi _{+}^{2}+\pi _+\pi _-}{\pi _{+}^{2}+2\pi _+\pi _-}p_+\left( x^1 \right) +\frac{\pi _+\pi _-}{\pi _{+}^{2}+2\pi _+\pi _-}p_-\left( x^1 \right) 
\\
\tilde{p}_{P2}\left( x^2 \right) =\frac{\pi _{+}^{2}+\pi _+\pi _-}{\pi _{+}^{2}+2\pi _+\pi _-}p_+\left( x^2 \right) +\frac{\pi _+\pi _-}{\pi _{+}^{2}+2\pi _+\pi _-}p_-\left( x^2 \right) 
\end{align*}
\end{lemma}

\subsubsection{Triple Dominant Positive Data}

For triples satisfying the weak supervision condition of positive dominance (positive samples $\ge$ negative samples), four distinct label configurations are possible:
\begin{align*} 
\left\{ \left( +1,+1,+1 \right) ,\left( +1,+1,-1 \right) ,\left( -1,+1,+1 \right) ,\left( +1,-1,+1 \right) \right\}
\end{align*}

\begin{lemma}\label{lemma1}
Data tuples in $D_{Triple}$ are independently drawn from:
\begin{equation}
\begin{aligned}
\tilde{p}\left( x^1,x^2,x^3 \right) =&\frac{1}{\pi _{+}^{3}+3\pi _{+}^{2}\pi _-}\left[ \pi _{+}^{3}p_+\left( x^1 \right) p_+\left( x^2 \right) p_+\left( x^3 \right) \right. \\& \left.    +\pi _{+}^{2}\pi _-p_+\left( x^1 \right) p_+\left( x^2 \right) p_-\left( x^3 \right) \right. \\& \left.      +\pi _{+}^{2}\pi _-p_+\left( x^1 \right) p_-\left( x^2 \right) p_+\left( x^3 \right) \right. \\& \left.    +\pi _{+}^{2}\pi _-p_-\left( x^1 \right) p_+\left( x^2 \right) p_+\left( x^3 \right) \right] 
\end{aligned}
\end{equation}
\end{lemma}

The pointwise samples $\left\{ x_{i}^{1} \right\} _{i=1}^{n_{Triple}}$ in $\tilde{\mathcal{D}}_{T1}$, $\left\{ x_{i}^{2} \right\} _{i=1}^{n_{Triple}}$ in $\tilde{\mathcal{D}}_{T2}$ and $\left\{ x_{i}^{3} \right\} _{i=1}^{n_{Triple}}$ in $\tilde{\mathcal{D}}_{T3}$ are independently and identically generated from marginal distributions $\tilde{p}_{T1}\left( x^1 \right)$, $\tilde{p}_{T2}\left( x^2 \right)$, $\tilde{p}_{T3}\left( x^3 \right)$, respectively. Similarly, these distributions can be expressed as a combination of positive and negative class components.
\begin{lemma} 
The marginal distributions $\tilde{p}_{T1}\left( x^1 \right)$, $\tilde{p}_{T2}\left( x^2 \right)$, $\tilde{p}_{T3}\left( x^3 \right)$ can be expressed as:
\begin{align*}
\tilde{p}_{T1}\left( x^1 \right) =\frac{\pi _{+}^{3}+2\pi _{+}^{2}\pi _-}{\pi _{+}^{3}+3\pi _{+}^{2}\pi _-}p_+\left( x^1 \right) +\frac{\pi _{+}^{2}\pi _-}{\pi _{+}^{3}+3\pi _{+}^{2}\pi _-}p_-\left( x^1 \right) 
\\
\tilde{p}_{T2}\left( x^2 \right) =\frac{\pi _{+}^{3}+2\pi _{+}^{2}\pi _-}{\pi _{+}^{3}+3\pi _{+}^{2}\pi _-}p_+\left( x^2 \right) +\frac{\pi _{+}^{2}\pi _-}{\pi _{+}^{3}+3\pi _{+}^{2}\pi _-}p_-\left( x^2 \right) 
\\
\tilde{p}_{T3}\left( x^3 \right) =\frac{\pi _{+}^{3}+2\pi _{+}^{2}\pi _-}{\pi _{+}^{3}+3\pi _{+}^{2}\pi _-}p_+\left( x^3 \right) +\frac{\pi _{+}^{2}\pi _-}{\pi _{+}^{3}+3\pi _{+}^{2}\pi _-}p_-\left( x^3 \right) 
\end{align*}
\end{lemma}
The condition that the proportion of positive samples is no less than that of negative samples applies to tuples of any size. Therefore, the size of such data tuples is not limited to two or three but can be extended to tuples of arbitrary size $M$.

As a result, this paper extends the idea of pairs and triplets to propose a generalized learning framework that builds classifiers by leveraging the dominance of positive samples in $M$-sized tuples. Furthermore, the process of generating data with a higher proportion of positive samples among $M$ tuples is described as follows:

\subsubsection{$M$-tuple Dominant Positive Data}
The size of the sample tuple is an arbitrary value $M$, and the proportion of unlabeled samples in the tuple belonging to the positive class is greater than or equal to that of the negative class. This paper denotes the dataset of data tuples that satisfy this constraint as $D_{M-tuple}=\left\{ \left( x_{i}^{1},x_{i}^{2},\cdots ,x_{i}^{M} \right) \right\} _{i=1}^{n_M}$. The distribution that $D_{M-tuple}$ follows is shown in Lemma \ref{lemma-5} below.

\begin{lemma}\label{lemma-5}
Data tuples in $D_{M-tuple}$ are independently drawn from:
\begin{equation}
\begin{footnotesize} 
\begin{aligned}  
&p_{MDP}\left( x^1,x^2,...,x^M \right) =\\&\frac{\sum_{k=0}^{\lfloor M/2 \rfloor}{\pi _{+}^{M-k}}\pi _{-}^{k}\sum_{S\subseteq \{1,2,...,M\},|S|=k}{\left( \prod_{i\in S}{p_-}(x^i)\prod_{i\notin S}{p_+}(x^i) \right)}}{\sum_{k=0}^{\lfloor M/2 \rfloor}{\binom{M}{k}}\pi _{+}^{M-k}\pi _{-}^{k}}
\end{aligned}
\end{footnotesize}
\end{equation}
\end{lemma}

The samples $\left\{ x_{i}^{1} \right\} _{i=1}^{n_M}$ in $\hat{\mathcal{D}}_1$, $\left\{ x_{i}^{2} \right\} _{i=1}^{n_M}$ in $\hat{\mathcal{D}}_2$, $\cdots$, and $\left\{ x_{i}^{M} \right\} _{i=1}^{n_M}$ in $\hat{\mathcal{D}}_M$ are generated independently and identically from the marginal distributions $\hat{p}_{1DP}\left( x^1 \right)$, $\hat{p}_{2DP}\left( x^2 \right)$, $\cdots$, $\hat{p}_{MDP}\left( x^M \right)$, respectively. A detailed mathematical characterization of these marginal distributions $\hat{p}_{1DP}\left( x^1 \right)$, $\hat{p}_{2DP}\left( x^2 \right)$, $\cdots$, $\hat{p}_{MDP}\left( x^M \right)$ is provided in Lemma \ref{lemma-6}.

\begin{lemma}\label{lemma-6}
The marginal distributions $\hat{p}_{1DP}\left( x^1 \right)$, $\hat{p}_{2DP}\left( x^2 \right)$, $\cdots$, $\hat{p}_{MDP}\left( x^M \right)$ can be expressed as:
\begin{align*} 
&\hat{p}_{1DP}\left( x^1 \right) =ap_+\left( x^1 \right) +bp_-\left( x^1 \right) 
\\&
\hat{p}_{2DP}\left( x^2 \right) =ap_+\left( x^2 \right) +bp_-\left( x^2 \right)
\end{align*} 
$$
\vdots 
$$
\begin{align*} 
\hat{p}_{MDP}\left( x^M \right) =ap_+\left( x^M \right) +bp_-\left( x^M \right) 
\end{align*} 
where,
\begin{align*} 
a=\frac{\sum_{k=0}^{\lfloor M/2 \rfloor}{\left( \begin{array}{c}
			M-1\\
			k\\
		\end{array} \right) \pi _{+}^{M-k}\pi _{-}^{k}}}{\sum_{k=0}^{\lfloor M/2 \rfloor}{\left( \begin{array}{c}
			M\\
			k\\
		\end{array} \right) \pi _{+}^{M-k}\pi _{-}^{k}}},
\\
b=\frac{\sum_{k=1}^{\lfloor M/2 \rfloor}{\left( \begin{array}{c}
			M-1\\
			k-1\\
		\end{array} \right) \pi _{+}^{M-k}\pi _{-}^{k}}}{\sum_{k=0}^{\lfloor M/2 \rfloor}{\left( \begin{array}{c}
			M\\
			k\\
		\end{array} \right) \pi _{+}^{M-k}\pi _{-}^{k}}}
\end{align*} 
\end{lemma}

{\bf Pointwise Unlabeled Data:} It is assumed that pointwise unlabeled data $\left\{ x_{i}^{U} \right\} _{i=1}^{n_U}$ is independently sampled from $p_U\left( x \right)$. The distribution of pointwise unlabeled data can be expressed as follows:
\begin{equation}
	\begin{aligned}
		p_U\left( x \right) =\pi _+p_+\left( x \right) +\pi _-p_-\left( x \right) 
	\end{aligned} 
\end{equation}

By extracting information about the positive and negative sample distributions from each pointwise sample distribution in the $M$-tuple, we can derive an unbiased risk estimator for binary classification using the distributions of positive and negative samples.

The specific expressions for the distribution of positive samples $p_+\left( x \right)$ and negative samples $p_-\left( x \right)$ are provided in Lemma \ref{lemma-7}.

\begin{lemma}\label{lemma-7}
The distributions of positive and negative samples, represented as $p_+(x)$ and $p_-(x)$ respectively, can be formulated in terms of $\hat{p}_{MDP}(x)$ and $p_U(x)$.

\begin{equation}
	\begin{aligned}
p_+\left( x \right) =\frac{\pi _-}{a\pi _--b\pi _+}\hat{p}_{MDP}\left( x \right) -\frac{b}{a\pi _--b\pi _+}p_U\left( x \right) 
	\end{aligned} 
\end{equation}
\begin{equation}
	\begin{aligned}
p_-\left( x \right) =\frac{-\pi _+}{a\pi _--b\pi _+}\hat{p}_{MDP}\left( x \right) +\frac{a}{a\pi _--b\pi _+}p_U\left( x \right) 
	\end{aligned} 
\end{equation}
\end{lemma}

\subsection{Unbiased Risk Estimator with MDPU Data}
\begin{theorem}\label{Th1}
The classification risk $R(g)$ (\ref{eq1}) is formulated based on MDPU data:

\begin{equation}\label{eq7}
\begin{aligned}
&R_{MDPU}\left( g \right) 
\\
=&\pi _+\pi _-\underset{\left( \boldsymbol{x}^1,\boldsymbol{x}^2,\cdots ,\boldsymbol{x}^{\boldsymbol{M}} \right) \sim p_{MDP}\left( \boldsymbol{x}^1,\boldsymbol{x}^2,\cdots ,\boldsymbol{x}^{\boldsymbol{M}} \right)}{\mathbb{E}}\left[ \tilde{\mathcal{L}}_{MDP}\left( g\left( x \right) \right) \right] 
\\
&+\underset{x\sim p_U\left( x \right)}{\mathbb{E}}\left[ \mathcal{L} _U\left( g\left( x \right) \right) \right] 
\end{aligned}
\end{equation}
where,
$$
\tilde{\mathcal{L}}_{MDP}\left( g\left( x \right) \right) \triangleq \frac{\mathcal{L} _{MDP}\left( g\left( x^1 \right) \right) +\cdots +\mathcal{L} _{MDP}\left( g\left( x^M \right) \right)}{M}
$$
$$
\mathcal{L} _{MDP}\left( g\left( x^i \right) \right) \triangleq \frac{\ell \left( g\left( x^i \right) ,+1 \right) -\ell \left( g\left( x^i \right) ,-1 \right)}{a\pi _--b\pi _+}
$$
$$
\mathcal{L} _U\left( g\left( x \right) \right) \triangleq\frac{\left( -b\pi _+ \right) \ell \left( g\left( x \right) ,+1 \right) +\left( a\pi _- \right) \ell \left( g\left( x \right) ,-1 \right)}{a\pi _--b\pi _+}
$$  

\end{theorem} 

The classification risk derived from \emph{MDPU} data in Theorem \ref{Th1} can be approximated by the empirical risk, which is computed using the sample means of \emph{MDPU} data samples:

\begin{equation}\label{eq-7}
\begin{aligned}
\hat{R}_{MDPU}\left( g \right) =&\frac{\pi _+\pi _-}{Mn_{MDP}}\sum_{i=1}^{Mn_{MDP}}{\mathcal{L} _{MDP}\left( g\left( x_{MDP,i} \right) \right)}\\&+\frac{1}{n_U}\sum_{j=1}^{n_U}{\left[ \mathcal{L} _U\left( g\left( x_{U,i} \right) \right) \right]}
\end{aligned}
\end{equation}

where,
$$
\mathcal{L} _{MDP}\left( g\left( x_{MDP,i} \right) \right) \triangleq \frac{1}{a\pi _--b\pi _+}\tilde{\ell}\left( g\left( x^i \right) \right) 
$$
$$ 
\tilde{\ell}\left( g\left( x^i \right) \right) \triangleq \ell \left( g\left( x^i \right) ,+1 \right) -\ell \left( g\left( x^i \right) ,-1 \right) 
$$

\subsection{Estimation Error Bound}
We establish an estimation error bound for \emph{MDPU learning} to analyze the convergence property of the proposed risk estimator $\hat{R}_{\text{MDPU}}(g)$. A pivotal definition that we introduce here is Rademacher complexity, a standard metric for quantifying the complexity of the sample space.

\begin{definition}\label{Def1} (Rademacher complexity)\cite{mohri2018foundations}
	Let $Z_1,\cdots ,Z_n$ be i.i.d. random variables drawn from a probability distribution with density $\mu$, $\mathcal{G} =\left\{ g: \mathcal{X} \rightarrow \mathbb{R} \right\} $ be a class of measurable functions. Then the Rademacher complexity of $\mathcal{G} $ is defined as:
	\begin{equation}
	\Re \left( \mathcal{G}\right) =\mathbb{E} _{Z_1,\cdots ,Z_n\thicksim \mu}\mathbb{E} _{\sigma}\left[ \underset{g\in \mathcal{G}}{sup}\frac{1}{n}\sum_{i=1}^n{\sigma _ig\left( Z_i \right)} \right] 
	\end{equation}
	where $n$ is a positive integer, and $\sigma =\left( \sigma _1,\cdots ,\sigma _n \right) $ are Rademacher variables that taking from $\left\{ -1,+1 \right\}$ uniformly.
\end{definition}

Assume that there exists a constant $C_g>0$ that $sup_{g\in \mathcal{G}}\left\| g \right\| _{\infty}\leq C_g$ and $C_{\ell}>0$ such that $sup_{|z|\leq C_g}\ell \left( z,y \right) \leq C_{\ell}$ holds for all $y$. It is also assumed that the loss function $\ell \left( z,y \right) $ is Lipschitz continuous for $z$	with $|z|\leq C_g$ and $y$ with a Lipschitz constant $L_{\ell}$. Let $g^*=arg\min _{g\in \mathcal{G}}R\left( g \right) $ be the minimizer of true risk in Eq.(\ref{eq1}), and $\hat{g}=arg\min _{g\in \mathcal{G}}\hat{R}\left( g \right) $ be the minimizer of empirical risk in Eq.(\ref{eq-7}). 

\begin{theorem}	\label{Th-2} For any $\delta >0$, the estimation error bound holds with probability at least $1-\delta $:
\begin{equation}
\begin{aligned}
R(\hat{g})-R(g^*)
\leq& 2\pi _+\pi _-\frac{4\sqrt{2}\rho C_{\mathcal{G}}+2C_{\ell}\sqrt{\log \frac{4}{\delta}}}{\left| a\pi _--b\pi _+ \right|\sqrt{2Mn_{MDP}}}
\\&
+\frac{2\left( \left| -b\pi _+ \right|+\left| a\pi _- \right| \right) \sqrt{8}\rho C_{\mathcal{G}}}{\left| a\pi _--b\pi _+ \right|\sqrt{2n_U}}
\\&
+\frac{2\left( \left| -b\pi _+ \right|+\left| a\pi _- \right| \right) C_{\ell}\sqrt{\log \frac{4}{\delta}}}{\left| a\pi _--b\pi _+ \right|\sqrt{2n_U}}
\end{aligned}
\end{equation}
\end{theorem}

The estimation error bound provided in Theorem \ref{Th-2} theoretically demonstrates the consistency of the proposed \emph{MDPU learning} approach, with $R(\hat{g}_{MDPU}) \to R(g^*)$ as $n_{MDP} \to \infty$ and $n_U \to \infty$. In addition, the estimation error bound for \emph{MDPU learning} converges at a rate of $\mathcal{O}_p(1/\sqrt{n_{MDP}} + 1/\sqrt{n_U})$, achieving the optimal parametric rate for empirical risk minimization under minimal assumptions \cite{mendelson_lower_2008}.

\begin{figure}[!ht]
	\centering
	\includegraphics[width=0.35\textwidth]{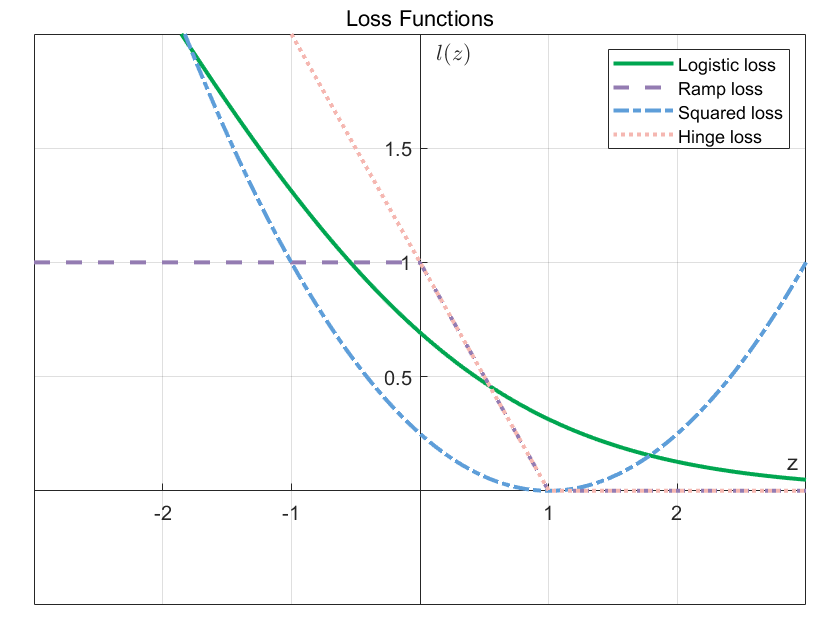}
	\caption{Illustrations of Different Loss Functions}
	\label{fig:2}
\end{figure}

\begin{table}[h]
	\centering
	\begin{tabular}{l l l}
		\toprule
		Loss Function & Notation & $\ell(t, z)$ \\
		\midrule
		Logistic loss & $\ell_{\text{Logistic}}(t,z)$ & $\log(1 + \exp(-tz))$ \\
		Ramp loss & $\ell_{\text{Ramp}}(t, z)$ & $\min(1, \max(0, 1 - tz))$ \\
		Squared loss & $\ell_{\text{Squared}}(t, z)$ & $\frac{1}{4}(tz - 1)^2$ \\
		Hinge loss & $\ell_{\text{Hinge}}(t, z)$ & $\max(0, 1 - tz)$ \\
		\bottomrule
	\end{tabular}
	\caption{Mathematical expressions for four distinct loss functions}
	\label{table:1}
\end{table}

\section{Risk Correction Method}\label{Section:4}
Based on the empirical risk function derived from \emph{MDPU} data (Eq.(\ref{eq7})), which includes two custom loss functions:
$$
\mathcal{L} _{MDP}\left( g\left( x^i \right) \right) \triangleq \frac{\ell \left( g\left( x^i \right) ,+1 \right) -\ell \left( g\left( x^i \right) ,-1 \right)}{a\pi _--b\pi _+}
$$
$$
\mathcal{L} _U\left( g\left( x \right) \right) \triangleq \frac{\left( -b\pi _+ \right) \ell \left( g\left( x \right) ,+1 \right) +\left( a\pi _- \right) \ell \left( g\left( x \right) ,-1 \right)}{a\pi _--b\pi _+}
$$

The denominator $a\pi _--b\pi _+$ in the customized loss function can be explicitly expanded as follows: 
\begin{align*}  
a\pi _--b\pi _+=&\frac{\pi _-\sum_{k=0}^{\lfloor M/2 \rfloor}{\left( \begin{array}{c}
			M-1\\
			k\\
		\end{array} \right) \pi _{+}^{M-k}\pi _{-}^{k}}}{\sum_{k=0}^{\lfloor M/2 \rfloor}{\left( \begin{array}{c}
			M\\
			k\\
		\end{array} \right) \pi _{+}^{M-k}\pi _{-}^{k}}}
\\&
-\frac{\pi _+\sum_{k=1}^{\lfloor M/2 \rfloor}{\left( \begin{array}{c}
			M-1\\
			k-1\\
		\end{array} \right) \pi _{+}^{M-k}\pi _{-}^{k}}}{\sum_{k=0}^{\lfloor M/2 \rfloor}{\left( \begin{array}{c}
			M\\
			k\\
		\end{array} \right) \pi _{+}^{M-k}\pi _{-}^{k}}}
\\
=&\frac{\sum_{k=0}^{\lfloor M/2 \rfloor}{\left( \begin{array}{c}
			M-1\\
			k\\
		\end{array} \right) \pi _{+}^{M-k}\pi _{-}^{k+1}}}{\sum_{k=0}^{\lfloor M/2 \rfloor}{\left( \begin{array}{c}
			M\\
			k\\
		\end{array} \right) \pi _{+}^{M-k}\pi _{-}^{k}}}
\\&
-\frac{\sum_{k=1}^{\lfloor M/2 \rfloor}{\left( \begin{array}{c}
			M-1\\
			k-1\\
		\end{array} \right) \pi _{+}^{M-k+1}\pi _{-}^{k}}}{\sum_{k=0}^{\lfloor M/2 \rfloor}{\left( \begin{array}{c}
			M\\
			k\\
		\end{array} \right) \pi _{+}^{M-k}\pi _{-}^{k}}}
\end{align*}   
If we let $k'=k-1$, then we have:
\begin{align*}
a\pi _--b\pi _+=&\frac{\sum_{k=0}^{\lfloor M/2 \rfloor}{\left( \begin{array}{c}
			M-1\\
			k\\
		\end{array} \right) \pi _{+}^{M-k}\pi _{-}^{k+1}}}{\sum_{k=0}^{\lfloor M/2 \rfloor}{\left( \begin{array}{c}
			M\\
			k\\
		\end{array} \right) \pi _{+}^{M-k}\pi _{-}^{k}}}
\\&
-\frac{\sum_{k'=0}^{\lfloor M/2 \rfloor -1}{\left( \begin{array}{c}
			M-1\\
			k'\\
		\end{array} \right) \pi _{+}^{M-k'}\pi _{-}^{k'+1}}}{\sum_{k=0}^{\lfloor M/2 \rfloor}{\left( \begin{array}{c}
			M\\
			k\\
		\end{array} \right) \pi _{+}^{M-k}\pi _{-}^{k}}}
\\
=&\frac{\left( \begin{array}{c}
		M-1\\
		\lfloor M/2 \rfloor\\
	\end{array} \right) \pi _{+}^{M-\lfloor M/2 \rfloor}\pi _{-}^{\lfloor M/2 \rfloor +1}}{\sum_{k=0}^{\lfloor M/2 \rfloor}{\left( \begin{array}{c}
			M\\
			k\\
		\end{array} \right) \pi _{+}^{M-k}\pi _{-}^{k}}}
\end{align*}    

Both the numerator $\left( \begin{array}{c}
	M-1\\
	\lfloor M/2 \rfloor\\
\end{array} \right) \pi _{+}^{M-\lfloor M/2 \rfloor}\pi _{-}^{\lfloor M/2 \rfloor +1}$ and denominator $\left( \begin{array}{c}
M-1\\
\lfloor M/2 \rfloor\\
\end{array} \right) \pi _{+}^{M-\lfloor M/2 \rfloor}\pi _{-}^{\lfloor M/2 \rfloor +1}$ remain strictly positive for all parity cases of $M$, ensuring the positivity of $a\pi _--b\pi _+$. However, due to the non-negativity constraint of the loss function, two negative terms (i.e., $\frac{-\ell \left( g\left( x^i \right) ,-1 \right)}{a\pi _--b\pi _+}$ and $\frac{\left( -b\pi _+ \right) \ell \left( g\left( x \right) ,+1 \right)}{a\pi _--b\pi _+}$) inherently exist in the custom-designed loss function, which may contribute to overfitting issues during model optimization. Extensive experiments demonstrate that when the training loss becomes negative, the test accuracy decreases to varying degrees, indicating a clear overfitting phenomenon (see Fig.(\ref{fig:3}) for experimental results based on the MNIST, EMNIST-Digits, and CIFAR-10). However, increasing weight decay or employing other methods to mitigate overfitting does not effectively address the issue. To resolve this, we propose a method using risk correction functions to reformulate the empirical risk function. 

Following the reference \cite{lu2020mitigating}, we select the absolute value correction function (ABS) and the rectified linear unit (ReLU) to encapsulate the empirical risk. The formulations of these two correction functions are given below:
\begin{equation}
	\tilde{R}\left( g \right) =f\left( \hat{R}_{MDPU}\left( g \right) \right)
\end{equation}
where $f\left( x \right) =\left\{ \begin{array}{c}
	x,      x\geq 0,\\
	k|x|,   x<0.\\
\end{array} \right. $ and $k>0$. 

Depending on the value of $k$, the correction function corresponds to either the rectified linear unit (ReLU) when $k=0$ or the absolute value correction function (ABS) when $k=1$, and the corrected empirical risk can be represented as follows:
\begin{equation}\label{eq-15}
	\tilde{R}_{ReLU}\left( g \right) =\max \left\{ 0,\hat{R}_{MDPU}\left( g \right) \right\}
\end{equation}
\begin{equation}\label{eq-16}
	\tilde{R}_{ABS}\left( g \right) =|\hat{R}_{MDPU}\left( g \right) |
\end{equation}

We subsequently present theoretical analysis to establish the consistency of the corrected risk estimator.
\begin{theorem}	\label{Th3} (Consistency of $
	\tilde{R}_{MDPU}\left( g \right)$)
	Suppose that there exists two positive constant $\zeta$ and $\eta $ satisfy that $R_{MDP}\left( g \right) \geq \zeta$ and $R_U\left( g \right) \geq \eta$. As stated in Theorem \ref{Th-2}, the bias of $\tilde{R}_{MDPU}(g)$ decreases at an exponential rate as $n \to \infty$:
\begin{equation}
\begin{aligned}
&\mathbb{E} [\tilde{R}_{MDPU}(g)]-R(g)
\\&
\leq \left[ \frac{\left( L_f+1 \right) \left( \pi _+\pi _- \right) MC_{\ell}}{\left( a\pi _--b\pi _+ \right)}+\left( L_f+1 \right) C_{\ell} \right] \varDelta
\end{aligned}
\end{equation}
where, 
\begin{align*}
\varDelta =&\exp \left( -\frac{2\zeta ^2M^2\left( a\pi _--b\pi _+ \right) ^2n_{MDP}}{\left( \pi _+\pi _- \right) ^2C_{\ell}^{2}} \right) \\&+\exp \left( -\frac{2\eta ^2n_U}{C_{\ell}^{2}} \right) 
\end{align*}
\end{theorem}

The following inequality holds with probability at least $1-\delta$:
\begin{equation}
	\begin{aligned} 
&\left| \tilde{R}_{MDPU}(g)-R(g) \right|
\\
\le& C_{\ell}L_f\sqrt{\frac{1}{2}\ln \frac{2}{\delta}\left( \frac{\left( \pi _+\pi _- \right) ^2}{M^2\left( a\pi _--b\pi _+ \right) ^2n_{MDP}}+\frac{1}{n_U} \right)}
\\&
+\left[ \frac{\left( L_f+1 \right) \left( \pi _+\pi _- \right) MC_{\ell}}{\left( a\pi _--b\pi _+ \right)}+\left( L_f+1 \right) C_{\ell} \right] \varDelta 
\end{aligned} 
\end{equation}
where, 
\begin{align*}
\varDelta =&\exp \left( -\frac{2\zeta ^2M^2\left( a\pi _--b\pi _+ \right) ^2n_{MDP}}{\left( \pi _+\pi _- \right) ^2C_{\ell}^{2}} \right) 
\\&
+\exp \left( -\frac{2\eta ^2n_U}{C_{\ell}^{2}} \right) 
\end{align*}

Through the analysis of Theorem \ref{Th3}, it can be concluded that as the number of training samples $n_{MDP}$ and $n_U$ approaches infinity, the corrected risk $\tilde{R}_{MDPU}(g)$ converges to the true risk $R(g)$ at an optimal rate of $\mathcal{O}_p(1/\sqrt{n_{MDP}} + 1/\sqrt{n_U})$. 

\section{Experiments}\label{Section:5}
This section experimentally validates the proposed \emph{MDPU} algorithm across multiple datasets. The algorithm employs two configurations with $M$ values set to 2 and 3, corresponding to``Pairwise Dominant Positive and Unlabeled Learning (Pairwise-DPU)" and ``Triple Dominant Positive and Unlabeled Learning (Triple-DPU)", respectively. The section systematically details dataset selection, experimental parameter configurations, and result analysis.

\begin{table*}
	\begin{center}
		\caption{The classification accuracy (mean ± standard deviation) of three \emph{Pairwise-DPU learning} methods is reported across 7 datasets under three distinct class-prior configurations, with results averaged over 100 training epochs and three independent experimental trials. The number of training samples for each dataset is configured as follows: $n=10,\!000$ for MNIST, Fashion-MNIST, EMNIST-Digits, CIFAR-10 and SVHN; $n=7,\!000$ for EMNIST-Letters; and $n=4,\!000$ for EMNIST-Balanced.}
		\label{table:2}
		\begin{tabular}{c|c|ccc|ccc} % <-- Alignments: 1st column left, 2nd middle and 3rd right, with vertical lines in between
			\hline
			Class-prior & Datasets & URE & ReLU & ABS & Siamese & Contrastive & K-Means \\
			\hline
			
			\multirow{7}{*}{$\pi _p=0.4$} 
			& MNIST & 87.91$\pm$0.40 & {\bf90.55$\pm$0.58}  & 89.90$\pm$0.93  & 55.12$\pm$2.92 & 53.55$\pm$1.05  & 55.50$\pm$1.12 \\
			
			& Fashion-MNIST & 92.03$\pm$0.37  & 93.71$\pm$0.38  & {\bf94.49$\pm$0.60}  & 65.19$\pm$0.34  & 54.54$\pm$1.71  & 56.36$\pm$0.22 \\
			
			& EMNIST-Digits & 88.52$\pm$0.78  & {\bf91.16$\pm$0.45}  & 90.73$\pm$0.41  &  60.70$\pm$1.49 & 57.54$\pm$1.09  & 54.70$\pm$0.41 \\
			
			& EMNIST-Letters & 83.71$\pm$0.53  & {\bf86.75$\pm$0.55}  & 85.50$\pm$0.45  & 52.77$\pm$1.57 &  51.32$\pm$0.71 & 53.86$\pm$4.56 \\
			
			& EMNIST-Balanced & 86.92$\pm$0.95  & {\bf88.95$\pm$0.36}  & 88.18$\pm$0.38  & 52.26$\pm$0.34  & 51.19$\pm$0.29  & 54.17$\pm$2.19 \\
			
			& CIFAR-10 & 57.69$\pm$6.56  & {\bf72.06$\pm$2.88}  & 71.33$\pm$3.54  & 54.07$\pm$1.65  & 59.73$\pm$0.99  & 50.52$\pm$2.06 \\
			
			& SVHN & 60.17$\pm$4.26  & 70.31$\pm$3.05  & {\bf73.55$\pm$1.42}  & 53.61$\pm$4.57  & 62.26$\pm$1.83  & 51.38$\pm$3.78 \\
			\hline

			\multirow{7}{*}{$\pi _p=0.5$} 
			& MNIST  & 85.01$\pm$1.07  & {\bf88.99$\pm$0.63}  & 88.76$\pm$0.67  & 57.39$\pm$5.44  & 51.33$\pm$0.43  & 53.42$\pm$1.10 \\
			
			& Fashion-MNIST & 92.00$\pm$0.71  & 94.45$\pm$0.37  & {\bf94.86$\pm$0.21}  & 61.36$\pm$3.79  & 56.51$\pm$5.06  & 51.97$\pm$6.31 \\
			
			& EMNIST-Digits & 86.28$\pm$0.80  & {\bf90.10$\pm$0.35}  & 90.05$\pm$0.35  & 61.02$\pm$1.17  & 56.67$\pm$3.72  & 55.58$\pm$0.11 \\
			
			& EMNIST-Letters & 81.80$\pm$2.11  & {\bf84.41$\pm$1.74}  & 84.09$\pm$1.72  & 53.63$\pm$1.78  & 51.77$\pm$0.37  & 57.49$\pm$0.24 \\
			
			& EMNIST-Balanced & 86.64$\pm$0.31  & 87.92$\pm$0.29  & {\bf88.20$\pm$0.38}  & 51.47$\pm$0.76  & 51.59$\pm$1.26  & 50.98$\pm$0.34 \\
			
			& CIFAR-10 & 61.94$\pm$2.67  & {\bf71.91$\pm$0.83}  & 68.01$\pm$4.01  & 51.98$\pm$0.95  & 58.30$\pm$0.36  & 50.88$\pm$0.96\\
			
			& SVHN & 58.29$\pm$4.30  & {\bf69.22$\pm$1.34}  & 67.57$\pm$2.66  & 54.25$\pm$0.76  & 60.50$\pm$1.16 & 51.35$\pm$2.88\\
			\hline

			\multirow{6}{*}{$\pi _p=0.6$} 
			& MNIST  & 78.04$\pm$1.22  & {\bf85.10$\pm$3.41}  & 82.80$\pm$4.35  & 57.91$\pm$6.30  & 51.88$\pm$0.75  & 52.63$\pm$1.01 \\
			
			& Fashion-MNIST & 88.62$\pm$2.38  & {\bf92.82$\pm$0.92}  & 92.42$\pm$0.41  & 60.52$\pm$2.36  & 54.83$\pm$5.14  & 48.43$\pm$6.36 \\
			
			& EMNIST-Digits & 79.61$\pm$1.95  & {\bf85.83$\pm$0.61}  & 85.25$\pm$1.28  & 60.03$\pm$2.15  & 54.39$\pm$1.64  & 55.10$\pm$0.25 \\
			
			& EMNIST-Letters & 74.40$\pm$1.14  & 79.15$\pm$1.51  & {\bf79.68$\pm$1.22}  & 52.92$\pm$4.71  & 51.62$\pm$1.65  & 56.01$\pm$2.37 \\
			
			& EMNIST-Balanced & 77.63$\pm$2.77  & {\bf79.50$\pm$4.02}  & 78.63$\pm$4.01  & 52.55$\pm$1.40  & 51.02$\pm$4.24  & 50.29$\pm$0.23 \\
			
			& CIFAR-10 & 57.73$\pm$1.20  & 65.30$\pm$1.34  & {\bf66.74$\pm$3.76}  & 53.96$\pm$1.70  & 58.10$\pm$0.45  & 52.61$\pm$0.83 \\
			
			& SVHN & 58.82$\pm$3.56  & {\bf68.95$\pm$0.50}  & 68.75$\pm$2.68  & 55.39$\pm$2.50  & 56.69$\pm$2.93  & 46.28$\pm$0.55 \\
			\hline
		\end{tabular}
	\end{center}
\end{table*}

\begin{table*}
	\begin{center}
		\caption{The classification accuracy (mean ± standard deviation) of three \emph{Pairwise-DPU learning} methods is reported across 7 datasets under three distinct class-prior configurations, with results averaged over 100 training epochs and three independent experimental trials. The number of training samples for each dataset is configured as follows: $n=12,\!000$ for MNIST, Fashion-MNIST, EMNIST-Digits, CIFAR-10 and SVHN; $n=8,\!000$ for EMNIST-Letters; and $n=5,\!000$ for EMNIST-Balanced.}
		\label{table:3}
		\begin{tabular}{c|c|ccc|ccc} % <-- Alignments: 1st column left, 2nd middle and 3rd right, with vertical lines in between
			\hline
			Class-prior & Datasets & URE & ReLU & ABS & Siamese & Contrastive & K-Means \\
			\hline
			
			\multirow{7}{*}{$\pi _p=0.4$} 
			& MNIST  & 87.72$\pm$0.20  & {\bf90.35$\pm$0.29}  & 89.57$\pm$0.83  & 54.49$\pm$2.01  & 51.89$\pm$1.66  & 54.76$\pm$0.36\\
			
			& Fashion-MNIST & 92.24$\pm$0.27  & 94.47$\pm$0.25  & {\bf95.05$\pm$0.24}  & 61.09$\pm$0.81  & 51.78$\pm$2.10  & 56.77$\pm$0.27\\
			
			& EMNIST-Digits & 89.32$\pm$0.78  & {\bf90.94$\pm$1.02}  & 90.43$\pm$0.35  & 64.19$\pm$2.39  & 58.16$\pm$0.74  & 55.02$\pm$0.53\\
			
			& EMNIST-Letters & 83.42$\pm$0.44  & {\bf86.71$\pm$0.25}  & 85.94$\pm$0.71  & 55.54$\pm$2.23  & 51.01$\pm$0.47  & 55.57$\pm$2.38\\
			
			& EMNIST-Balanced & 86.49$\pm$0.12  & {\bf89.10$\pm$0.88}  & 88.77$\pm$1.24  & 53.66$\pm$1.08  & 51.16$\pm$0.67  & 54.12$\pm$2.33\\
			
			& CIFAR-10 & 57.43$\pm$4.67  & 74.64$\pm$0.44  & {\bf75.98$\pm$1.38}  & 54.52$\pm$1.72  & 66.29$\pm$2.60  & 52.46$\pm$0.43\\
			
			& SVHN & 63.10$\pm$6.50  & {\bf69.41$\pm$2.56}  & 67.53$\pm$1.49  & 56.22$\pm$3.20  & 64.56$\pm$2.42  & 54.52$\pm$0.19\\
			\hline

			\multirow{7}{*}{$\pi _p=0.5$} 
			& MNIST  & 86.31$\pm$0.34  & {\bf89.37$\pm$0.11}  & 89.09$\pm$0.84  & 56.95$\pm$1.66  & 52.27$\pm$1.65  & 54.35$\pm$0.29\\
			
			& Fashion-MNIST & 92.56$\pm$0.46  & 94.44$\pm$0.55  & {\bf94.66$\pm$0.70}  & 64.12$\pm$3.21  & 54.97$\pm$3.12  & 57.10$\pm$0.42\\
			
			& EMNIST-Digits & 88.09$\pm$1.00  & 90.91$\pm$0.57  & {\bf90.95$\pm$0.39}  & 61.67$\pm$3.15  & 55.25$\pm$0.43  & 55.68$\pm$0.52\\
			
			& EMNIST-Letters & 82.79$\pm$0.40  & 84.68$\pm$0.34  & {\bf84.85$\pm$0.56}  & 51.68$\pm$0.26  & 50.69$\pm$0.68  & 56.95$\pm$0.43\\
			
			& EMNIST-Balanced & 85.41$\pm$0.42  & 87.81$\pm$0.64  & {\bf88.11$\pm$0.61}  & 51.49$\pm$0.74  & 51.22$\pm$0.64  & 52.26$\pm$2.95\\
			
			& CIFAR-10 & 57.07$\pm$1.20  & {\bf71.75$\pm$3.77}  & 69.98$\pm$3.03  & 51.62$\pm$0.70  & 59.43$\pm$0.18  & 52.54$\pm$0.71\\
			
			& SVHN & 58.27$\pm$5.36  & {\bf70.77$\pm$0.55}  & 69.69$\pm$3.58  & 56.91$\pm$3.69  & 63.24$\pm$4.78  & 54.03$\pm$0.25\\
			\hline

			\multirow{6}{*}{$\pi _p=0.6$} 
			& MNIST  & 79.06$\pm$0.48  & {\bf85.65$\pm$0.39}  & 84.23$\pm$0.61  & 56.54$\pm$4.37  & 53.24$\pm$1.92  & 54.29$\pm$0.55\\
			
			& Fashion-MNIST & 86.72$\pm$3.61  & 92.32$\pm$0.57  & {\bf93.08$\pm$0.13}  & 61.44$\pm$2.85  & 54.11$\pm$1.87  & 57.03$\pm$0.57\\
			
			& EMNIST-Digits & 80.44$\pm$2.53  & {\bf86.42$\pm$1.74}  & 84.96$\pm$2.05  & 63.14$\pm$1.35  & 55.52$\pm$1.54  & 48.90$\pm$5.35\\
			
			& EMNIST-Letters & 76.27$\pm$1.37  & 80.17$\pm$0.81  & {\bf80.42$\pm$0.21}  & 52.78$\pm$0.84  & 50.97$\pm$0.51  & 57.50$\pm$0.58\\
			
			& EMNIST-Balanced & 74.59$\pm$2.95  & {\bf79.01$\pm$2.34}  & 78.69$\pm$1.42  & 52.66$\pm$2.09  & 51.26$\pm$0.29  & 56.07$\pm$0.68\\
			
			& CIFAR-10 & 56.08$\pm$0.94  & 67.68$\pm$1.53  & {\bf70.25$\pm$2.66}  & 54.40$\pm$3.83  & 59.42$\pm$0.62  & 51.34$\pm$0.25\\
			
			& SVHN & 57.03$\pm$1.97  & {\bf70.10$\pm$0.77}  & 69.09$\pm$1.03  & 53.91$\pm$3.31  & 58.11$\pm$0.65  & 54.24$\pm$0.37\\
			\hline
		\end{tabular}
	\end{center}
\end{table*}

\begin{table*}
	\begin{center}
		\caption{The classification accuracy (mean ± standard deviation) of three \emph{Triple-DPU learning} methods is reported across 7 datasets under three distinct class-prior configurations, with results averaged over 100 training epochs and three independent experimental trials. The number of training samples for each dataset is configured as follows: $n=10,\!000$ for MNIST, Fashion-MNIST, EMNIST-Digits, CIFAR-10 and SVHN; $n=7,\!000$ for EMNIST-Letters; and $n=4,\!000$ for EMNIST-Balanced.}
		\label{table:4}
		\begin{tabular}{c|c|ccc|ccc} % <-- Alignments: 1st column left, 2nd middle and 3rd right, with vertical lines in between
			\hline
			Class-prior & Datasets & URE & ReLU & ABS & Siamese & Contrastive & K-Means \\
			\hline
			
			\multirow{7}{*}{$\pi _p=0.4$} 
			& MNIST  & 90.85$\pm$0.66  & {\bf92.09$\pm$0.44}  & 91.84$\pm$0.13  & 62.35$\pm$2.64  & 51.14$\pm$0.79  & 72.82$\pm$0.31\\
			
			& Fashion-MNIST & 94.23$\pm$0.33  & 95.13$\pm$0.20  & {\bf95.26$\pm$0.41}  & 57.96$\pm$1.25  & 51.96$\pm$1.74  & 59.74$\pm$0.30\\
			
			& EMNIST-Digits & 91.34$\pm$0.34  & {\bf93.30$\pm$0.11}  & 93.12$\pm$0.10  & 64.07$\pm$2.23  & 51.37$\pm$1.63  & 54.11$\pm$0.60\\
			
			& EMNIST-Letters & 87.98$\pm$0.60  & {\bf89.20$\pm$0.36}  & 89.06$\pm$0.21  & 54.22$\pm$3.74  & 50.40$\pm$0.57 & 56.71$\pm$1.95\\
			
			& EMNIST-Balanced & 89.64$\pm$1.01  & {\bf90.69$\pm$0.69}  & 90.11$\pm$0.25  & 57.23$\pm$2.00  & 51.43$\pm$0.81  & 57.87$\pm$0.91\\
			
			& CIFAR-10 & 72.12$\pm$0.41  & 73.35$\pm$2.77  & {\bf74.65$\pm$0.70}  & 57.23$\pm$2.00  & 54.60$\pm$1.74  & 52.69$\pm$0.54\\
			
			& SVHN & 71.88$\pm$0.08  & 72.88$\pm$2.07  & {\bf73.58$\pm$1.74}  & 60.17$\pm$0.84  & 54.47$\pm$2.22 & 62.72$\pm$1.05\\
			\hline

			\multirow{7}{*}{$\pi _p=0.5$} 
			& MNIST  & 87.80$\pm$0.56  & 90.59$\pm$0.77  & {\bf90.67$\pm$0.32}  & 61.09$\pm$4.99  & 50.72$\pm$0.34  & 72.44$\pm$0.71\\
			
			& Fashion-MNIST & 92.23$\pm$0.22  & 94.56$\pm$0.21  & {\bf95.26$\pm$0.41}  & 57.45$\pm$2.90  & 54.68$\pm$5.39  & 59.85$\pm$0.42\\
			
			& EMNIST-Digits & 88.69$\pm$0.39  & 91.41$\pm$0.79  & {\bf92.16$\pm$0.17}  & 64.70$\pm$1.35  & 51.37$\pm$1.94  & 53.78$\pm$1.33\\
			
			& EMNIST-Letters & 85.14$\pm$1.24  & {\bf87.46$\pm$0.39}  & 87.32$\pm$0.42  & 53.88$\pm$2.41  & 50.94$\pm$0.90  & 57.77$\pm$0.75\\
			
			& EMNIST-Balanced & 86.10$\pm$1.05  & 90.09$\pm$0.98  & {\bf90.34$\pm$0.64}  & 50.98$\pm$0.24  & 51.29$\pm$0.43  & 58.02$\pm$0.31\\
			
			& CIFAR-10 & 65.49$\pm$1.24  & {\bf71.30$\pm$1.65}  & 70.26$\pm$3.27  & 53.77$\pm$3.72  & 53.29$\pm$0.74  & 52.18$\pm$1.04\\
			
			& SVHN & 66.59$\pm$3.20  & 72.50$\pm$0.97  & {\bf74.38$\pm$1.78}  & 54.53$\pm$3.87  & 56.09$\pm$3.93  & 60.41$\pm$0.68\\
			\hline

			\multirow{6}{*}{$\pi _p=0.6$} 
			& MNIST  & 85.17$\pm$0.70  & {\bf88.95$\pm$0.47}  & 88.70$\pm$0.45  & 63.10$\pm$4.08  & 52.00$\pm$1.80  & 73.74$\pm$0.35\\
			
			& Fashion-MNIST & 91.51$\pm$0.64  & 94.52$\pm$0.39  & {\bf94.76$\pm$0.33}  & 56.83$\pm$1.37  & 58.34$\pm$3.33  & 59.80$\pm$0.26\\
			
			& EMNIST-Digits & 86.04$\pm$0.92  & 91.03$\pm$0.25  & {\bf91.17$\pm$0.57}  & 60.52$\pm$0.50  & 50.78$\pm$1.10  & 53.28$\pm$0.16\\
			
			& EMNIST-Letters & 81.55$\pm$0.54  & {\bf84.95$\pm$0.76}  & 84.93$\pm$0.78  & 56.19$\pm$1.01  & 50.24$\pm$0.33  & 56.82$\pm$0.40\\
			
			& EMNIST-Balanced & 83.52$\pm$0.82  & {\bf87.54$\pm$0.85}  & 87.37$\pm$0.51  & 51.19$\pm$0.02  & 51.99$\pm$0.67  & 58.42$\pm$0.40\\
			
			& CIFAR-10 & 60.72$\pm$1.53  & {\bf71.26$\pm$0.92}  & 70.07$\pm$1.60  & 56.93$\pm$4.88  & 52.46$\pm$1.74  & 55.00$\pm$0.36\\
			
			& SVHN & 62.25$\pm$3.72  & {\bf73.75$\pm$1.43}  & 70.29$\pm$1.39  & 58.83$\pm$0.77  & 51.51$\pm$1.67  & 62.99$\pm$1.50\\
			\hline
		\end{tabular}
	\end{center}
\end{table*}

\begin{table*}
	\begin{center}
		\caption{The classification accuracy (mean ± standard deviation) of three \emph{Triple-DPU learning} methods is reported across 7 datasets under three distinct class-prior configurations, with results averaged over 100 training epochs and three independent experimental trials. The number of training samples for each dataset is configured as follows: $n=12,\!000$ for MNIST, Fashion-MNIST, EMNIST-Digits, CIFAR-10 and SVHN; $n=8,\!000$ for EMNIST-Letters; and $n=5,\!000$ for EMNIST-Balanced.}
		\label{table:5}
		\begin{tabular}{c|c|ccc|ccc} % <-- Alignments: 1st column left, 2nd middle and 3rd right, with vertical lines in between
			\hline
			Class-prior & Datasets & URE & ReLU & ABS & Siamese & Contrastive & K-Means \\
			\hline
			
			\multirow{7}{*}{$\pi _p=0.4$} 
			& MNIST  & 90.83$\pm$0.34  & {\bf92.39$\pm$0.39}  & 92.19$\pm$0.26  & 66.76$\pm$6.18  & 51.07$\pm$0.48  & 72.93$\pm$0.41\\
			
			& Fashion-MNIST & 94.06$\pm$0.22  & {\bf95.34$\pm$0.33}  & 92.19$\pm$0.26  & 59.10$\pm$3.04  & 52.00$\pm$1.80  & 59.65$\pm$0.28\\
			
			& EMNIST-Digits & 91.37$\pm$0.05  & 93.11$\pm$0.06  & {\bf93.15$\pm$0.25}  & 63.39$\pm$1.35  & 50.35$\pm$0.49  & 54.43$\pm$0.11\\
			
			& EMNIST-Letters & 88.35$\pm$0.66  & 89.76$\pm$0.26  & {\bf89.67$\pm$0.28}  & 51.86$\pm$1.09  & 51.92$\pm$1.39  & 56.10$\pm$1.86\\
			
			& EMNIST-Balanced & 90.81$\pm$0.56  & {\bf93.11$\pm$0.06}  & 91.27$\pm$0.25  & 51.61$\pm$0.33  & 50.73$\pm$0.46  & 58.03$\pm$0.96\\
			
			& CIFAR-10 & 70.09$\pm$3.31  & {\bf72.96$\pm$1.46}  & 70.45$\pm$1.26  & 54.93$\pm$2.71  & 52.32$\pm$1.59  & 52.79$\pm$0.41\\
			
			& SVHN & 73.90$\pm$0.65  & 74.67$\pm$1.67  & {\bf76.14$\pm$1.43} & 60.15$\pm$0.06  & 53.47$\pm$2.40  & 62.63$\pm$1.03\\
			\hline

			\multirow{7}{*}{$\pi _p=0.5$} 
			& MNIST  & 88.28$\pm$0.42  & 91.18$\pm$0.08  & {\bf91.60$\pm$0.20}  & 70.13$\pm$1.56  & 51.00$\pm$0.38  & 71.98$\pm$0.59\\
			
			& Fashion-MNIST & 92.92$\pm$0.62  & 94.67$\pm$0.36  & {\bf95.64$\pm$0.07}  & 56.43$\pm$2.10  & 50.73$\pm$0.47  & 59.83$\pm$0.27\\
			
			& EMNIST-Digits & 89.88$\pm$0.63  & 92.49$\pm$0.46  & {\bf92.85$\pm$0.58} & 63.12$\pm$1.52  & 51.53$\pm$1.08  & 53.80$\pm$1.33\\
			
			& EMNIST-Letters & 84.48$\pm$1.55  & 87.41$\pm$1.04  & {\bf87.51$\pm$1.35}  & 62.62$\pm$2.62  & 50.34$\pm$0.48  & 56.42$\pm$0.86\\
			
			& EMNIST-Balanced & 87.41$\pm$0.32  & 91.20$\pm$0.47  & {\bf91.30$\pm$0.38}  & 51.45$\pm$0.30  & 51.51$\pm$1.74  & 58.60$\pm$0.65\\
			
			& CIFAR-10 & 64.85$\pm$6.40  & {\bf75.84$\pm$2.91}  & 75.72$\pm$1.56  & 59.54$\pm$0.78  & 51.52$\pm$0.93  & 53.83$\pm$1.14\\
			
			& SVHN & 66.77$\pm$0.28  & 75.71$\pm$1.89  & {\bf75.95$\pm$0.87}  & 57.93$\pm$1.65  & 53.00$\pm$1.81 & 60.75$\pm$1.06\\
			\hline

			\multirow{6}{*}{$\pi _p=0.6$} 
			& MNIST  & 86.40$\pm$0.25  & {\bf90.32$\pm$0.26}  & 90.12$\pm$0.23  & 64.71$\pm$3.66  & 50.65$\pm$0.11  & 73.01$\pm$0.46\\
			
			& Fashion-MNIST & 92.48$\pm$0.58  & 94.60$\pm$0.20  & {\bf94.90$\pm$0.40}  & 59.30$\pm$1.01  & 51.07$\pm$0.48  & 59.87$\pm$0.30\\
			
			& EMNIST-Digits & 87.28$\pm$0.24  & 90.66$\pm$0.61  & {\bf91.02$\pm$0.55}  & 62.03$\pm$1.30  & 53.54$\pm$2.29  & 54.71$\pm$1.41\\
			
			& EMNIST-Letters & 83.33$\pm$0.77  & {\bf86.05$\pm$0.38}  & 85.84$\pm$0.58  & 54.65$\pm$2.27  & 51.75$\pm$1.54  & 56.81$\pm$0.62\\
			
			& EMNIST-Balanced & 84.74$\pm$0.92  & {\bf87.94$\pm$0.78}  & 87.90$\pm$0.82  & 51.65$\pm$0.26  & 50.88$\pm$0.24  & 58.68$\pm$0.85\\
			
			& CIFAR-10 & 59.67$\pm$2.58  & {\bf74.07$\pm$1.67}  & 71.71$\pm$1.46  & 51.62$\pm$0.70  & 53.38$\pm$2.97  & 54.08$\pm$0.43\\
			
			& SVHN & 65.52$\pm$1.91  & {\bf72.77$\pm$1.66}  & 72.66$\pm$2.85  & 53.31$\pm$2.01  & 57.13$\pm$2.14  & 62.45$\pm$1.30\\
			\hline
		\end{tabular}
	\end{center}
\end{table*}

\subsection{Datasets}
In this paper, seven datasets were chosen to experimentally verify the proposed method's validity and efficacy. Given that each dataset contains samples from multiple distinct classes, they are typically employed for multi-classification tasks. Nevertheless, to conform to the binary classification objective of this research, the samples were manually reclassified.

{\bf MNIST:}\cite{lecun_gradient-based_1998} This dataset comprises grayscale images of handwritten digits, with each image having an original feature dimension of 28$\times$28 pixels. It contains 60,000 training samples and 10,000 test samples, covering digits from 0 to 9. For our binary classification experiments, we designate the digits in even-numbered positions \{0, 2, 4, 6, 8\} as the positive class and the digits in odd-numbered positions \{1, 3, 5, 7, 9\} as the negative class.

{\bf Fashion-MNIST:}\cite{xiao2017fashion} This dataset consists of grayscale images of fashion items, each with a feature dimension of 28$\times$28. It includes 60,000 training samples and 10,000 testing samples. The label space contains the following categories: \{`T-shirt', `Pullover', `Dress', `Shirt', `Trouser', `Coat', `Sandal', `Sneaker', `Bag', `Ankle boot'\}. In our experimental framework, the classes \{`T-shirt', `Pullover', `Shirt', `Bag', `Coat'\} are labeled as the positive class, while the remaining categories \{`Dress', `Trouser', `Sandal', `Sneaker', `Ankle boot'\} are treated as the negative class.

{\bf EMNIST-Digits:}\cite{cohen2017emnist} The EMNIST-Digits dataset is a subset of the EMNIST dataset, featuring grayscale images of handwritten digits with an original feature dimension of 28x28. It comprises 240,000 training examples and 40,000 test examples. The label space includes digits 0-9. In our experimental setup, digits in even-numbered positions \{0, 2, 4, 6, 8\} are classified as the positive class, while digits in odd-numbered positions \{1, 3, 5, 7, 9\} are designated as the negative class.

{\bf EMNIST-Letters:}\cite{cohen2017emnist} The EMNIST-Letters dataset is a subset of the EMNIST dataset, consisting of grayscale images of handwritten uppercase letters with a spatial resolution of 28x28 pixels. It contains 124,800 training samples and 20,800 test samples. The label space spans all uppercase letters from A to Z. In our experimental configuration, letters occupying even ordinal positions in the alphabet \{A, C, E, G, I, K, M, O, Q, S, U, W, Y\} are designated as the positive class, while those in odd positions \{B, D, F, H, J, L, N, P, R, T, V, X, Z\} form the negative class.

{\bf EMNIST-Balanced:}\cite{cohen2017emnist} The EMNIST-Balanced dataset, a subset of the EMNIST collection, consists of grayscale images of handwritten digits (0-9) and uppercase letters (A-Z), with an original feature dimension of 28x28. It contains 112,800 training samples and 18,800 test samples, and the class distribution is balanced. The label space includes digits 0-9 and letters A-Z. In the experimental setup, classes are grouped by parity: the positive class consists of even - numbered digits {0, 2, 4, 6, 8} and uppercase letters at even ordinal positions (assuming "A" starts at position 0: {A, C, E, G, I, K, M, O, Q, S, U, W, Y}); the negative class consists of odd - numbered digits {1, 3, 5, 7, 9} and uppercase letters at odd ordinal positions ({B, D, F, H, J, L, N, P, R, T, V, X, Z}).

{\bf CIFAR-10:}\cite{krizhevsky2009learning} This dataset consists of 32$\times$32$\times$3 color images of various objects, with a total of 60,000 training examples and 10,000 test examples. The label space includes the following categories: \{`airplane', `automobile', `bird', `cat', `deer', `dog', `frog', `horse', `ship', `truck'\}. In our experimental framework, the classes \{`bird', `cat', `deer', `dog', `frog', `horse'\} are defined as the positive class, while \{`airplane', `automobile', `ship', `truck'\} are classified as the negative class.

{\bf SVHN:}\cite{37648} The SVHN dataset features images of house numbers with an original feature dimension of 32x32x3. It consists of 73,257 training examples and 26,032 test examples. The label space includes the digits: \{`0', `1', `2', `3', `4', `5', `6', `7', `8', `9'\}. In our
experimental setting, digits \{`2', `3', `4', `5', `6', `7'\} are defined as the positive class, while digits \{`0', `1', `8', `9'\} are classified as the negative class.

\begin{figure*}
	\centering
	\scriptsize
	\begin{tabular}{ccc}
		\includegraphics[width=5.5cm]{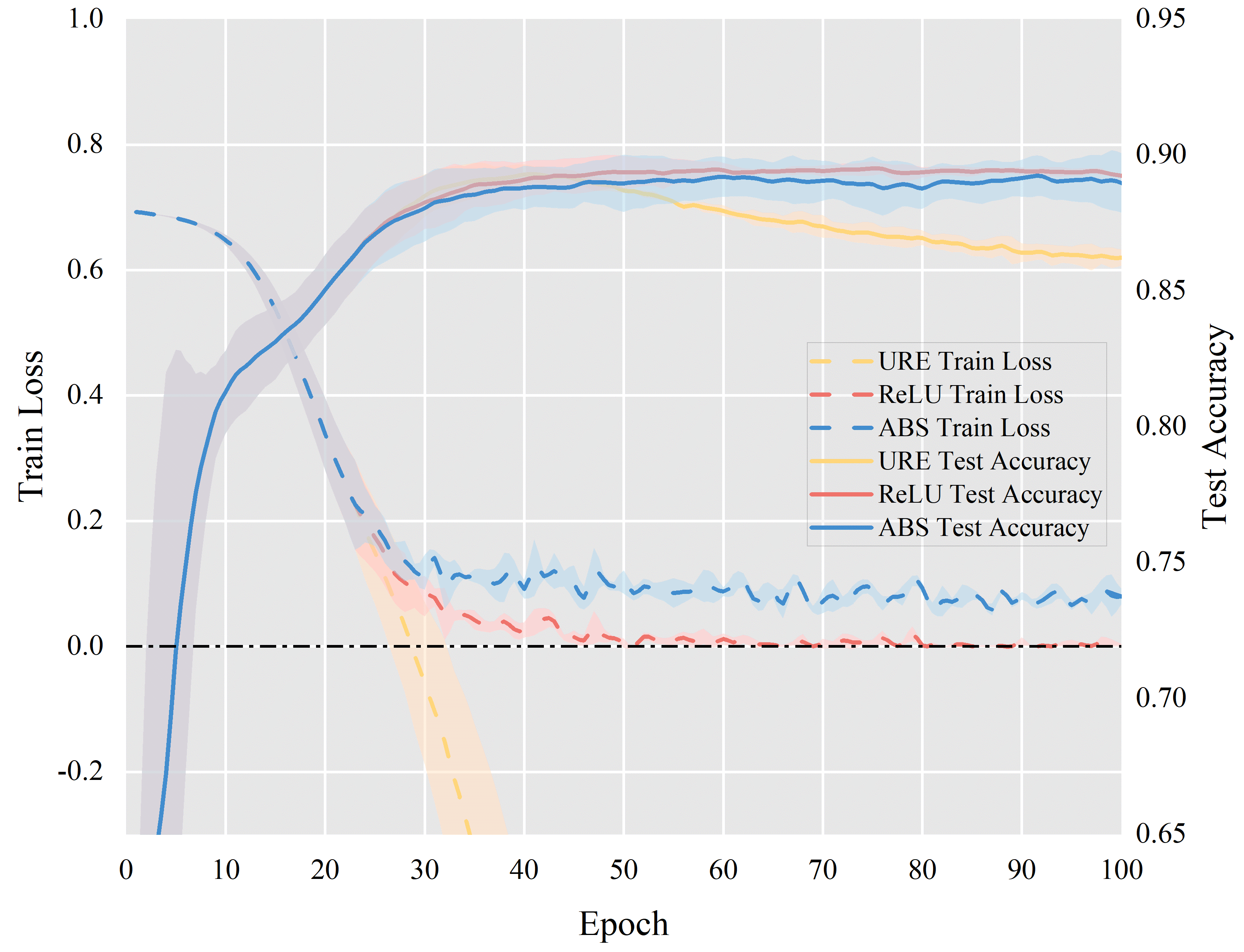} & 
		\includegraphics[width=5.5cm]{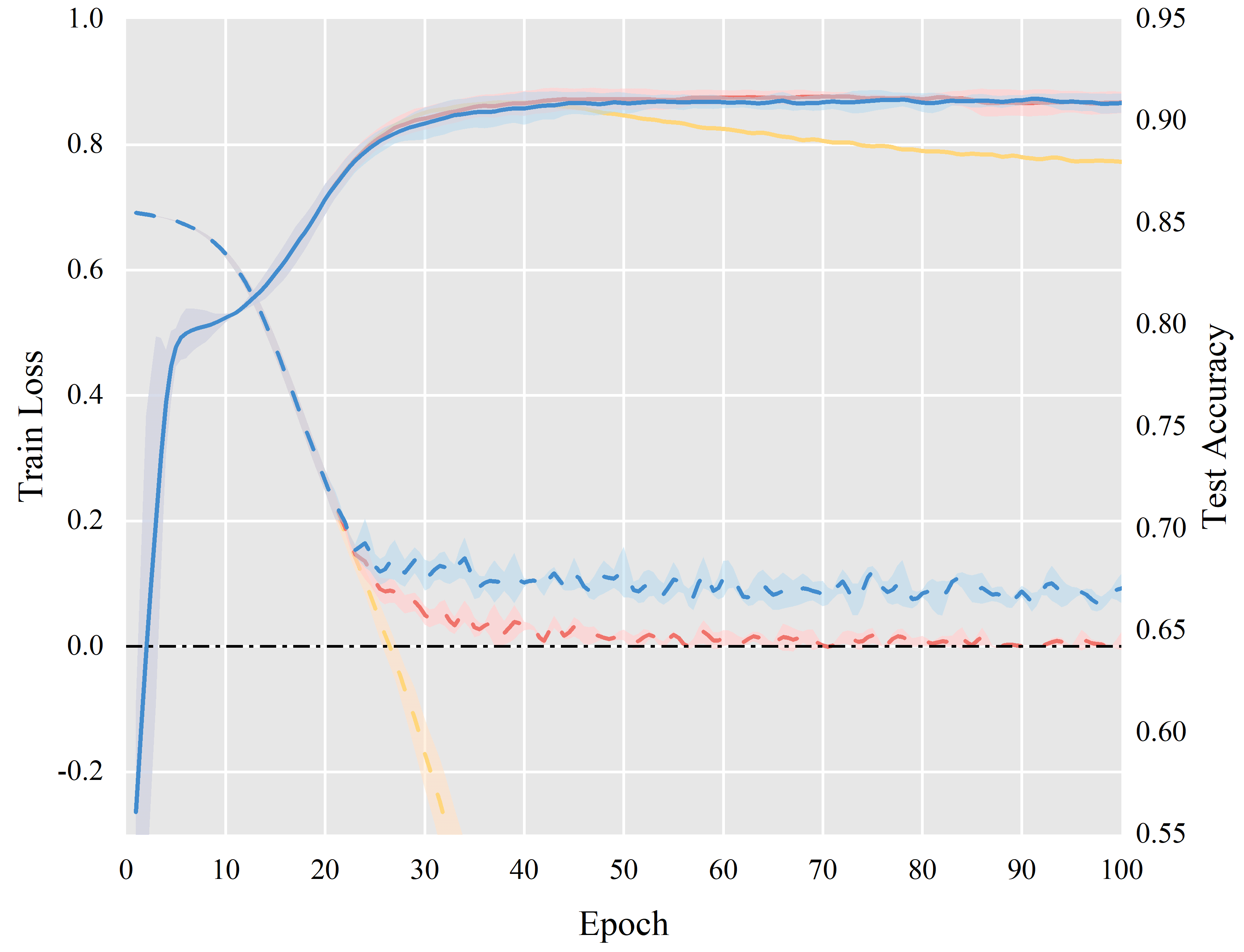} &
		\includegraphics[width=5.5cm]{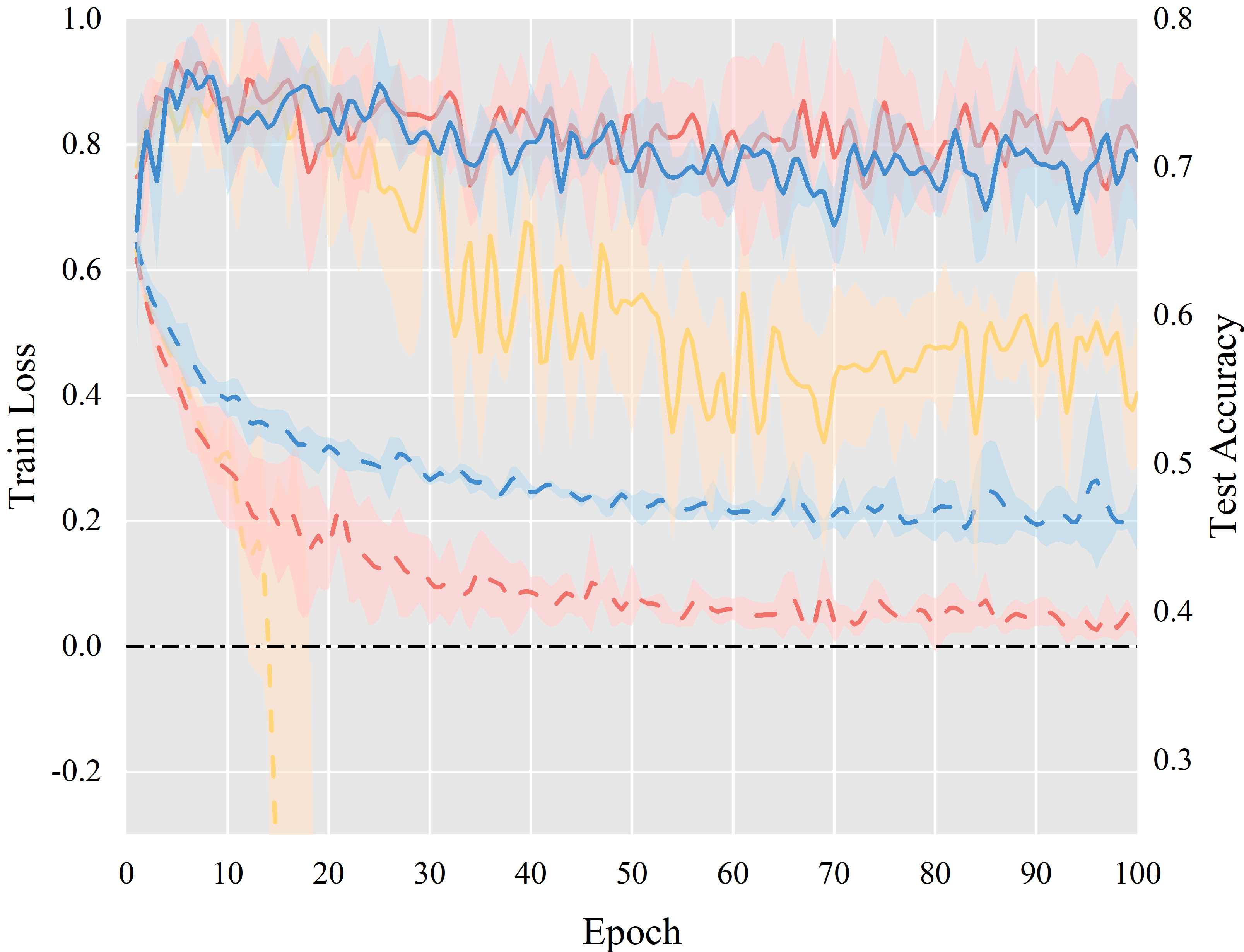} \\
		(a.1) (Pairwise) MNIST with $\pi_p=0.5$ & (a.2) (Pairwise) EMNIST-Digits with $\pi_p=0.5$ & (a.3) (Pairwise) CIFAR-10 with $\pi_p=0.5$T \\ 
		
		\includegraphics[width=5.5cm]{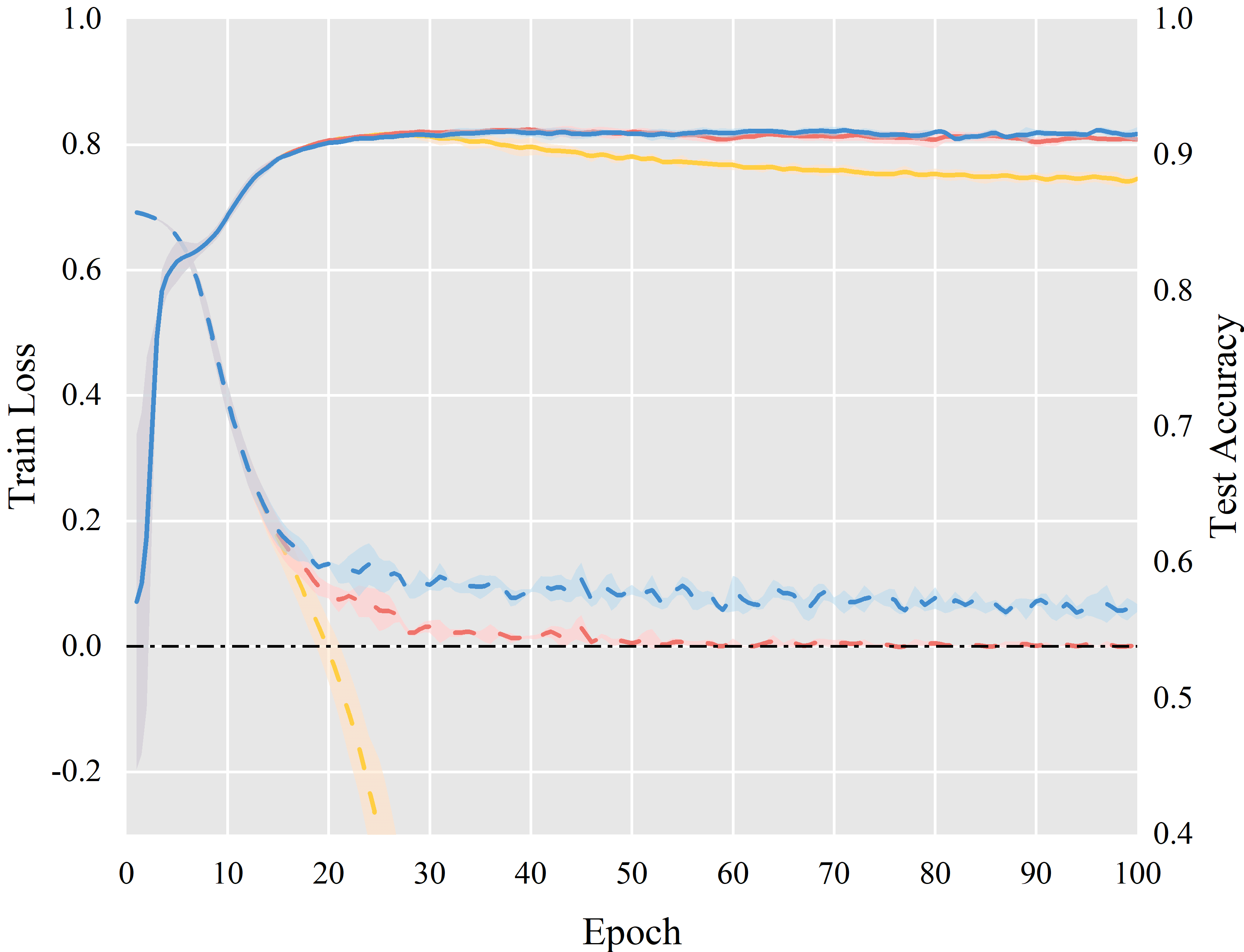} & 
		\includegraphics[width=5.5cm]{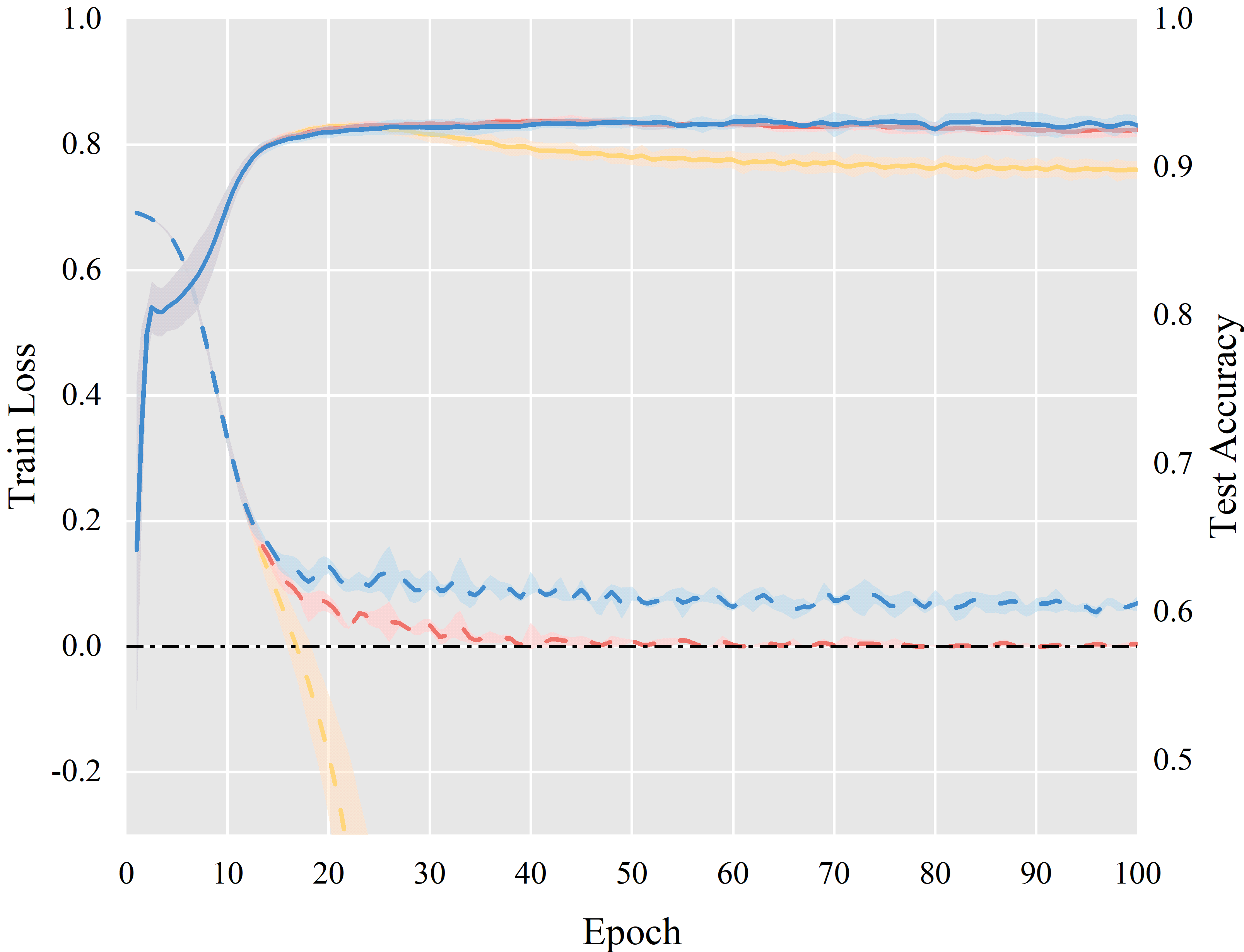} &
		\includegraphics[width=5.5cm]{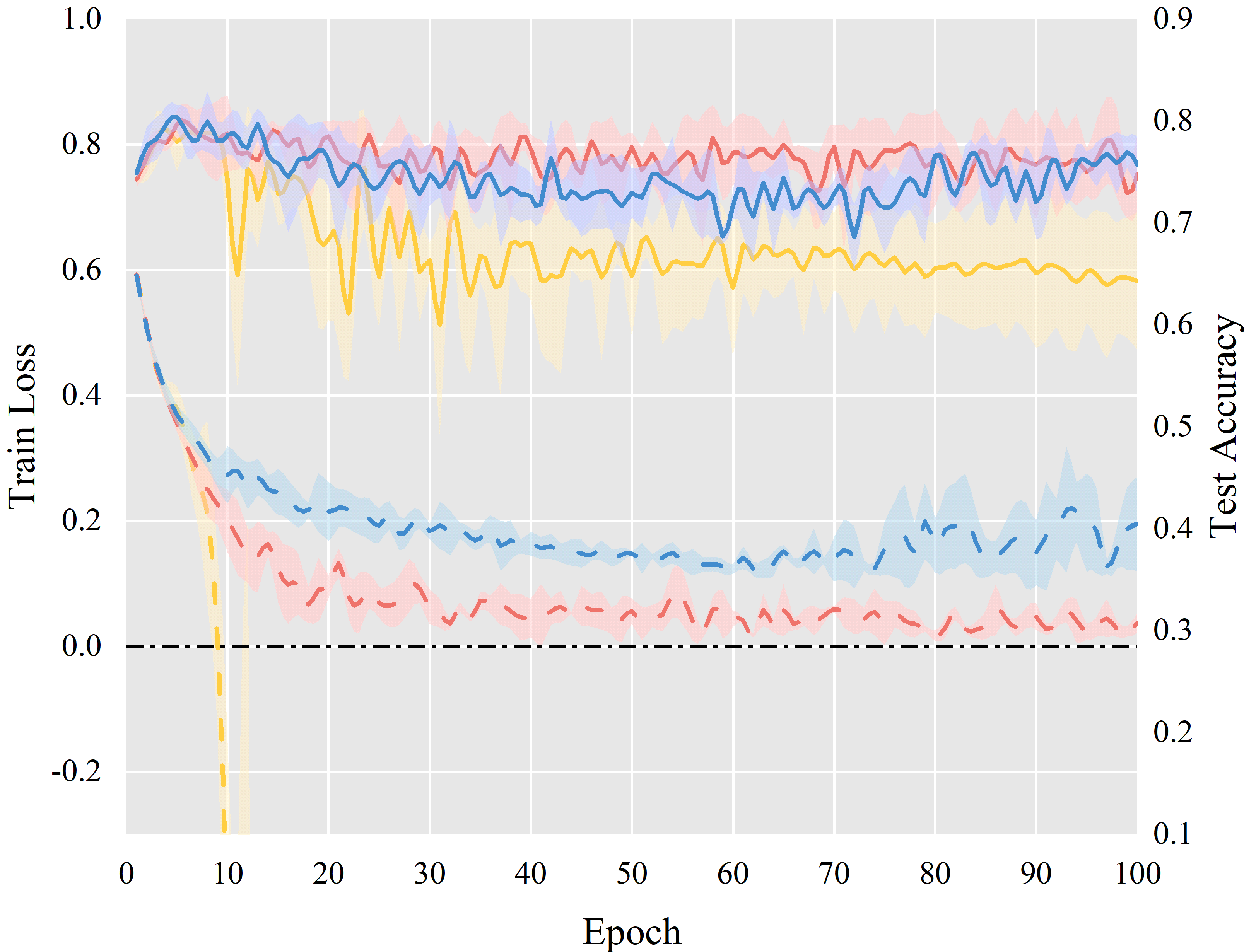} \\
		(b.1) (Triple) MNIST with $\pi_p=0.5$ & (b.2) (Triple) EMNIST-Digits with $\pi_p=0.5$ & (b.3) (Triple) CIFAR-10 with $\pi_p=0.5$ \\
		
	\end{tabular}
	\caption{Classification performance of Pairwise-MDPU Learning and Triple-MDPU Learning with logistic loss under $\pi _+=0.5$ on MNIST, EMNIST-Digits, and CIFAR-10 datasets. The mean accuracy is represented by dark colors, while the standard deviation is indicated by light colors. Specifically, (a.1)-(a.3) denotes the outcomes of Pairwise-DPU learning on MNIST, EMNIST-Digits and CIFAR-10; (b.1)-(b.3) denotes the outcomes of Triple-DPU learning on MNIST, EMNIST-Digits and CIFAR-10.}
	\vspace{-0.5em}
	\label{fig:3}
\end{figure*}

Hyper-parameter configurations were systematically determined for each dataset. Specifically, learning rates were selected from the set $\{2 \times 10^{-5}, 4 \times 10^{-5}, 1 \times 10^{-3}, 2 \times 10^{-3}\}$, while weight decay values were searched within $\{7 \times 10^{-4}, 6 \times 10^{-4}, 5 \times 10^{-4}, 2 \times 10^{-3}\}$. For batch size selection, grayscale datasets (MNIST, Fashion-MNIST, EMNIST) adopted 3,000, whereas color datasets (CIFAR-10, SVHN) utilized 256. Model architectures included a 300-unit three-layer perceptron for grayscale datasets and ResNet-34 for color datasets. Systematic evaluation involved three class priors \{0.4, 0.5, 0.6\} and sample quantity variation analysis. Final validation compared URE, ReLU, and ABS algorithms across four loss function configurations.

All the methods are implemented by Pytorch, and conducted the experiments on NVIDIA Geforce RTX 5080.

\subsection{Methods}
The experimental methodology is based on the Empirical Risk Minimization (ERM) framework, implemented by minimizing the proposed unbiased risk function (Eq.\ref{eq1}) and its two rectified risk functions modified via ReLU and ABS correction functions (Eqs.\ref{eq-15} and \ref{eq-16}). These methods are denoted as URE, ReLU, and ABS in the experiments.

\subsection{Baseline methods}
{\bf Siamese:} \cite{Koch2015SiameseNN}
The Siamese architecture represents a neural network paradigm specifically designed for pairwise input comparison, leveraging shared-weight configurations to extract discriminative feature embeddings. Our experimental design employs $M=2$ (pairwise) and $M=3$ (triple) configurations to systematically analyze multi-instance similarity assessment capabilities.

{\bf Contrastive:} \cite{hadsell2006dimensionality}
Contrastive Loss optimizes the embedding space by minimizing intra-class distances and enforcing a margin between inter-class pairs. During evaluation, we randomly select one prototype per class and compute the cosine similarity between test instances and these prototypes. To ensure robustness, the results are averaged over 10 independent prototype selections.

{\bf K-Means:} \cite{macqueen1967some}
As a foundational unsupervised learning paradigm, K-Means clustering partitions data into $K$ clusters via iterative centroid optimization. Aligned with our binary classification objective, we configure $K=2$ for all clustering experiments. To preserve the holistic feature correlations inherent in the training data tuple (pairs/triplets), this study innovatively concatenates feature vectors from three constituent images into extended feature representations. These composite vectors serve as unified input instances for the K-Means algorithm, enabling the preservation of inter-instance spatial relationships while maintaining computational tractability.

\subsection{Experimental Results}
The classification accuracy performance of Pairwise-DPU Learning and Triple-DPU Learning using the Logistic loss function on the seven benchmark datasets---MNIST, Fashion-MNIST, EMNIST-Digits, EMNIST-Letters, EMNIST-Balanced, CIFAR-10, and SVHN---is recorded in Tables \ref{table:2} through \ref{table:5}. Specifically: Tables \ref{table:2} and \ref{table:4} record the classification accuracy performance of Pairwise-DPU Learning and Triple-DPU Learning under the following training sample size settings: $n=10,\!000$ for MNIST, Fashion-MNIST, EMNIST-Digits, CIFAR-10, and SVHN; $n=7,\!000$ for EMNIST-Letters; and $n=4,\!000$ for EMNIST-Balanced. Tables \ref{table:3} and \ref{table:5} record the classification accuracy performance of Pairwise-DPU Learning and Triple-DPU Learning under the following training sample size settings: $n=12,\!000$ for MNIST, Fashion-MNIST, EMNIST-Digits, CIFAR-10, and SVHN; $n=8,\!000$ for EMNIST-Letters; and $n=5,\!000$ for EMNIST-Balanced.

\begin{figure*}
	\centering
	\scriptsize
	\begin{tabular}{ccc}
		\includegraphics[width=5cm]{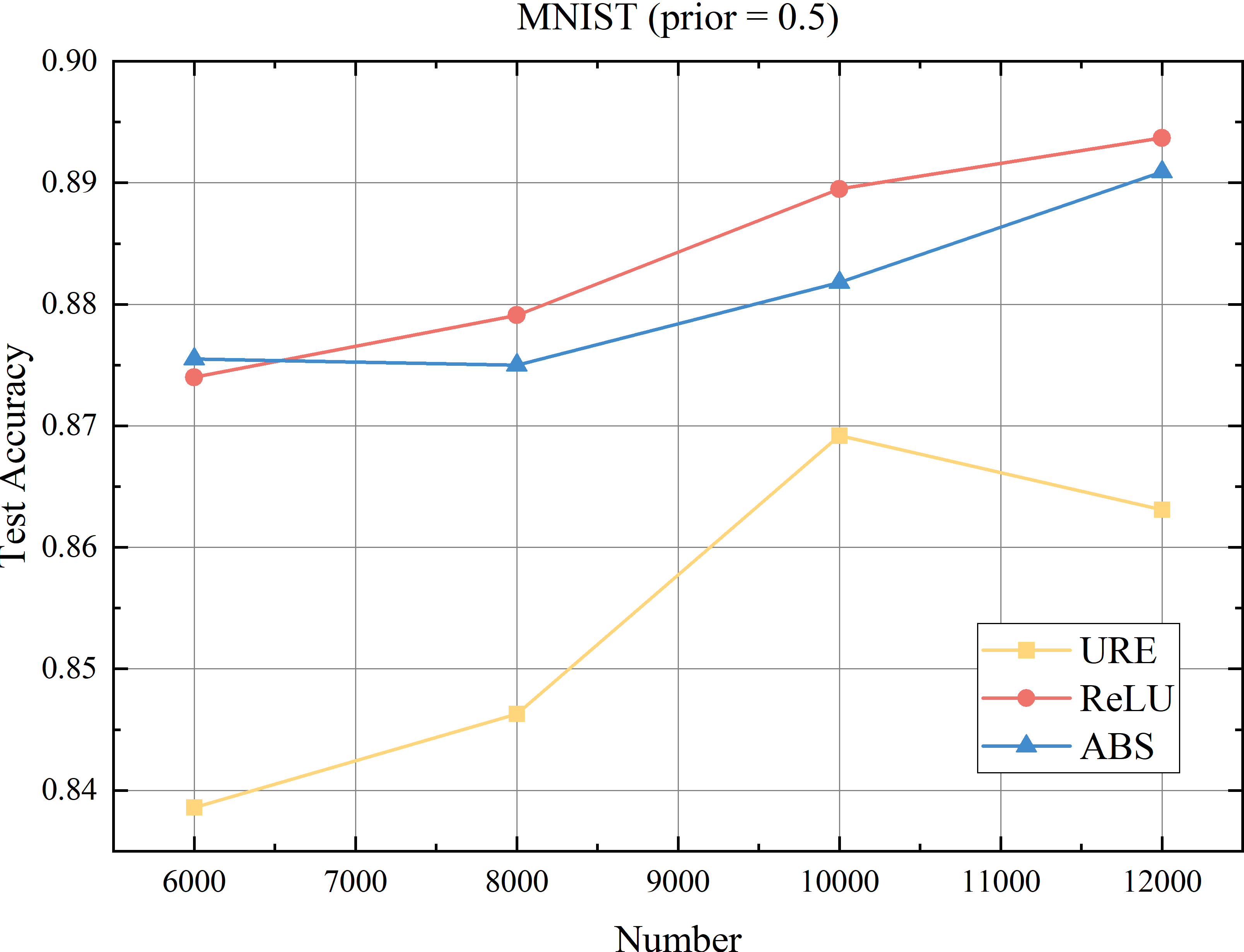} & 
		\includegraphics[width=5cm]{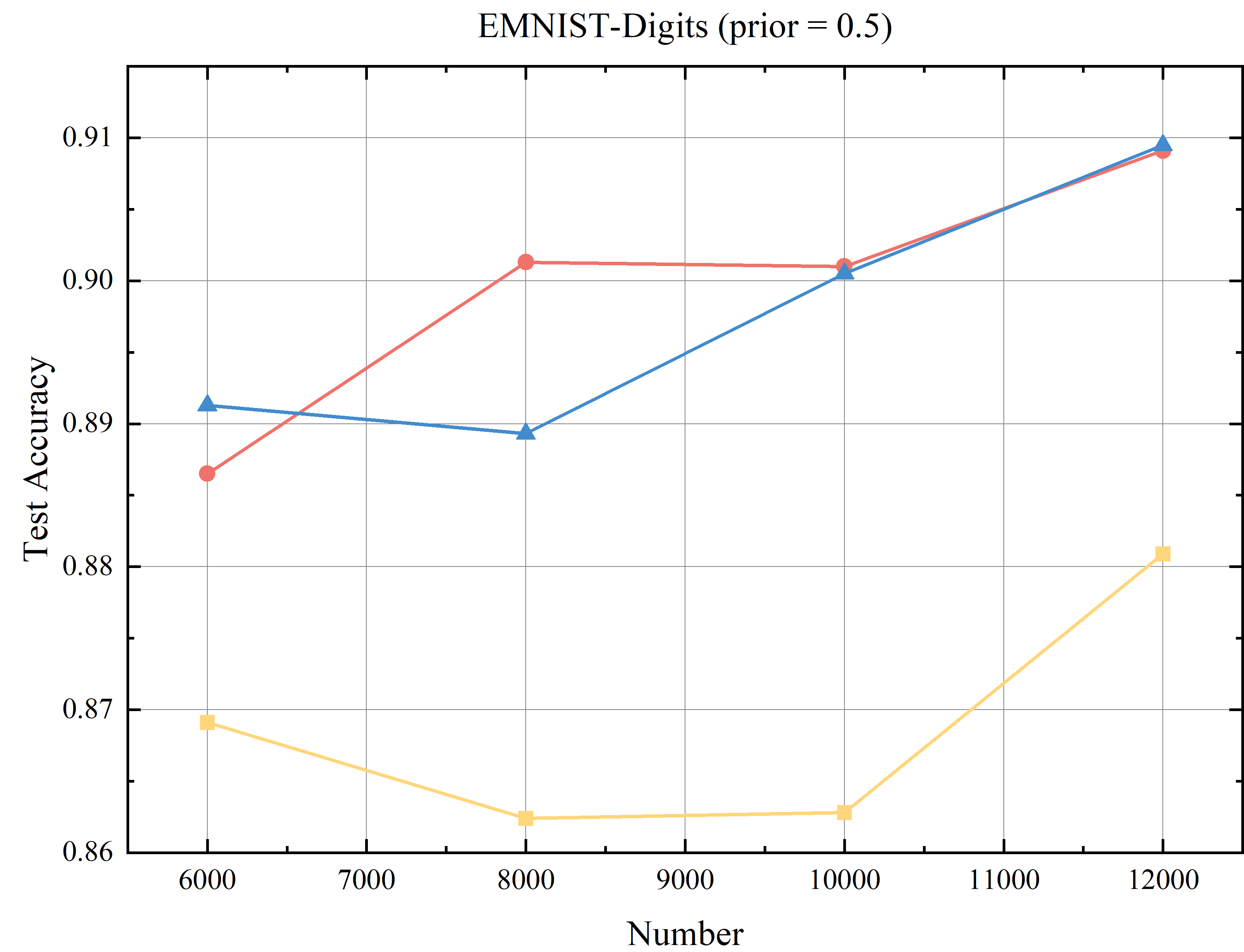} &
		\includegraphics[width=5cm]{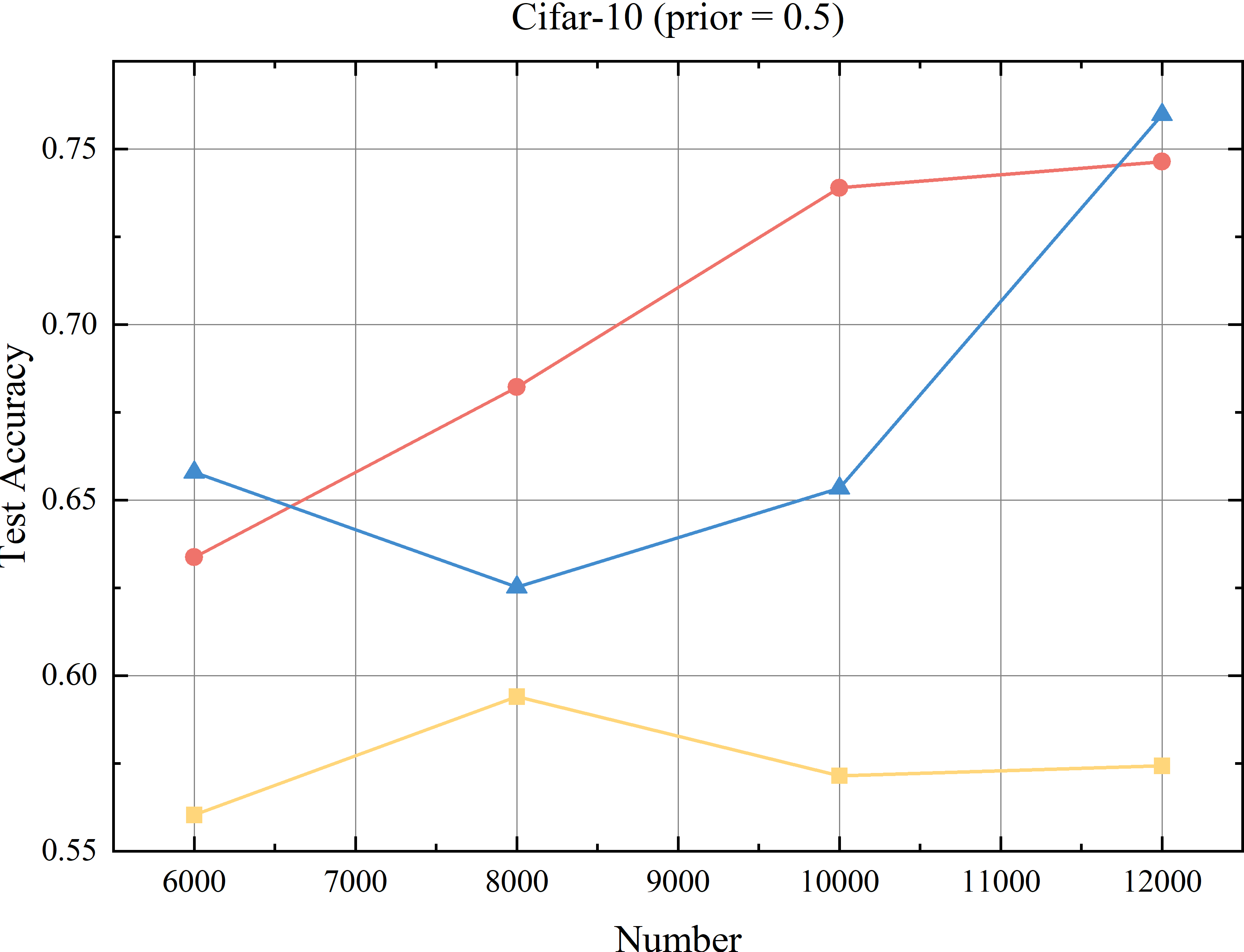} \\
		(a.1) (Pairwise-DPU) MNIST & (a.2) (Pairwise-DPU) EMNIST-Digits & (a.3) (Pairwise-DPU) CIFAR-10 \\ 
		
		\includegraphics[width=5cm]{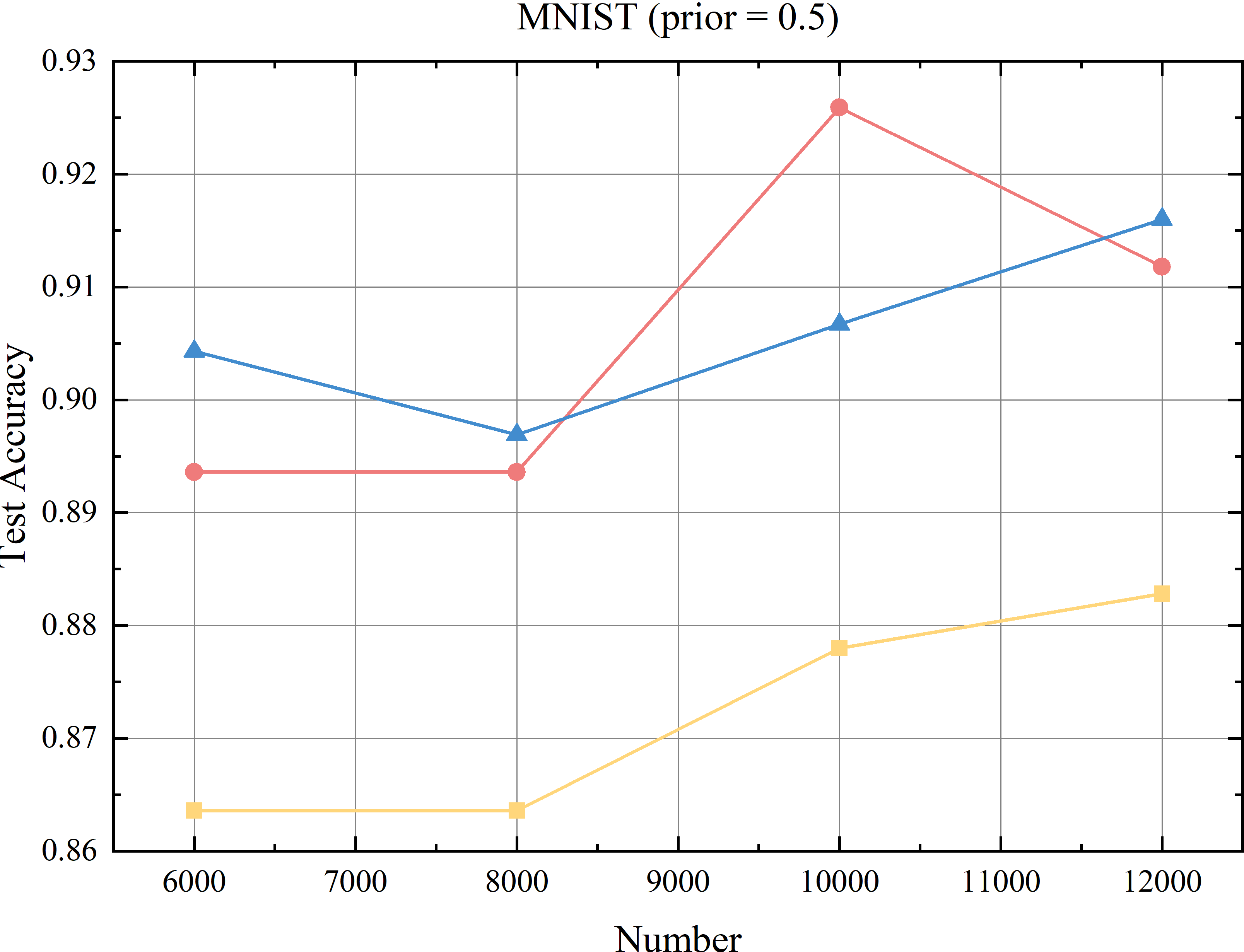} & 
		\includegraphics[width=5cm]{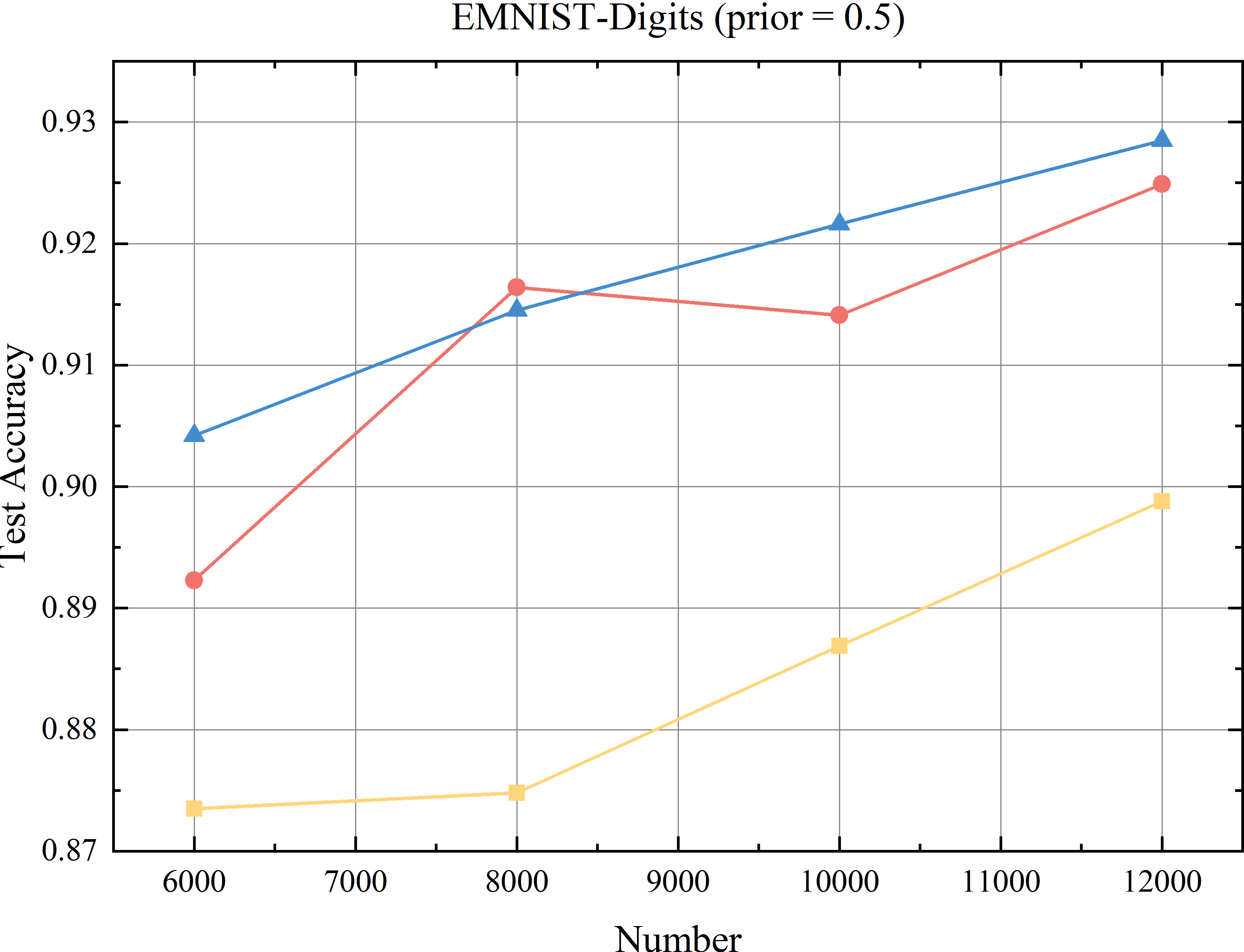} &
		\includegraphics[width=5cm]{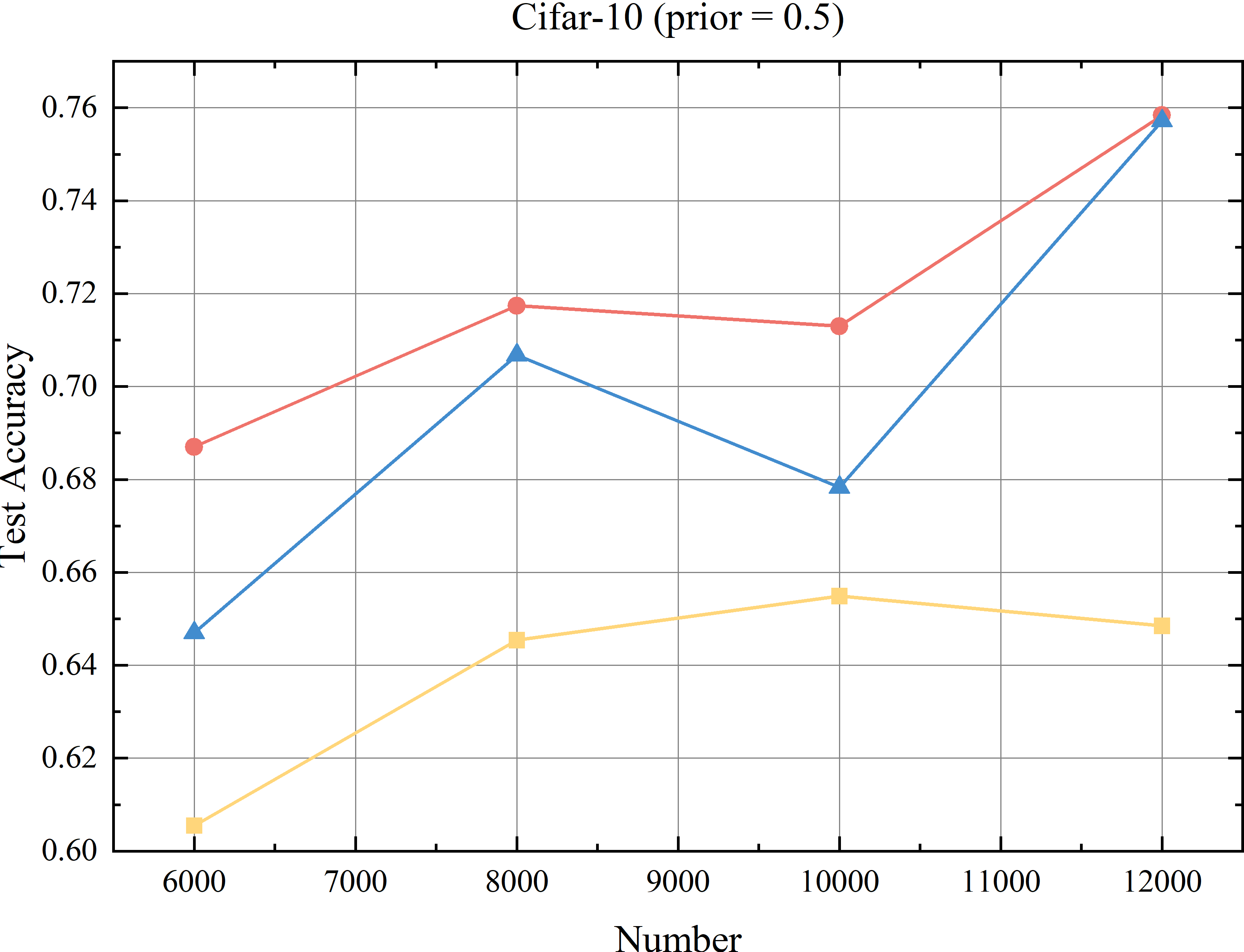} \\
		(b.1) (Triple-DPU) MNIST & (b.2) (Triple-DPU) EMNIST-Digits & (b.3) (Triple-DPU) CIFAR-10  \\
		
	\end{tabular}
	\caption{Test accuracy trends of Pairwise-DPU Learning and Triple-DPU Learning on MNIST, EMNIST-Digits, and CIFAR-10 datasets with varying training sample group sizes.}
	\vspace{-0.5em}
	\label{fig:4}
\end{figure*}

\begin{figure*}[!ht]
	\centering
	\scriptsize
	\begin{tabular}{ccc}
		\includegraphics[width=5cm]{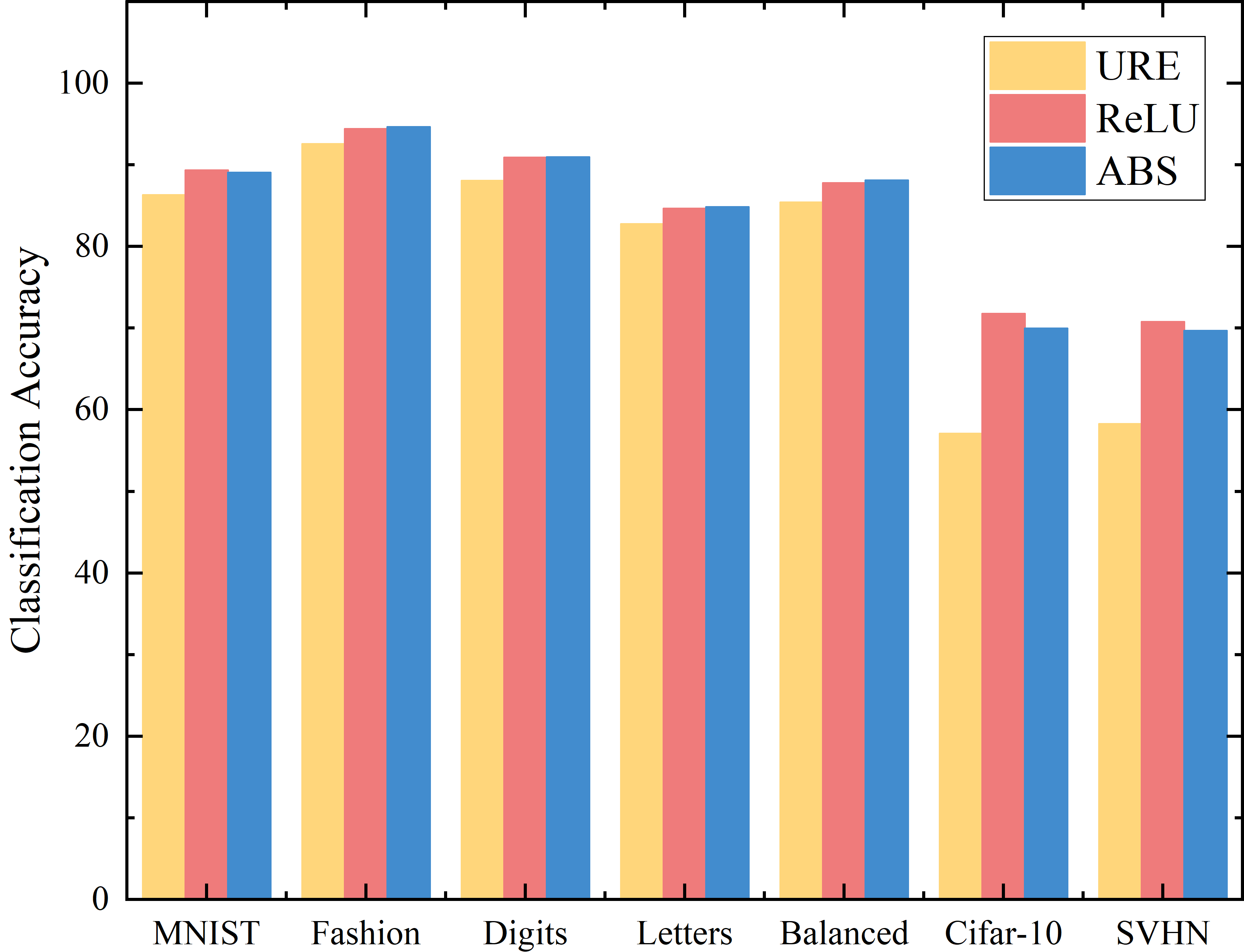} & 
		\includegraphics[width=5cm]{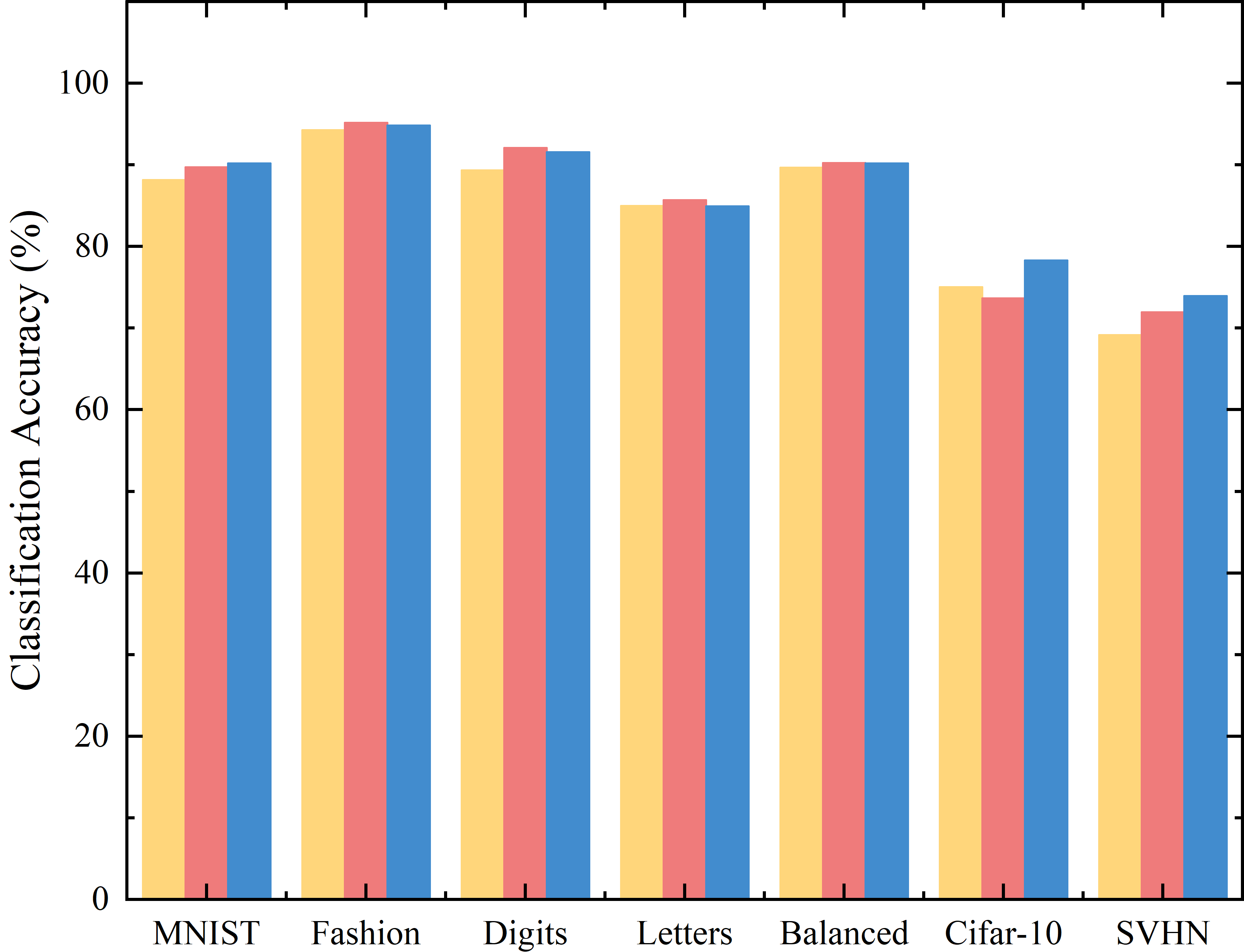} &
		\includegraphics[width=5cm]{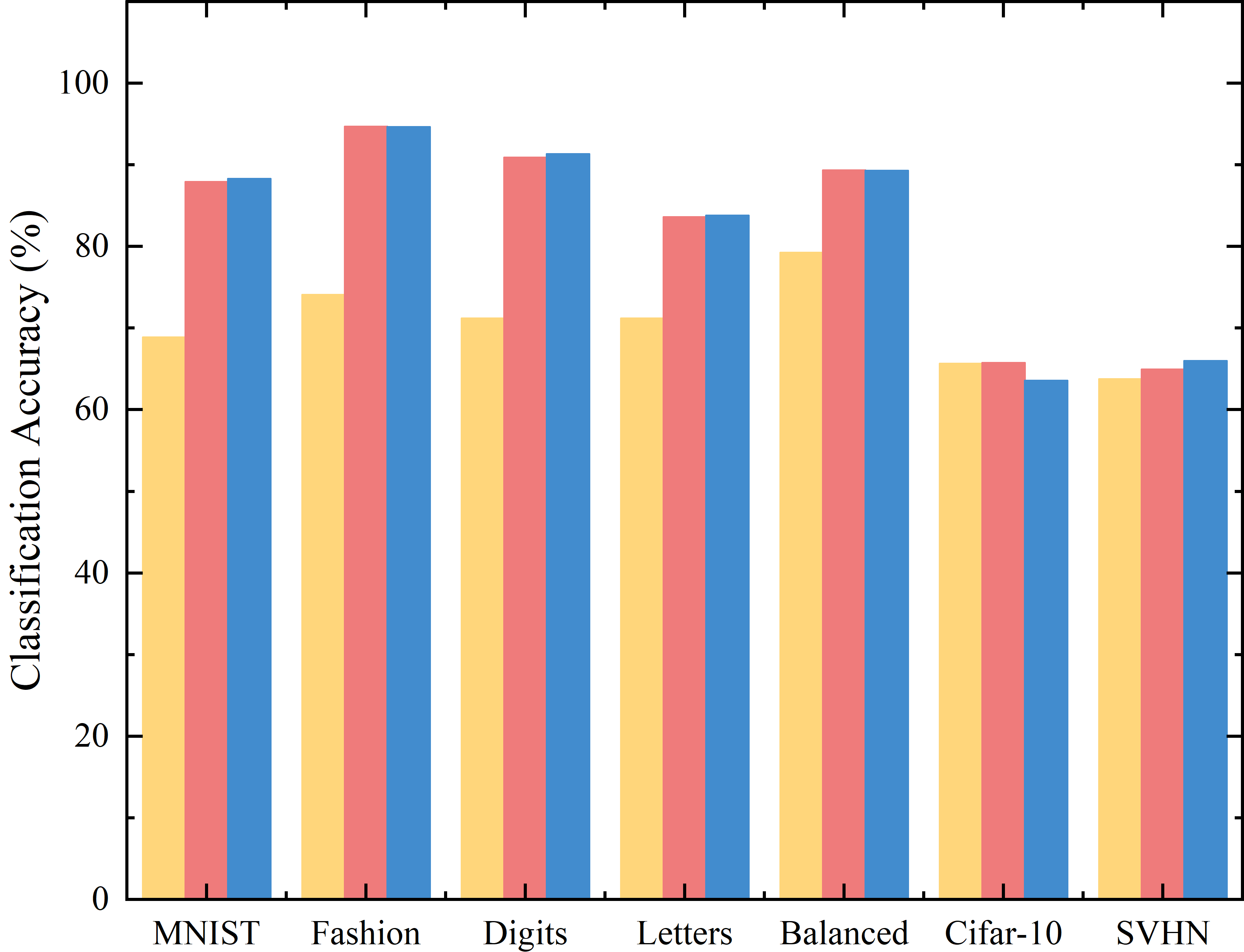} \\
		(a.1) Pairwise-DPU (Logistic loss) & 
		(a.2) Pairwise-DPU (Ramp loss) &
		(a.3) Pairwise-DPU (Squared loss) \\ 
		
		\includegraphics[width=5cm]{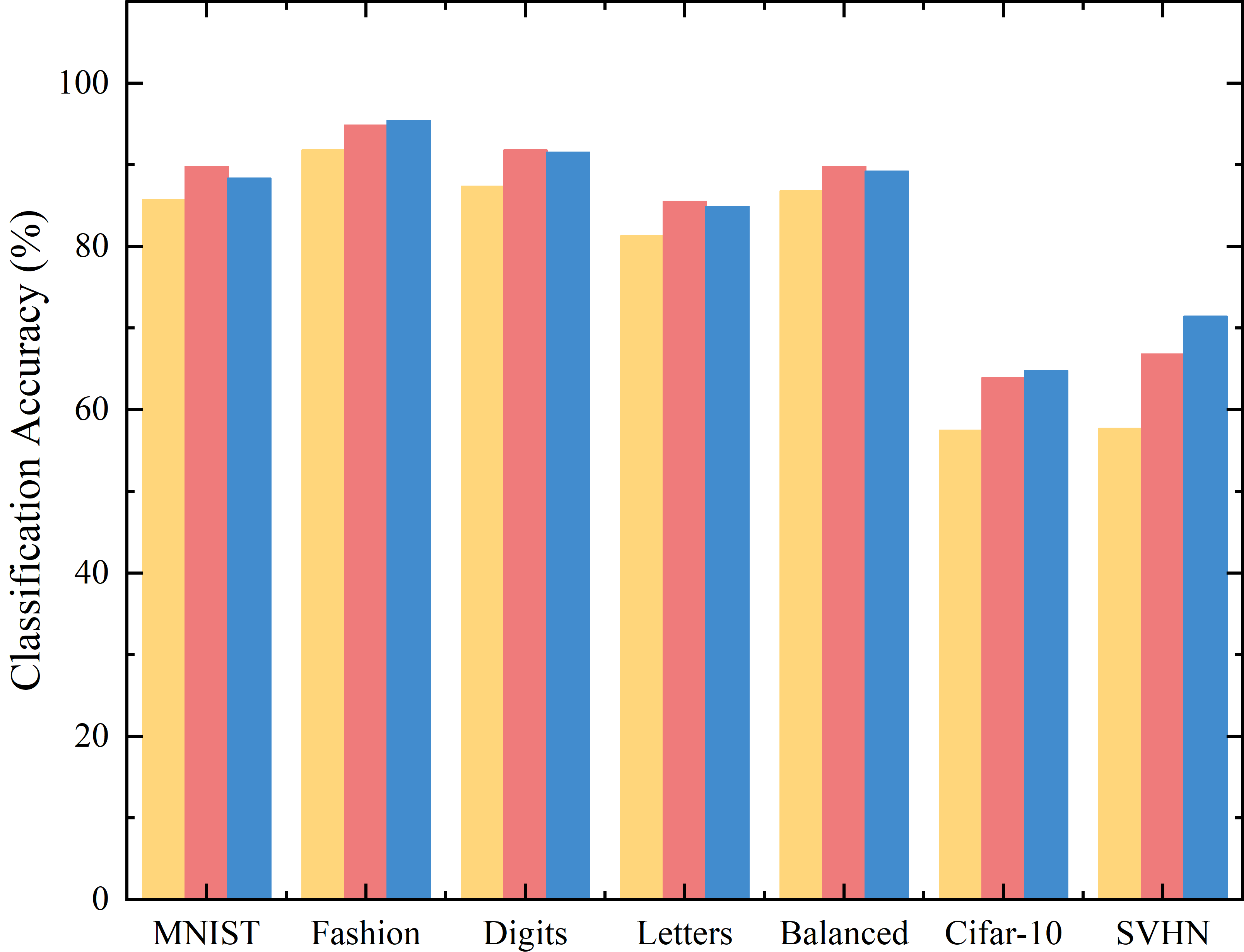} & 
		\includegraphics[width=5cm]{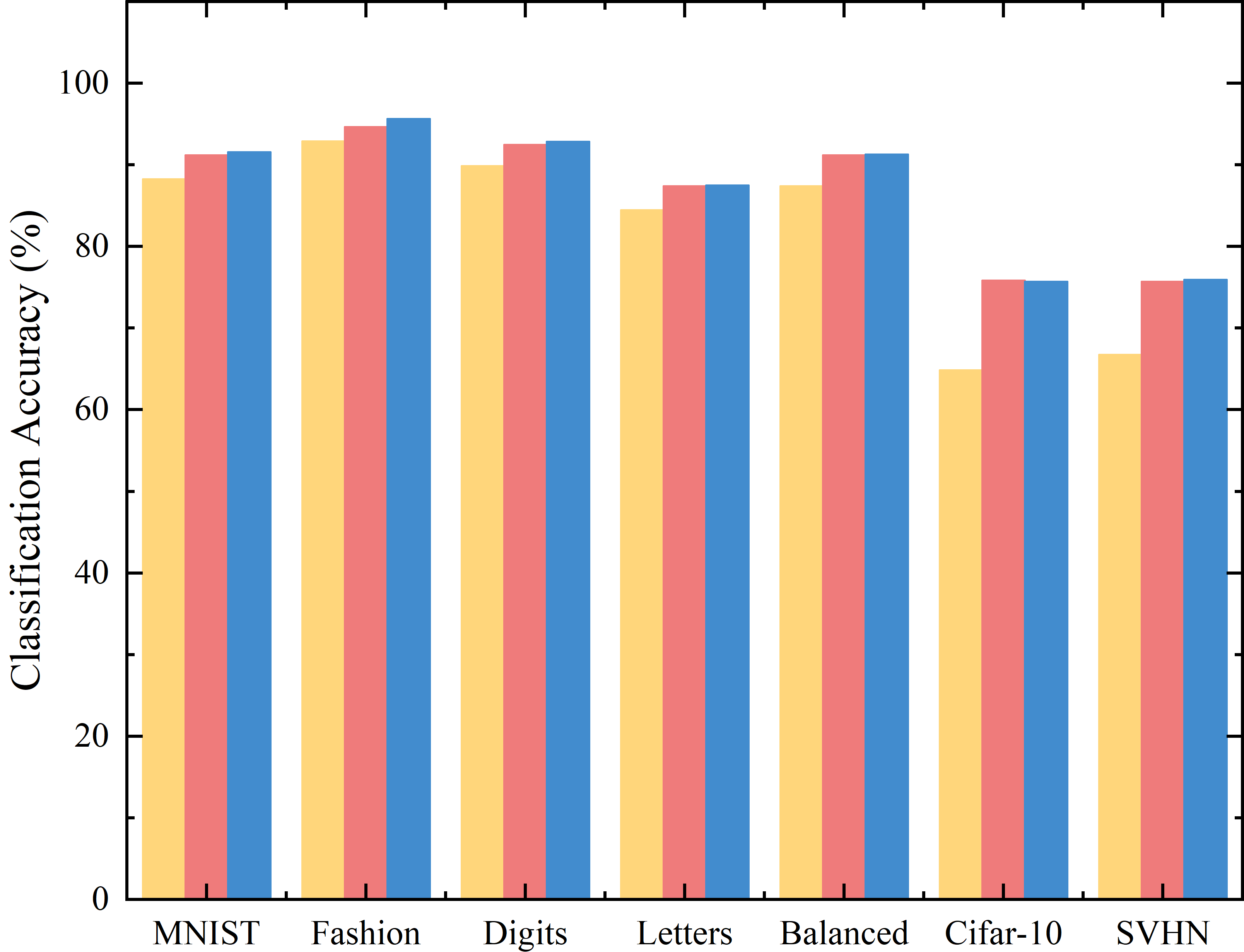} &
		\includegraphics[width=5cm]{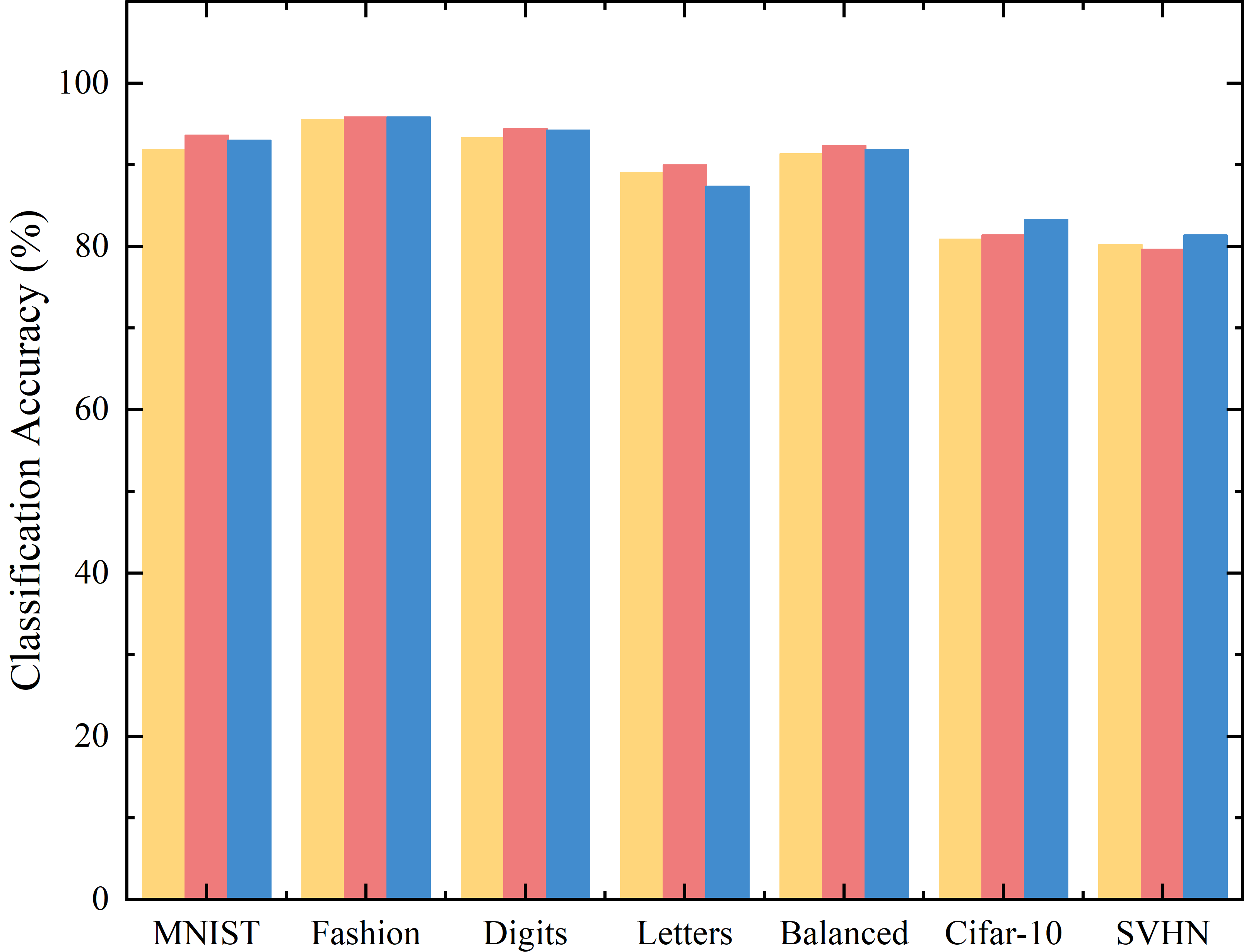} \\
		(b.1) Pairwise-DPU (Hinge loss) & 
		(b.2) Triple-DPU MNIST (Logistic loss) &
		(b.3) Triple-DPU EMNIST-Digits (Ramp loss) \\ 
		
		\multicolumn{3}{c}{%
			\begin{tabular}{cc}
				\includegraphics[width=5cm]{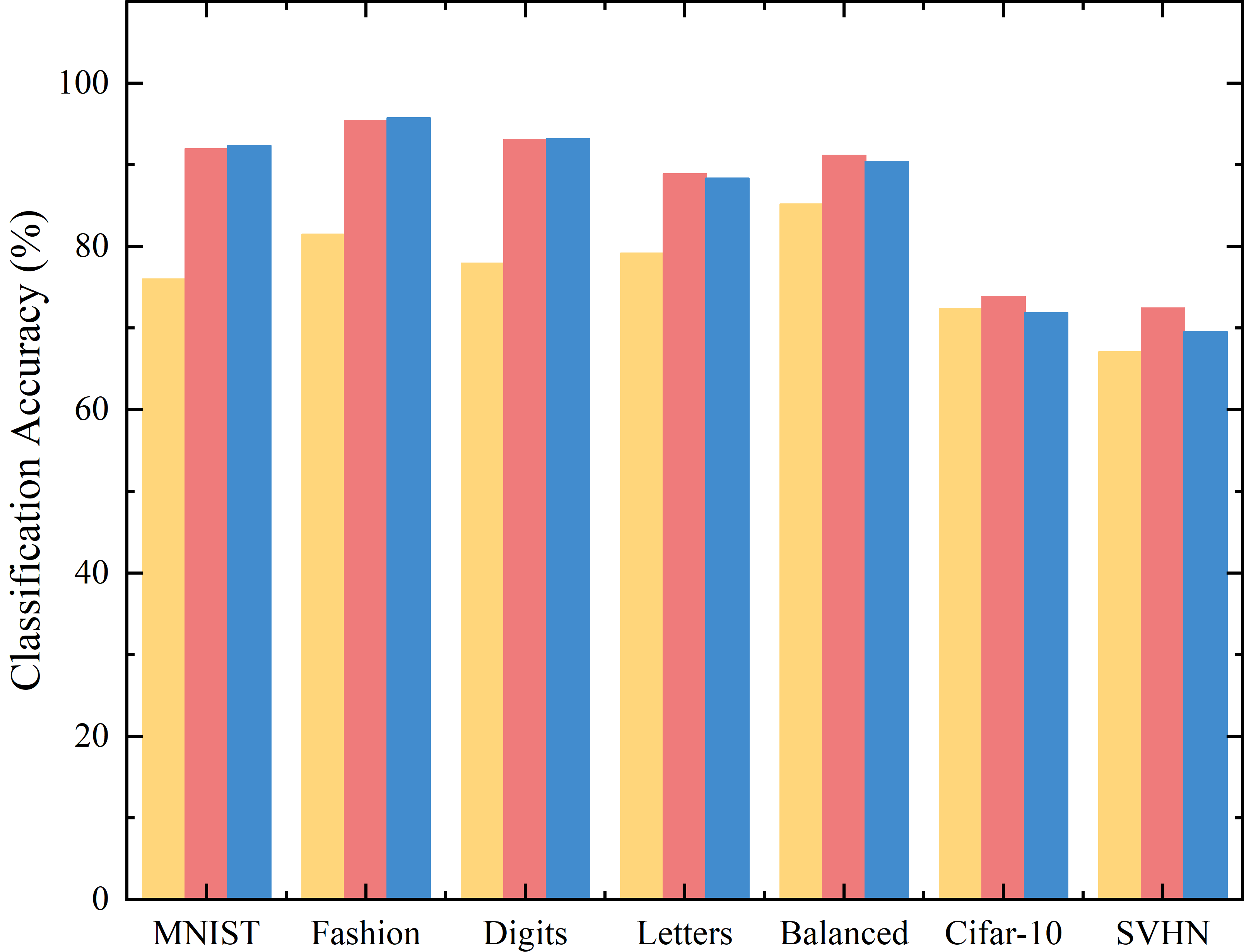} & 
				\includegraphics[width=5cm]{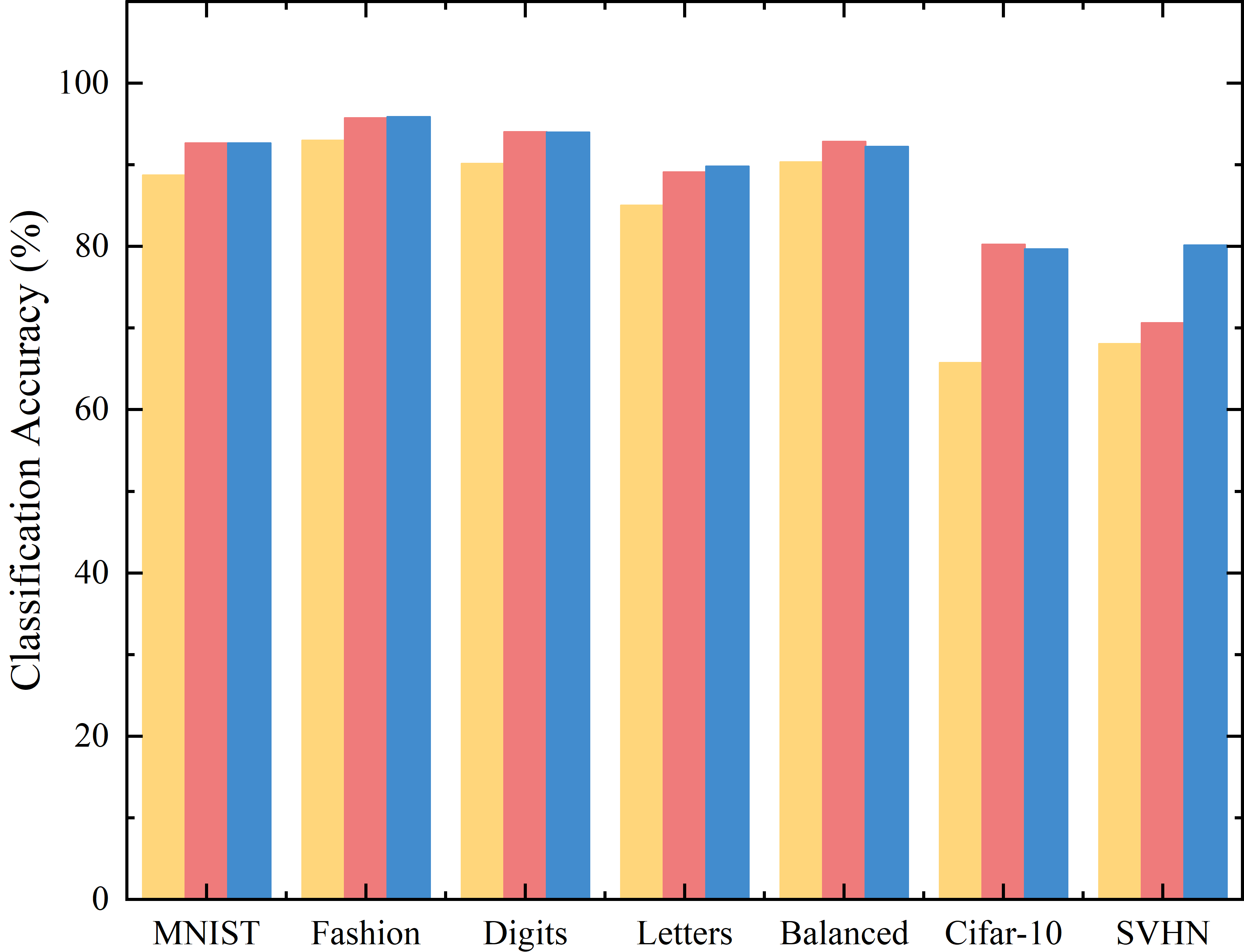} \\
				(c.1) Triple-DPU (Squared loss) & 
				(c.2) Triple-DPU (Hinge loss) \\
			\end{tabular}
		}
	\end{tabular}
	\caption{Comparative classification accuracy of Pairwise-DPU Learning and Triple-DPU Learning using Logistic, Ramp, Squared, and Hinge loss functions across MNIST, Fashion-MNIST, EMNIST-Digits, EMNIST-Letters, EMNIST-Balanced, CIFAR-10, and SVHN datasets.}
	\vspace{-0.5em}
	\label{fig:5}
\end{figure*}

The proposed weakly supervised learning framework, evaluated on MNIST, EMNIST-Digits, and CIFAR-10 under pairwise and triple configurations ($\pi _+=0.5$) (Fig.\ref{fig:3}), demonstrates that the Unbiased Risk Estimator (URE) suffers from training instability due to unbounded negative terms, leading to negative empirical risk and overfitting. In contrast, ReLU and ABS risk estimators effectively stabilize training by enforcing non-negative loss surfaces, maintaining consistent test accuracy across all datasets. Notably, on triple EMNIST-Digits, risk-corrected methods (ReLU and ABS) achieve superior classification accuracy compared to URE, validating the necessity of risk correction for robust generalization. Experimental results demonstrate that risk-corrected methods maintain superior classification performance across diverse datasets and class prior configurations (omitted results due to space constraints).

To evaluate the impact of training sample size on the \emph{MDPU}, experiments were conducted by adjusting the training sample scale. Under $\pi _+=0.5$, classification results of Pairwise-DPU and Triple-DPU algorithms on MNIST, EMNIST-Digits, and CIFAR-10 datasets are presented in Fig.\ref{fig:4}. The horizontal axis indicates training sample size, while the vertical axis corresponds to test accuracy. Results demonstrate that increasing training samples enhances classification performance for both URE, ReLU/ABS methods. Notably, the ReLU and ABS methods exhibit stronger stability during sample expansion: their accuracy growth slopes significantly surpass those of URE, with no performance saturation observed in high-noise scenarios (e.g., CIFAR-10 and SVHN). 

The efficacy of the proposed URE, ReLU, and ABS learning algorithms under varying loss functions was systematically examined. Four representative loss functions in machine learning—Logistic loss, Ramp loss, Squared loss, and Hinge loss—were selected for comparative analysis, with their mathematical definitions and visual characteristics documented in Table \ref{table:1} and Fig.\ref{fig:2}. As Fig.\ref{fig:5} demonstrates, the \emph{MDPU} algorithm shows minimal sensitivity to loss function selection. Crucially, regardless of the loss function employed, the algorithm maintains consistently high classification performance across all benchmark datasets, validating the robustness and broad applicability of the proposed \emph{MDPU} algorithm.

\section{Conclusion}\label{Section:6}
This paper introduces a novel weakly supervised learning framework termed \emph{Learning from M-Tuple Dominant Positive and Unlabeled Data (MDPU)} . By overcoming the precise proportion annotation requirements of traditional \emph{Learning from Label Proportions (LLP)}, \emph{MDPU} proposes an unbiased risk estimator and two corrected risk estimators to mitigate overfitting. Theoretical contributions include rigorous derivations of generalization error bounds for all risk estimators, thereby establishing mathematical foundations for algorithm reliability. Extensive experiments on benchmark datasets demonstrate that \emph{MDPU} achieves superior classification performance compared to existing methods.

\bibliographystyle{ieeetr}
\bibliography{MDPU_classification}

\begin{thebibliography}{10}

\bibitem{zhou_brief_2018}
Z.-H. Zhou, ``A brief introduction to weakly supervised learning,'' {\em
  National Science Review}, vol.~5, pp.~44--53, Jan. 2018.

\bibitem{Oquab_2015_CVPR}
M.~Oquab, L.~Bottou, I.~Laptev, and J.~Sivic, ``Is object localization for
  free? - weakly-supervised learning with convolutional neural networks,'' in
  {\em Proceedings of the IEEE Conference on Computer Vision and Pattern
  Recognition (CVPR)}, June 2015.

\bibitem{7243351}
X.~Xu, W.~Li, D.~Xu, and I.~W. Tsang, ``Co-labeling for multi-view weakly
  labeled learning,'' {\em IEEE Transactions on Pattern Analysis and Machine
  Intelligence}, vol.~38, no.~6, pp.~1113--1125, 2016.

\bibitem{8735810}
Y.-F. Li, L.-Z. Guo, and Z.-H. Zhou, ``Towards safe weakly supervised
  learning,'' {\em IEEE Transactions on Pattern Analysis and Machine
  Intelligence}, vol.~43, no.~1, pp.~334--346, 2021.

\bibitem{9294086}
C.~Gong, J.~Yang, J.~You, and M.~Sugiyama, ``Centroid estimation with
  guaranteed efficiency: A general framework for weakly supervised learning,''
  {\em IEEE Transactions on Pattern Analysis and Machine Intelligence},
  vol.~44, no.~6, pp.~2841--2855, 2022.

\bibitem{10025822}
H.~Wang, Z.~Zhang, Z.~Fan, J.~Chen, L.~Zhang, R.~Shibasaki, and X.~Song,
  ``Multi-task weakly supervised learning for origin–destination travel time
  estimation,'' {\em IEEE Transactions on Knowledge and Data Engineering},
  vol.~35, no.~11, pp.~11628--11641, 2023.

\bibitem{elkan_learning_2008}
C.~Elkan and K.~Noto, ``Learning classifiers from only positive and unlabeled
  data,'' in {\em Proceedings of the 14th {ACM} {SIGKDD} international
  conference on {Knowledge} discovery and data mining}, (Las Vegas Nevada USA),
  pp.~213--220, ACM, Aug. 2008.

\bibitem{4796195}
K.~Zhou, G.-R. Xue, Q.~Yang, and Y.~Yu, ``Learning with positive and unlabeled
  examples using topic-sensitive plsa,'' {\em IEEE Transactions on Knowledge
  and Data Engineering}, vol.~22, no.~1, pp.~46--58, 2010.

\bibitem{zhou2009learning}
K.~Zhou, G.-R. Xue, Q.~Yang, and Y.~Yu, ``Learning with positive and unlabeled
  examples using topic-sensitive plsa,'' {\em IEEE Transactions on Knowledge
  and Data Engineering}, vol.~22, no.~1, pp.~46--58, 2009.

\bibitem{du2015convex}
M.~Du~Plessis, G.~Niu, and M.~Sugiyama, ``Convex formulation for learning from
  positive and unlabeled data,'' in {\em International conference on machine
  learning}, pp.~1386--1394, PMLR, 2015.

\bibitem{kiryo2017positive}
R.~Kiryo, G.~Niu, M.~C. Du~Plessis, and M.~Sugiyama, ``Positive-unlabeled
  learning with non-negative risk estimator,'' {\em Advances in neural
  information processing systems}, vol.~30, 2017.

\bibitem{gong2019large}
C.~Gong, T.~Liu, J.~Yang, and D.~Tao, ``Large-margin label-calibrated support
  vector machines for positive and unlabeled learning,'' {\em IEEE transactions
  on neural networks and learning systems}, vol.~30, no.~11, pp.~3471--3483,
  2019.

\bibitem{sansone2019efficient}
E.~Sansone, F.~G. De~Natale, and Z.-H. Zhou, ``Efficient training for positive
  unlabeled learning,'' {\em IEEE Transactions on Pattern Analysis \& Machine
  Intelligence}, vol.~41, no.~11, pp.~2584--2598, 2019.

\bibitem{bekker2020learning}
J.~Bekker and J.~Davis, ``Learning from positive and unlabeled data: A
  survey,'' {\em Machine Learning}, vol.~109, pp.~719--760, 2020.

\bibitem{jiang2023positive}
Y.~Jiang, Q.~Xu, Y.~Zhao, Z.~Yang, P.~Wen, X.~Cao, and Q.~Huang,
  ``Positive-unlabeled learning with label distribution alignment,'' {\em IEEE
  Transactions on Pattern Analysis and Machine Intelligence}, 2023.

\bibitem{ishida2018binary}
T.~Ishida, G.~Niu, and M.~Sugiyama, ``Binary classification from
  positive-confidence data,'' {\em Advances in neural information processing
  systems}, vol.~31, 2018.

\bibitem{chen2014ambiguously}
Y.-C. Chen, V.~M. Patel, R.~Chellappa, and P.~J. Phillips, ``Ambiguously
  labeled learning using dictionaries,'' {\em IEEE Transactions on Information
  Forensics and Security}, vol.~9, no.~12, pp.~2076--2088, 2014.

\bibitem{patrini2017making}
G.~Patrini, A.~Rozza, A.~Krishna~Menon, R.~Nock, and L.~Qu, ``Making deep
  neural networks robust to label noise: A loss correction approach,'' in {\em
  Proceedings of the IEEE conference on computer vision and pattern
  recognition}, pp.~1944--1952, 2017.

\bibitem{han2018co}
B.~Han, Q.~Yao, X.~Yu, G.~Niu, M.~Xu, W.~Hu, I.~Tsang, and M.~Sugiyama,
  ``Co-teaching: Robust training of deep neural networks with extremely noisy
  labels,'' {\em Advances in neural information processing systems}, vol.~31,
  2018.

\bibitem{9361098}
Z.-Y. Zhang, P.~Zhao, Y.~Jiang, and Z.-H. Zhou, ``Learning from incomplete and
  inaccurate supervision,'' {\em IEEE Transactions on Knowledge and Data
  Engineering}, vol.~34, no.~12, pp.~5854--5868, 2022.

\bibitem{cheng2020weakly}
L.~Cheng, X.~Zhou, L.~Zhao, D.~Li, H.~Shang, Y.~Zheng, P.~Pan, and Y.~Xu,
  ``Weakly supervised learning with side information for noisy labeled
  images,'' in {\em Computer Vision--ECCV 2020: 16th European Conference,
  Glasgow, UK, August 23--28, 2020, Proceedings, Part XXX 16}, pp.~306--321,
  Springer, 2020.

\bibitem{jiang2021learning}
L.~Jiang, H.~Zhang, F.~Tao, and C.~Li, ``Learning from crowds with multiple
  noisy label distribution propagation,'' {\em IEEE Transactions on Neural
  Networks and Learning Systems}, vol.~33, no.~11, pp.~6558--6568, 2021.

\bibitem{wu2022learning}
S.~Wu, T.~Liu, B.~Han, J.~Yu, G.~Niu, and M.~Sugiyama, ``Learning from noisy
  pairwise similarity and unlabeled data,'' {\em The Journal of Machine
  Learning Research}, vol.~23, no.~1, pp.~13851--13884, 2022.

\bibitem{song2022learning}
H.~Song, M.~Kim, D.~Park, Y.~Shin, and J.-G. Lee, ``Learning from noisy labels
  with deep neural networks: A survey,'' {\em IEEE Transactions on neural
  networks and learning systems}, vol.~34, no.~11, pp.~8135--8153, 2022.

\bibitem{10934976}
H.~Chen, Z.~Wang, R.~Tao, H.~Wei, X.~Xie, M.~Sugiyama, B.~Raj, and J.~Wang,
  ``Impact of noisy supervision in foundation model learning,'' {\em IEEE
  Transactions on Pattern Analysis and Machine Intelligence}, pp.~1--19, 2025.

\bibitem{10689264}
W.~Luo, S.~Chen, T.~Liu, B.~Han, G.~Niu, M.~Sugiyama, D.~Tao, and C.~Gong,
  ``Estimating per-class statistics for label noise learning,'' {\em IEEE
  Transactions on Pattern Analysis and Machine Intelligence}, vol.~47, no.~1,
  pp.~305--322, 2025.

\bibitem{lu2018minimal}
N.~Lu, G.~Niu, A.~K. Menon, and M.~Sugiyama, ``On the minimal supervision for
  training any binary classifier from only unlabeled data,'' in {\em
  International Conference on Learning Representations}, 2018.

\bibitem{lu2020mitigating}
N.~Lu, T.~Zhang, G.~Niu, and M.~Sugiyama, ``Mitigating overfitting in
  supervised classification from two unlabeled datasets: A consistent risk
  correction approach,'' in {\em International Conference on Artificial
  Intelligence and Statistics}, pp.~1115--1125, PMLR, 2020.

\bibitem{lu2021binary}
N.~Lu, S.~Lei, G.~Niu, I.~Sato, and M.~Sugiyama, ``Binary classification from
  multiple unlabeled datasets via surrogate set classification,'' in {\em
  International Conference on Machine Learning}, pp.~7134--7144, PMLR, 2021.

\bibitem{wei2024consistent}
Z.~Wei, S.~Shu, Y.~Cao, H.~Wei, B.~An, and L.~Feng, ``Consistent multi-class
  classification from multiple unlabeled datasets,'' in {\em The Twelfth
  International Conference on Learning Representations}, 2024.

\bibitem{tang2023multi}
Y.~Tang, N.~Lu, T.~Zhang, and M.~Sugiyama, ``Multi-class classification from
  multiple unlabeled datasets with partial risk regularization,'' in {\em Asian
  Conference on Machine Learning}, pp.~990--1005, PMLR, 2023.

\bibitem{bao2018classification}
H.~Bao, G.~Niu, and M.~Sugiyama, ``Classification from pairwise similarity and
  unlabeled data,'' in {\em International Conference on Machine Learning},
  pp.~452--461, PMLR, 2018.

\bibitem{shimada2021classification}
T.~Shimada, H.~Bao, I.~Sato, and M.~Sugiyama, ``Classification from pairwise
  similarities/dissimilarities and unlabeled data via empirical risk
  minimization,'' {\em Neural Computation}, vol.~33, no.~5, pp.~1234--1268,
  2021.

\bibitem{cao2021learning}
Y.~Cao, L.~Feng, Y.~Xu, B.~An, G.~Niu, and M.~Sugiyama, ``Learning from
  similarity-confidence data,'' in {\em International Conference on Machine
  Learning}, pp.~1272--1282, PMLR, 2021.

\bibitem{feng2021pointwise}
L.~Feng, S.~Shu, N.~Lu, B.~Han, M.~Xu, G.~Niu, B.~An, and M.~Sugiyama,
  ``Pointwise binary classification with pairwise confidence comparisons,'' in
  {\em International Conference on Machine Learning}, pp.~3252--3262, PMLR,
  2021.

\bibitem{wang2023binary}
W.~Wang, L.~Feng, Y.~Jiang, G.~Niu, M.-L. Zhang, and M.~Sugiyama, ``Binary
  classification with confidence difference,'' in {\em Thirty-seventh
  Conference on Neural Information Processing Systems}, 2023.

\bibitem{quadrianto2009estimating}
N.~Quadrianto, A.~J. Smola, T.~S. Caetano, and Q.~V. Le, ``Estimating labels
  from label proportions,'' {\em Journal of Machine Learning Research},
  vol.~10, pp.~2349--2374, 2009.

\bibitem{yu2014learning}
F.~X. Yu, K.~Choromanski, S.~Kumar, T.~Jebara, and S.-F. Chang, ``On learning
  from label proportions,'' {\em arXiv preprint arXiv:1402.5902}, 2014.

\bibitem{yu2013proptosvm}
F.~Yu, D.~Liu, S.~Kumar, J.~Tony, and S.-F. Chang, ``$\backslash$proptosvm for
  learning with label proportions,'' in {\em International Conference on
  Machine Learning}, pp.~504--512, PMLR, 2013.

\bibitem{scott2020learning}
C.~Scott and J.~Zhang, ``Learning from label proportions: A mutual
  contamination framework,'' {\em Advances in neural information processing
  systems}, vol.~33, pp.~22256--22267, 2020.

\bibitem{mohri2018foundations}
M.~Mohri, A.~Rostamizadeh, and A.~Talwalkar, {\em Foundations of machine
  learning}.
\newblock MIT press, 2018.

\bibitem{mendelson_lower_2008}
S.~Mendelson, ``Lower {Bounds} for the {Empirical} {Minimization}
  {Algorithm},'' {\em IEEE Transactions on Information Theory}, vol.~54,
  pp.~3797--3803, Aug. 2008.

\bibitem{lecun_gradient-based_1998}
Y.~Lecun, L.~Bottou, Y.~Bengio, and P.~Haffner, ``Gradient-based learning
  applied to document recognition,'' {\em Proceedings of the IEEE}, vol.~86,
  pp.~2278--2324, Nov. 1998.

\bibitem{xiao2017fashion}
H.~Xiao, K.~Rasul, and R.~Vollgraf, ``Fashion-mnist: a novel image dataset for
  benchmarking machine learning algorithms,'' {\em arXiv preprint
  arXiv:1708.07747}, 2017.

\bibitem{cohen2017emnist}
G.~Cohen, S.~Afshar, J.~Tapson, and A.~Van~Schaik, ``Emnist: Extending mnist to
  handwritten letters,'' in {\em 2017 international joint conference on neural
  networks (IJCNN)}, pp.~2921--2926, IEEE, 2017.

\bibitem{krizhevsky2009learning}
A.~Krizhevsky, ``Learning multiple layers of features from tiny images,'' {\em
  Master's thesis, University of Tront}, 2009.

\bibitem{37648}
Y.~Netzer, T.~Wang, A.~Coates, A.~Bissacco, B.~Wu, and A.~Y. Ng, ``Reading
  digits in natural images with unsupervised feature learning,'' in {\em NIPS
  Workshop on Deep Learning and Unsupervised Feature Learning 2011}, 2011.

\bibitem{Koch2015SiameseNN}
G.~R. Koch, ``Siamese neural networks for one-shot image recognition,'' 2015.

\bibitem{hadsell2006dimensionality}
R.~Hadsell, S.~Chopra, and Y.~LeCun, ``Dimensionality reduction by learning an
  invariant mapping,'' in {\em 2006 IEEE computer society conference on
  computer vision and pattern recognition (CVPR'06)}, vol.~2, pp.~1735--1742,
  IEEE, 2006.

\bibitem{macqueen1967some}
J.~MacQueen, ``Some methods for classification and analysis of multivariate
  observations,'' in {\em Proceedings of the Fifth Berkeley Symposium on
  Mathematical Statistics and Probability, Volume 1: Statistics}, vol.~5,
  pp.~281--298, University of California press, 1967.

\end{thebibliography}

\setcounter{equation}{0}

\renewcommand\theequation{A.\arabic{equation}}

	\onecolumn
	\section{Appendix}\label{appendix}
	\subsection{Proof of lemma 1}
	\setcounter{lemma}{8}
	\setcounter{theorem}{8}
	\noindent \emph{Proof.}
	The distribution of Pairwise Dominant Positive data can be expressed as follows:
	\begin{equation}
		\begin{aligned}
			\tilde{p}\left( x^1,x^2 \right) &=\frac{p\left( x^1,x^2\mid \left( y^1,y^2 \right) \in \mathcal{Y} _2 \right) *p\left( \left( y^1,y^2 \right) \in \mathcal{Y} _2 \right)}{p\left( \left( y^1,y^2 \right) \in \mathcal{Y} _2 \right)}
			\\&
			=\frac{\sum_{\left( y^1,y^2 \right) \in \mathcal{Y} _2}{p\left( x^1,x^2\mid \left( y^1,y^2 \right) \right) *p\left( y^1,y^2 \right)}}{p\left( \left( y^1,y^2 \right) \in \mathcal{Y} _2 \right)}
			\\&
			=\frac{1}{\pi _{+}^{2}+2\pi _+\pi _-}\left[ \pi _{+}^{2}p_+\left( x^1 \right) p_+\left( x^2 \right) +\pi _+\pi _-p_+\left( x^1 \right) p_-\left( x^2 \right) +\pi _+\pi _-p_-\left( x^1 \right) p_+\left( x^2 \right) \right] 
		\end{aligned}
	\end{equation}
	
	\subsection{Proof of lemma 2}
	\setcounter{equation}{0}
	\renewcommand\theequation{B.\arabic{equation}}
	\noindent \emph{Proof.}
	According to lemma 1, then the marginal distributions $\tilde{p}_{P1}\left( x^1 \right)$, $\tilde{p}_{P2}\left( x^2 \right)$can be computed as:
	
	\begin{equation}
		\begin{aligned}   
			\tilde{p}_{P1}\left( x^1 \right) =\int{\tilde{p}\left( x^1,x^2 \right) dx^2}=\frac{1}{\pi _{+}^{2}+2\pi _+\pi _-}\left[ \left( \pi _{+}^{2}+\pi _+\pi _- \right) p_+\left( x^1 \right) +\pi _+\pi _-p_-\left( x^1 \right) \right] 
		\end{aligned} 
	\end{equation}
	
	\begin{equation}
		\begin{aligned}  
			\tilde{p}_{P2}\left( x^2 \right) =\int{\tilde{p}\left( x^1,x^2 \right) dx^1}=\frac{1}{\pi _{+}^{2}+2\pi _+\pi _-}\left[ \left( \pi _{+}^{2}+\pi _+\pi _- \right) p_+\left( x^2 \right) +\pi _+\pi _-p_-\left( x^2 \right) \right] 
		\end{aligned} 
	\end{equation}
	
	\subsection{Proof of lemma 3}
	\setcounter{lemma}{7}
	\setcounter{theorem}{7}
	\setcounter{equation}{0}
	\renewcommand\theequation{C.\arabic{equation}}
	\noindent \emph{Proof.}
	The distribution of Triple Dominant Positive data can be expressed as follows:
	\begin{equation}
		\begin{aligned}
			\tilde{p}\left( x^1,x^2,x^3 \right) =&\frac{p\left( x^1,x^2,x^3\mid \left( y^1,y^2,y^3 \right) \in \mathcal{Y} _3 \right) *p\left( \left( y^1,y^2,y^3 \right) \in \mathcal{Y} _3 \right)}{p\left( \left( y^1,y^2,y^3 \right) \in \mathcal{Y} _3 \right)}
			\\
			=&\frac{\sum_{\left( y^1,y^2,y^3 \right) \in \mathcal{Y} _3}{p\left( x^1,x^2,x^3\mid \left( y^1,y^2,y^3 \right) \right) *p\left( y^1,y^2,y^3 \right)}}{p\left( \left( y^1,y^2,y^3 \right) \in \mathcal{Y} _3 \right)}
			\\
			=&\frac{1}{\pi _{+}^{3}+3\pi _{+}^{2}\pi _-}\left[ \pi _{+}^{3}p_+\left( x^1 \right) p_+\left( x^2 \right) p_+\left( x^3 \right) +\pi _{+}^{2}\pi _-p_+\left( x^1 \right) p_+\left( x^2 \right) p_-\left( x^3 \right) \right. 
			\\&
			\left. +\pi _{+}^{2}\pi _-p_+\left( x^1 \right) p_-\left( x^2 \right) p_+\left( x^3 \right) +\pi _{+}^{2}\pi _-p_-\left( x^1 \right) p_+\left( x^2 \right) p_+\left( x^3 \right) \right] 
		\end{aligned}
	\end{equation}
	
	\subsection{Proof of lemma 4}
	\setcounter{equation}{0}
	\renewcommand\theequation{D.\arabic{equation}}
	\noindent \emph{Proof.}
	According to lemma 3, then the marginal distributions $\tilde{p}_{T1}\left( x^1 \right)$, $\tilde{p}_{T2}\left( x^2 \right)$, $\tilde{p}_{T3}\left( x^3 \right)$ can be computed as:
	
	\begin{equation}
		\begin{aligned}   
			\tilde{p}_{T1}\left( x^1 \right) =\iint{\tilde{p}\left( x^1,x^2,x^3 \right) dx^2dx^3}=\frac{1}{\pi _{+}^{3}+3\pi _{+}^{2}\pi _-}\left[ \left( \pi _{+}^{3}+2\pi _{+}^{2}\pi _- \right) p_+\left( x^1 \right) +\pi _{+}^{2}\pi _-p_-\left( x^1 \right) \right] 
		\end{aligned} 
	\end{equation}
	
	\begin{equation}
		\begin{aligned}   
			\tilde{p}_{T2}\left( x^2 \right) =\iint{\tilde{p}\left( x^1,x^2,x^3 \right) dx^1dx^3}=\frac{1}{\pi _{+}^{3}+3\pi _{+}^{2}\pi _-}\left[ \left( \pi _{+}^{3}+2\pi _{+}^{2}\pi _- \right) p_+\left( x^2 \right) +\pi _{+}^{2}\pi _-p_-\left( x^2 \right) \right] 
		\end{aligned} 
	\end{equation}
	
	\begin{equation}
		\begin{aligned}   
			\tilde{p}_{T3}\left( x^3 \right) =\iint{\tilde{p}\left( x^1,x^2,x^3 \right) dx^1dx^2}=\frac{1}{\pi _{+}^{3}+3\pi _{+}^{2}\pi _-}\left[ \left( \pi _{+}^{3}+2\pi _{+}^{2}\pi _- \right) p_+\left( x^3 \right) +\pi _{+}^{2}\pi _-p_-\left( x^3 \right) \right] 
		\end{aligned} 
	\end{equation}

	\subsection{Proof of lemma 5}
	\setcounter{equation}{0}
	\renewcommand\theequation{E.\arabic{equation}}
	\noindent \emph{Proof.}
	The distribution of $M$-tuple Dominant Positive and Unlabeled $(MDPU)$ data can be expressed as:
	\begin{equation}
		\begin{aligned} 
			p_{MDP}\left( x^1,x^2,...,x^M \right)=&\frac{p\left( x^1,x^2,\cdots ,x^M\mid \left( y^1,y^2,\cdots ,y^M \right) \in \mathcal{Y} _M \right) *p\left( \left( y^1,y^2,\cdots ,y^M \right) \in \mathcal{Y} _M \right)}{p\left( \left( y^1,y^2,\cdots ,y^M \right) \in \mathcal{Y} _M \right)}
			\\
			=&\frac{\sum_{\left( y^1,y^2,\cdots ,y^M \right) \in \mathcal{Y} _M}{p\left( x^1,x^2,\cdots ,x^M\mid \left( y^1,y^2,\cdots ,y^M \right) \right) *p\left( y^1,y^2,\cdots ,y^M \right)}}{p\left( \left( y^1,y^2,\cdots ,y^M \right) \in \mathcal{Y} _M \right)}
			\\
			=&\frac{\sum_{k=0}^{\lfloor M/2 \rfloor}{\pi _{+}^{M-k}}\pi _{-}^{k}\sum_{S\subseteq \{1,2,...,M\},|S|=k}{\left( \prod_{i\in S}{p_-}(x^i)\prod_{i\notin S}{p_+}(x^i) \right)}}{\sum_{k=0}^{\lfloor M/2 \rfloor}{\binom{M}{k}}\pi _{+}^{M-k}\pi _{-}^{k}}
		\end{aligned} 
	\end{equation}
	
	\subsection{Proof of lemma 6}
	\setcounter{equation}{0}
	\renewcommand\theequation{F.\arabic{equation}}
	\noindent \emph{Proof.}
	The marginal distributions $\hat{p}_{1DP}\left( x^1 \right)$, $\hat{p}_{2DP}\left( x^2 \right)$, $\cdots$, $\hat{p}_{MDP}\left( x^M \right)$ can be computed as:
	\begin{equation}
		\begin{footnotesize}
			\begin{aligned}
				\hat{p}_{1DP}\left( x^1 \right) =&\frac{1}{\sum_{k=0}^{\lfloor M/2 \rfloor}{\binom{M}{k}}\pi _{+}^{M-k}\pi _{-}^{k}}
				\\&
				*\left( \int{\cdots}\int{\pi _{+}^{M}p_+\left( x^1 \right) \cdots p_+\left( x^M \right) dx_2dx_3\cdots dx_M}+\cdots +\int{\cdots}\int{\pi _{+}^{M-1}\pi _-p_+\left( x^1 \right) \cdots p_-\left( x^M \right) dx_2dx_3\cdots dx_M} \right) 
				\\
				=&\frac{1}{\sum_{k=0}^{\lfloor M/2 \rfloor}{\left( \begin{array}{c}
							M\\
							k\\
						\end{array} \right) \pi _{+}^{M-k}\pi _{-}^{k}}}\left[ \left( \sum_{k=0}^{\lfloor M/2 \rfloor}{\left( \begin{array}{c}
						M-1\\
						k\\
					\end{array} \right) \pi _{+}^{M-k}\pi _{-}^{k}} \right) p_+\left( x^1 \right) +\left( \sum_{k=1}^{\lfloor M/2 \rfloor}{\left( \begin{array}{c}
						M-1\\
						k-1\\
					\end{array} \right) \pi _{+}^{M-k}\pi _{-}^{k}} \right) p_-\left( x^1 \right) \right] 
			\end{aligned} 
		\end{footnotesize}
	\end{equation}
	
	\begin{equation}
		\begin{footnotesize}
			\begin{aligned}
				\hat{p}_{2DP}\left( x^2 \right) =&\frac{1}{\sum_{k=0}^{\lfloor M/2 \rfloor}{\binom{M}{k}}\pi _{+}^{M-k}\pi _{-}^{k}}
				\\&
				*\left( \int{\cdots}\int{\pi _{+}^{M}p_+\left( x^1 \right) \cdots p_+\left( x^M \right) dx_1dx_3\cdots dx_M}+\cdots +\int{\cdots}\int{\pi _{+}^{M-1}\pi _-p_+\left( x^1 \right) \cdots p_-\left( x^M \right) dx_1dx_3\cdots dx_M} \right) 
				\\
				=&\frac{1}{\sum_{k=0}^{\lfloor M/2 \rfloor}{\left( \begin{array}{c}
							M\\
							k\\
						\end{array} \right) \pi _{+}^{M-k}\pi _{-}^{k}}}\left[ \left( \sum_{k=0}^{\lfloor M/2 \rfloor}{\left( \begin{array}{c}
						M-1\\
						k\\
					\end{array} \right) \pi _{+}^{M-k}\pi _{-}^{k}} \right) p_+\left( x^2 \right) +\left( \sum_{k=1}^{\lfloor M/2 \rfloor}{\left( \begin{array}{c}
						M-1\\
						k-1\\
					\end{array} \right) \pi _{+}^{M-k}\pi _{-}^{k}} \right) p_-\left( x^2 \right) \right] 
			\end{aligned} 
		\end{footnotesize}
	\end{equation}
	$$
	\vdots 
	$$
	\begin{equation}
		\begin{footnotesize}
			\begin{aligned}
				\hat{p}_{MDP}\left( x^M \right) =&\frac{1}{\sum_{k=0}^{\lfloor M/2 \rfloor}{\binom{M}{k}}\pi _{+}^{M-k}\pi _{-}^{k}}
				\\&
				*\left( \int{\cdots}\int{\pi _{+}^{M}p_+\left( x^1 \right) \cdots p_+\left( x^M \right) dx_1dx_2\cdots dx_{M-1}}+\cdots +\int{\cdots}\int{\pi _{+}^{M-1}\pi _-p_+\left( x^1 \right) \cdots p_-\left( x^M \right) dx_1dx_2\cdots dx_{M-1}} \right) 
				\\
				=&\frac{1}{\sum_{k=0}^{\lfloor M/2 \rfloor}{\left( \begin{array}{c}
							M\\
							k\\
						\end{array} \right) \pi _{+}^{M-k}\pi _{-}^{k}}}\left[ \left( \sum_{k=0}^{\lfloor M/2 \rfloor}{\left( \begin{array}{c}
						M-1\\
						k\\
					\end{array} \right) \pi _{+}^{M-k}\pi _{-}^{k}} \right) p_+\left( x^M \right) +\left( \sum_{k=1}^{\lfloor M/2 \rfloor}{\left( \begin{array}{c}
						M-1\\
						k-1\\
					\end{array} \right) \pi _{+}^{M-k}\pi _{-}^{k}} \right) p_-\left( x^M \right) \right] 
			\end{aligned} 
		\end{footnotesize}
	\end{equation}
	
	\subsection{Proof of lemma 7}
	\setcounter{equation}{0}
	\renewcommand\theequation{G.\arabic{equation}}
	\noindent \emph{Proof.}
	According to Lemma 6, $\hat{p}_{MDP}\left( x \right)$ can be expressed as:
	
	\begin{equation}
		\begin{footnotesize}
			\begin{aligned}
				\hat{p}_{MDP}\left( x \right) =\frac{1}{\sum_{k=0}^{\lfloor M/2 \rfloor}{\left( \begin{array}{c}
							M\\
							k\\
						\end{array} \right) \pi _{+}^{M-k}\pi _{-}^{k}}}\left[ \left( \sum_{k=0}^{\lfloor M/2 \rfloor}{\left( \begin{array}{c}
						M-1\\
						k\\
					\end{array} \right) \pi _{+}^{M-k}\pi _{-}^{k}} \right) p_+\left( x \right) +\left( \sum_{k=1}^{\lfloor M/2 \rfloor}{\left( \begin{array}{c}
						M-1\\
						k-1\\
					\end{array} \right) \pi _{+}^{M-k}\pi _{-}^{k}} \right) p_-\left( x \right) \right] 
			\end{aligned} 
		\end{footnotesize}
	\end{equation}
	
	and the distribution of Unlabeled data can be expressed as follows:
	\begin{equation}
		\begin{aligned}
			p_U\left( x \right) =\pi _+p_+\left( x \right) +\pi _-p_-\left( x \right) 
		\end{aligned} 
	\end{equation}
	
	Hence, the following system of linear equations is obtained:
	\begin{equation}\label{g-3}
		\begin{aligned}
			\left[ \begin{array}{c}
				\hat{p}_{MDP}\left( x \right)\\
				p_U\left( x \right)\\
			\end{array}\mathrm{} \right] =\left[ \begin{matrix}
				\frac{\sum_{k=0}^{\lfloor M/2 \rfloor}{\left( \begin{array}{c}
							M-1\\
							k\\
						\end{array} \right) \pi _{+}^{M-k}\pi _{-}^{k}}}{\sum_{k=0}^{\lfloor M/2 \rfloor}{\left( \begin{array}{c}
							M\\
							k\\
						\end{array} \right) \pi _{+}^{M-k}\pi _{-}^{k}}}&		\frac{\sum_{k=1}^{\lfloor M/2 \rfloor}{\left( \begin{array}{c}
							M-1\\
							k-1\\
						\end{array} \right) \pi _{+}^{M-k}\pi _{-}^{k}}}{\sum_{k=0}^{\lfloor M/2 \rfloor}{\left( \begin{array}{c}
							M\\
							k\\
						\end{array} \right) \pi _{+}^{M-k}\pi _{-}^{k}}}\\
				\pi _+&		\pi _-\\
			\end{matrix} \right] \left[ \begin{array}{c}
				p_+\left( x \right)\\
				p_-\left( x \right)\\
			\end{array}\mathrm{} \right] 
		\end{aligned} 
	\end{equation}
	
	To improve the clarity of the proof process, $a$ and $b$ are respectively used as simplified representations of the coefficients for the positive and negative sample distributions in distribution $\hat{p}_{MDP}$. The expressions for $a$ and $b$ are as follows:
	\begin{equation}
		\begin{aligned}
			a=\frac{\sum_{k=0}^{\lfloor M/2 \rfloor}{\left( \begin{array}{c}
						M-1\\
						k\\
					\end{array} \right) \pi _{+}^{M-k}\pi _{-}^{k}}}{\sum_{k=0}^{\lfloor M/2 \rfloor}{\left( \begin{array}{c}
						M\\
						k\\
					\end{array} \right) \pi _{+}^{M-k}\pi _{-}^{k}}},b=\frac{\sum_{k=1}^{\lfloor M/2 \rfloor}{\left( \begin{array}{c}
						M-1\\
						k-1\\
					\end{array} \right) \pi _{+}^{M-k}\pi _{-}^{k}}}{\sum_{k=0}^{\lfloor M/2 \rfloor}{\left( \begin{array}{c}
						M\\
						k\\
					\end{array} \right) \pi _{+}^{M-k}\pi _{-}^{k}}}
		\end{aligned} 
	\end{equation}
	
	By solving the above system of linear equations (Eq.(\ref{g-3})), we can obtain the following results:
	\begin{equation}
		\begin{aligned}
			\left[ \begin{array}{c}
				p_+(x)\\
				p_-(x)\\
			\end{array} \right] =\left[ \begin{matrix}
				\frac{\pi _-}{a\pi _--b\pi _+}&		\frac{-b}{a\pi _--b\pi _+}\\
				\frac{-\pi _+}{a\pi _--b\pi _+}&		\frac{a}{a\pi _--b\pi _+}\\
			\end{matrix} \right] \left[ \begin{array}{c}
				\hat{p}_{MDP}\left( x \right)\\
				p_U\left( x \right)\\
			\end{array}\mathrm{} \right] 
		\end{aligned} 
	\end{equation}
	
	Then we have the distributions of positive and negative samples:
	\begin{equation}
		p_+(x)=\frac{\pi _-}{a\pi _--b\pi _+}\hat{p}_{MDP}\left( x \right) -\frac{b}{a\pi _--b\pi _+}p_U\left( x \right) 
	\end{equation}
	\begin{equation}
		p_-(x)=\frac{-\pi _+}{a\pi _--b\pi _+}\hat{p}_{MDP}\left( x \right) +\frac{a}{a\pi _--b\pi _+}p_U\left( x \right) 
	\end{equation}

	\subsection{Proof of Theorem 1}
	\setcounter{equation}{0}
	\renewcommand\theequation{H.\arabic{equation}}
	\noindent \emph{Proof.}
	From the distributions of positive and negative samples, we can directly derive the classification risk for \emph{Learning from M-Tuple Dominant Positive and Unlabeled Data (MDPU)}:
	\begin{equation}
		\begin{aligned} 
			R_{MDPU}\left( g \right) =&\underset{x\sim p\left( x,y \right)}{\mathbb{E}}\left[ \ell \left( g\left( x \right) ,y \right) \right] 
			\\
			=&\pi _+\underset{x\sim p_+\left( x \right)}{\mathbb{E}}\left[ \ell \left( g\left( x \right) ,+1 \right) \right] +\pi _-\underset{x\sim p_-\left( x \right)}{\mathbb{E}}\left[ \ell \left( g\left( x \right) ,-1 \right) \right] 
			\\
			=&\pi _+\int{\left( \frac{\pi _-}{a\pi _--b\pi _+}\hat{p}_{MDP}\left( x \right) -\frac{b}{a\pi _--b\pi _+}p_U\left( x \right) \right) \ell \left( g\left( x \right) ,+1 \right)}dx
			\\&
			+\pi _-\int{\left( \frac{-\pi _+}{a\pi _--b\pi _+}\hat{p}_{MDP}\left( x \right) +\frac{a}{a\pi _--b\pi _+}p_U\left( x \right) \right) \ell \left( g\left( x \right) ,-1 \right)}dx
			\\
			=&\underset{x\sim \hat{p}_{MDP}}{\mathbb{E}}\left[ \frac{\pi _+\pi _-}{a\pi _--b\pi _+}\ell \left( g\left( x \right) ,+1 \right) -\frac{\pi _+\pi _-}{a\pi _--b\pi _+}\ell \left( g\left( x \right) ,-1 \right) \right] 
			\\&
			+\underset{x\sim p_U\left( x \right)}{\mathbb{E}}\left[ \frac{-b\pi _+}{a\pi _--b\pi _+}\ell \left( g\left( x \right) ,+1 \right) +\frac{a\pi _-}{a\pi _--b\pi _+}\ell \left( g\left( x \right) ,-1 \right) \right] 
			\\
			=&\pi _+\pi _-\underset{x\sim \hat{p}_{MDP}}{\mathbb{E}}\left[ \frac{\ell \left( g\left( x \right) ,+1 \right) -\ell \left( g\left( x \right) ,-1 \right)}{a\pi _--b\pi _+} \right] 
			\\&
			+\underset{x\sim p_U\left( x \right)}{\mathbb{E}}\left[ \frac{-b\pi _+}{a\pi _--b\pi _+}\ell \left( g\left( x \right) ,+1 \right) +\frac{a\pi _-}{a\pi _--b\pi _+}\ell \left( g\left( x \right) ,-1 \right) \right] 
			\\
			=&\pi _+\pi _-\underset{\left( \boldsymbol{x}^1,\boldsymbol{x}^2,\cdots ,\boldsymbol{x}^{\boldsymbol{M}} \right) \sim p_{MDP}}{\mathbb{E}}\left[ \frac{\ell \left( g\left( x^1 \right) ,+1 \right) -\ell \left( g\left( x^1 \right) ,-1 \right) +\cdots +\ell \left( g\left( x^M \right) ,+1 \right) -\ell \left( g\left( x^M \right) ,-1 \right)}{M} \right] \\&+\underset{x\sim p_U\left( x \right)}{\mathbb{E}}\left[ \mathcal{L} _U\left( g\left( x \right) \right) \right] 
			\\
			=&\pi _+\pi _-\underset{\left( \boldsymbol{x}^1,\boldsymbol{x}^2,\cdots ,\boldsymbol{x}^{\boldsymbol{M}} \right) \sim p_{MDP}\left( \boldsymbol{x}^1,\boldsymbol{x}^2,\cdots ,\boldsymbol{x}^{\boldsymbol{M}} \right)}{\mathbb{E}}\left[ \frac{\mathcal{L} _{MDP}\left( g\left( x^1 \right) \right) +\cdots +\mathcal{L} _{MDP}\left( g\left( x^M \right) \right)}{M} \right] +\underset{x\sim p_U\left( x \right)}{\mathbb{E}}\left[ \mathcal{L} _U\left( g\left( x \right) \right) \right] 
			\\
			=&\pi _+\pi _-\underset{\left( \boldsymbol{x}^1,\boldsymbol{x}^2,\cdots ,\boldsymbol{x}^{\boldsymbol{M}} \right) \sim p_{MDP}\left( \boldsymbol{x}^1,\boldsymbol{x}^2,\cdots ,\boldsymbol{x}^{\boldsymbol{M}} \right)}{\mathbb{E}}\left[ \tilde{\mathcal{L}}_{MDP}\left( g\left( x \right) \right) \right] +\underset{x\sim p_U\left( x \right)}{\mathbb{E}}\left[ \mathcal{L} _U\left( g\left( x \right) \right) \right] 
		\end{aligned} 
	\end{equation}
	
	where,
	
	\begin{equation}
		\tilde{\mathcal{L}}_{MDP}\left( g\left( x \right) \right) =\frac{\mathcal{L} _{MDP}\left( g\left( x^1 \right) \right) +\cdots +\mathcal{L} _{MDP}\left( g\left( x^M \right) \right)}{M}
	\end{equation}
	
	\begin{equation}
		\begin{aligned} 
			\mathcal{L} _{MDP}\left( g\left( x^i \right) \right) \triangleq \frac{1}{a\pi _--b\pi _+}\tilde{\ell}\left( g\left( x^i \right) \right) 
		\end{aligned} 
	\end{equation}
	
	\begin{equation}
		\begin{aligned} 
			\tilde{\ell}\left( g\left( x^i \right) \right) \triangleq \ell \left( g\left( x^i \right) ,+1 \right) -\ell \left( g\left( x^i \right) ,-1 \right) 
		\end{aligned} 
	\end{equation}
	
	\begin{equation}
		\begin{aligned} 
			\mathcal{L} _U\left( g\left( x \right) \right) =\frac{-b\pi _+}{a\pi _--b\pi _+}\ell \left( g\left( x \right) ,+1 \right) +\frac{a\pi _-}{a\pi _--b\pi _+}\ell \left( g\left( x \right) ,-1 \right) 
		\end{aligned} 
	\end{equation}
	
	\begin{equation}
		\begin{aligned} 
			\hat{R}_{MDPU}\left( g \right) =\frac{\pi _+\pi _-}{Mn_{MDP}}\sum_{i=1}^{Mn_{MDP}}{\mathcal{L} _{MDP}\left( g\left( \tilde{x}_{MDP,i} \right) \right)}+\frac{1}{n_U}\sum_{j=1}^{n_U}{\left[ \mathcal{L} _U\left( g\left( x_{U,i} \right) \right) \right]}
		\end{aligned} 
	\end{equation}

	\subsection{Proof of Theorem 2}
	\setcounter{equation}{0}
	\renewcommand\theequation{I.\arabic{equation}}
	\noindent \emph{Proof.}
	Firstly, we make the following definitions,
	$$
	\begin{cases}
		R_{\widehat{MDP}}(g)\triangleq \underset{x\sim \hat{p}_{MDP}\left( x \right)}{\mathbb{E}}\left[ \mathcal{L} _{MDP}\left( g\left( x \right) \right) \right]\\
		\hat{R}_{\widehat{MDP}}(g)\triangleq \frac{1}{Mn_{MDP}}\sum_{i=1}^{Mn_{MDP}}{\mathcal{L} _{MDP}\left( g\left( \hat{x}_{MDP,i} \right) \right)}\\
	\end{cases}\begin{cases}
		R_U(g)\triangleq \underset{x\sim p_U\left( x \right)}{\mathbb{E}}\left[ \mathcal{L} _U\left( g\left( x \right) \right) \right]\\
		\hat{R}_U(g)\triangleq \frac{1}{n_U}\sum_{i=1}^{n_U}{\mathcal{L} _U\left( g\left( x_{U,i} \right) \right)}\\
	\end{cases}
	$$
	
	Meanwhile,
	$$
	\hat{R}_{\widehat{MDP}U}(g)=\pi _+\pi _-\hat{R}_{\widehat{MDP}}(g)+\hat{R}_U(g)
	$$
	
	Specially,
	$$
	\hat{R}_{MDPU}(g)=\hat{R}_{\widehat{MDP}U}(g)
	$$

	We can obtain that:
	\begin{equation}
		\begin{aligned}   
			R(\hat{g})-R(g^*)&=R_{MDPU}(\hat{g})-R_{MDPU}(g^*)
			\\&
			=\left( R_{MDPU}(\hat{g})-\hat{R}_{MDPU}(\hat{g}) \right) +\left( \hat{R}_{MDPU}(\hat{g})-\hat{R}_{MDPU}(g^*) \right) +\left( \hat{R}_{MDPU}(g^*)-R_{MDPU}(g^*) \right) 
			\\&
			\leq \left( R_{MDPU}(\hat{g})-\hat{R}_{MDPU}(\hat{g}) \right) +0+\left( \hat{R}_{MDPU}(g^*)-R_{MDPU}(g^*) \right) 
			\\&
			\leq 2\underset{g\in \mathcal{G}}{sup}\left| R_{MDPU}(g)-\hat{R}_{MDPU}(g) \right|
			\\&
			=2\underset{g\in \mathcal{G}}{sup}\left| R_{\widehat{MDP}U}(g)-\hat{R}_{\widehat{MDP}U}(g) \right|
			\\&
			\leq 2\pi _+\pi _-\underset{g\in \mathcal{G}}{sup}\left| R_{\widehat{MDP}}(g)-\hat{R}_{\widehat{MDP}}(g) \right|+2\underset{g\in \mathcal{G}}{sup}\left| R_U(g)-\hat{R}_U(g) \right|
		\end{aligned} 
	\end{equation}
	
	The generalization error bound is established through Talagrand's lemma, which bounds the complexity of the composite function class $\mathcal{G} \circ \ell$ (where $\ell$ is Lipschitz-continuous) via the empirical Rademacher complexity of the hypothesis set.
	\begin{lemma}(Uniform deviation bound):\label{lemma:8}
		Let $Z$ be a random variable drawn from a probability distribution with density $\mu$, $\mathcal{H} =\left\{ h:\mathcal{Z} \mapsto \left[ 0,M \right] \right\} \left( M>0 \right)$ be a class of measurable function, $\left\{ z_i \right\} _{i=1}^{n}$ be i.i.d. examples drawn from the distribution with density $\mu$. Then, for any $\delta>0$, the inequality below hold with probability at least $1-\delta $:
		\begin{equation}
			\begin{aligned} 
				\underset{g\in \mathcal{G}}{sup}\left| \underset{Z\sim \mu}{\mathbb{E}}-\frac{1}{n}\sum_{i=1}^n{g\left( z_i \right)} \right|\le 2\Re \left( \mathcal{G} \right) +M\sqrt{\frac{\log \frac{2}{\delta}}{2n}}
			\end{aligned} 
		\end{equation}
	\end{lemma}
	
	As shown by Lemma \ref{lemma:8}, it can be derived that:
	\begin{equation}
		\begin{aligned}   
			\underset{g\in \mathcal{G}}{sup}\left| R_{\widehat{MDP}}(g)-\hat{R}_{\widehat{MDP}}(g) \right|=&\underset{g\in \mathcal{G}}{sup}\left| \underset{x\sim \hat{p}_{MDP}\left( x \right) }{\mathbb{E}}\left[ \mathcal{L} _{MDP}\left( g\left( x \right) \right) \right] -\frac{1}{Mn_{MDP}}\sum_{i=1}^{Mn_{MDP}}{\left[ \mathcal{L} _{MDP}\left( g\left( x^i \right) \right) \right]} \right|
			\\
			=&\frac{1}{\left| a\pi _--b\pi _+ \right|}\underset{g\in \mathcal{G}}{sup}\left| \underset{x\sim \hat{p}_{MDP}\left( x \right) }{\mathbb{E}}\left[ \tilde{\ell}\left( g\left( x \right) \right) \right] -\frac{1}{Mn_{MDP}}\sum_{i=1}^{Mn_{MDP}}{\left[ \tilde{\ell}\left( g\left( x_i \right) \right) \right]} \right|
			\\
			\leq& \frac{1}{\left| a\pi _--b\pi _+ \right|}\left\{ \underset{g\in \mathcal{G}}{sup}\left| \underset{x\sim \hat{p}_{MDP}\left( x \right) }{\mathbb{E}}\left[ \ell \left( g\left( x \right) ,+1 \right) \right] -\frac{1}{Mn_{MDP}}\sum_{i=1}^{Mn_{MDP}}{\left[ \ell \left( g\left( x_i \right) ,+1 \right) \right]} \right|  \right. \\& \left.    +\underset{g\in \mathcal{G}}{sup}\left| \underset{x\sim \hat{p}_{MDP}\left( x \right) }{\mathbb{E}}\left[ \ell \left( g\left( x \right) ,-1 \right) \right] -\frac{1}{Mn_{MDP}}\sum_{i=1}^{Mn_{MDP}}{\left[ \ell \left( g\left( x_i \right) ,-1 \right) \right]} \right| \right\} 
			\\
			\leq& \frac{4\Re \left( \ell \circ \mathcal{G} ;Mn_{MDP},p_{MDP} \right)}{\left| a\pi _--b\pi _+ \right|}+\frac{2C_{\ell}}{\left| a\pi _--b\pi _+ \right|}\sqrt{\frac{\log \frac{4}{\delta}}{2Mn_{MDP}}}
		\end{aligned} 
	\end{equation}  
	
	The following inequality is established via Talagrand's lemma:
	\begin{equation}\label{Talagrand}
		\begin{aligned} 
			\Re \left( \ell \circ \mathcal{G} ;Mn_{MDP},p_{MDP} \right) \leq \rho \Re \left( \mathcal{G} ;Mn_{MDP},p_{MDP} \right) 
		\end{aligned} 
	\end{equation}
	where $\ell \circ \mathcal{G} $ denotes that $\left\{ \ell \circ \mathcal{G} \mid g\in \mathcal{G} \right\} $.
	
	\begin{lemma}\label{lemma-9}
		Let $\mathcal{C} _{\mathcal{G}}$ represents a non-negative constant. It is assumed that the following inequality holds for the set of models $\mathcal{G}$ and any given probability density $\mu$ defined over the data space:
		\begin{equation}
			\Re \left( \mathcal{G}\right) \le \frac{\mathcal{C} _{\mathcal{G}}}{\sqrt{n}}
		\end{equation}
	\end{lemma}
	An estimation error bound for the unbiased risk estimator derived from \emph{MDPU} data can be established by leveraging Rademacher complexity, McDiarmid's inequality, and Talagrand's lemma. According to Eq.(\ref{Talagrand}) and lemma \ref{lemma-9}, the following inequality holds with a probability at least $1-\frac{\delta}{2} $:
	\begin{equation}
		\begin{aligned}
			\underset{g\in \mathcal{G}}{sup}\left| R_{\widehat{MDP}}(g)-\hat{R}_{\widehat{MDP}}(g) \right|&\leq \frac{4\rho C_{\mathcal{G}}}{\left| a\pi _--b\pi _+ \right|\sqrt{Mn_{MDP}}}+\frac{2C_{\ell}}{\left| a\pi _--b\pi _+ \right|}\sqrt{\frac{\log \frac{4}{\delta}}{2Mn_{MDP}}}
			\\&
			=\frac{4\sqrt{2}\rho C_{\mathcal{G}}}{\left| a\pi _--b\pi _+ \right|\sqrt{2Mn_{MDP}}}+\frac{2C_{\ell}}{\left| a\pi _--b\pi _+ \right|}\sqrt{\frac{\log \frac{4}{\delta}}{2Mn_{MDP}}}
			\\&
			=\frac{4\sqrt{2}\rho C_{\mathcal{G}}+2C_{\ell}\sqrt{\log \frac{4}{\delta}}}{\left| a\pi _--b\pi _+ \right|\sqrt{2Mn_{MDP}}}
		\end{aligned}
	\end{equation} 
	
	Following the same way, 
	
	\begin{equation}
		\begin{aligned}
			\underset{g\in \mathcal{G}}{sup}\left| R_U(g)-\hat{R}_U(g) \right|=&\underset{g\in \mathcal{G}}{sup}\left| \underset{x\sim p_U\left( x \right)}{\mathbb{E}}\left[ \mathcal{L} _U\left( g\left( x \right) \right) \right] -\frac{1}{n_U}\sum_{i=1}^{n_U}{\left[ \mathcal{L} _U\left( g\left( x \right) \right) \right]} \right|
			\\
			=&\left| \frac{-b\pi _+}{a\pi _--b\pi _+} \right|\underset{g\in \mathcal{G}}{sup}\left| \underset{x\sim p_U\left( x \right)}{\mathbb{E}}\left[ \ell \left( g\left( x \right) ,+1 \right) \right] -\frac{1}{n_U}\sum_{i=1}^{n_U}{\left[ \ell \left( g\left( x_i \right) ,+1 \right) \right]} \right|
			\\&
			+\left| \frac{a\pi _-}{a\pi _--b\pi _+} \right|\underset{g\in \mathcal{G}}{sup}\left| \underset{x\sim p_U\left( x \right)}{\mathbb{E}}\left[ \ell \left( g\left( x \right) ,-1 \right) \right] -\frac{1}{n_U}\sum_{i=1}^{n_U}{\left[ \ell \left( g\left( x_i \right) ,-1 \right) \right]} \right|
			\\
			\leq& \left| \frac{-b\pi _+}{a\pi _--b\pi _+} \right|\left\{ 2\Re \left( \ell \circ \mathcal{G} ;n_U,p_U \right) +C_{\ell}\sqrt{\frac{\log \frac{4}{\delta}}{2n_U}} \right\} \\&+\left| \frac{a\pi _-}{a\pi _--b\pi _+} \right|\left\{ 2\Re \left( \ell \circ \mathcal{G} ;n_U,p_U \right) +C_{\ell}\sqrt{\frac{\log \frac{4}{\delta}}{2n_U}} \right\} 
			\\
			=&\frac{\left| -b\pi _+ \right|+\left| a\pi _- \right|}{\left| a\pi _--b\pi _+ \right|}2\Re \left( \ell \circ \mathcal{G} ;n_U,p_U \right) +\frac{\left| -b\pi _+ \right|+\left| a\pi _- \right|}{\left| a\pi _--b\pi _+ \right|}C_{\ell}\sqrt{\frac{\log \frac{4}{\delta}}{2n_U}}
			\\
			\leq& \frac{\left( \left| -b\pi _+ \right|+\left| a\pi _- \right| \right) 2\rho C_{\mathcal{G}}}{\left| a\pi _--b\pi _+ \right|\sqrt{n_U}}+\frac{\left| -b\pi _+ \right|+\left| a\pi _- \right|}{\left| a\pi _--b\pi _+ \right|}C_{\ell}\sqrt{\frac{\log \frac{4}{\delta}}{2n_U}}
			\\
			=&\frac{\left( \left| -b\pi _+ \right|+\left| a\pi _- \right| \right) \sqrt{8}\rho C_{\mathcal{G}}+\left( \left| -b\pi _+ \right|+\left| a\pi _- \right| \right) C_{\ell}\sqrt{\log \frac{4}{\delta}}}{\left| a\pi _--b\pi _+ \right|\sqrt{2n_U}}
		\end{aligned}
	\end{equation}

	Therefore, we can conclude that,
	\begin{equation}
		\begin{aligned}   
			R(\hat{g})-R(g^*)\leq 2\pi _+\pi _-\frac{4\sqrt{2}\rho C_{\mathcal{G}}+2C_{\ell}\sqrt{\log \frac{4}{\delta}}}{\left| a\pi _--b\pi _+ \right|\sqrt{2Mn_{MDP}}}+\frac{2\left( \left| -b\pi _+ \right|+\left| a\pi _- \right| \right) \sqrt{8}\rho C_{\mathcal{G}}+2\left( \left| -b\pi _+ \right|+\left| a\pi _- \right| \right) C_{\ell}\sqrt{\log \frac{4}{\delta}}}{\left| a\pi _--b\pi _+ \right|\sqrt{2n_U}}
		\end{aligned} 
	\end{equation}
	
	\subsection{Proof of Theorem 3}
	\setcounter{equation}{0}
	\renewcommand\theequation{J.\arabic{equation}}
	\noindent \emph{Proof.}
	Assume that there exits $\zeta >0$ and $\eta >0$ such that $R_{MDP}\left( g \right) \geq \zeta $ and $R_U\left( g \right) \geq \eta $.
	\begin{equation}
		\begin{aligned}  
			P\left( \mathfrak{D} _{MDP},\mathfrak{D} _U \right) =P_{MDP}\left( x_{1}^{1},\cdots ,x_{1}^{M} \right) \cdots P_{MDP}\left( x_{n_{MDP}}^{1},\cdots ,x_{n_{MDP}}^{M} \right) P\left( x_{1}^{U} \right) \cdots P\left( x_{n_U}^{U} \right) 
		\end{aligned} 
	\end{equation}
	
	\begin{equation}
		\begin{aligned} 
			F\left( \mathfrak{D} _{MDP},\mathfrak{D} _U \right) =F\left( \mathfrak{D} _{MDP} \right) F\left( \mathfrak{D} _U \right) 
		\end{aligned} 
	\end{equation}

	Then we have:
	\begin{equation}
		\begin{aligned}  
			Pr\left( \mathfrak{D} ^-\left( g \right) \right) &=\,\,Pr\left( \hat{R}_{MDP}<0 \right) +\,\,Pr\left( \hat{R}_U<0 \right) 
			\\&
			\leq \,\,Pr\left( \hat{R}_{MDP}\leq R_{MDP}-\zeta \right) +\,\,Pr\left( \hat{R}_U\leq R_U-\eta \right) 
			\\&
			=Pr\left( \hat{R}_{MDP}-R_{MDP}\geq \zeta \right) +\,\,Pr\left( \hat{R}_U-R_U\geq \eta \right) 
			\\&
			\leq \exp \left( -\frac{2\zeta ^2M^2\left( a\pi _--b\pi _+ \right) ^2n_{MDP}^{2}}{n_{MDP}\left( \pi _+\pi _- \right) ^2C_{\ell}^{2}} \right) +\exp \left( -\frac{2\eta ^2n_{U}^{2}}{n_UC_{\ell}^{2}} \right) 
			\\&
			=\exp \left( -\frac{2\zeta ^2M^2\left( a\pi _--b\pi _+ \right) ^2n_{MDP}}{\left( \pi _+\pi _- \right) ^2C_{\ell}^{2}} \right) +\exp \left( -\frac{2\eta ^2n_U}{C_{\ell}^{2}} \right) 
		\end{aligned} 
	\end{equation}
	
	And,
	\begin{equation}
		\begin{aligned}  
			\mathbb{E} [\tilde{R}(g)]-R(g)=&\mathbb{E} [\tilde{R}(g)-\hat{R}(g)]
			\\
			=&\int_{\left( \mathfrak{D} _{MDP},\mathfrak{D} _U \right) \in \mathcal{D} ^+(g)}{(\tilde{R}(g)-\hat{R}(g))}dF\left( \mathfrak{D} _{MDP},\mathfrak{D} _U \right) \\&+\int_{\left( \mathfrak{D} _{MDP},\mathfrak{D} _U \right) \in \mathcal{D} ^-(g)}{(\tilde{R}(g)-\hat{R}(g))}dF\left( \mathfrak{D} _{MDP},\mathfrak{D} _U \right) 
			\\
			=&\int_{\left( \mathfrak{D} _{MDP},\mathfrak{D} _U \right) \in \mathcal{D} ^-(g)}{(\tilde{R}(g)-\hat{R}(g))}dF\left( \mathfrak{D} _{MDP},\mathfrak{D} _U \right) 
		\end{aligned} 
	\end{equation}

	\begin{equation}\label{eq-J5}
		\begin{aligned}  
			\mathbb{E} [\tilde{R}_{MDPU}(g)]-R(g)&=\int_{\left( \mathfrak{D} _{MDP},\mathfrak{D} _U \right) \in \mathcal{D} ^-(g)}{(\tilde{R}(g)-\hat{R}(g))}dF\left( \mathfrak{D} _{MDP},\mathfrak{D} _U \right) 
			\\&
			\leq \underset{\left( \mathfrak{D} _{MDP},\mathfrak{D} _U \right) \in \mathcal{D} ^-(g)}{sup}\left( \tilde{R}(g)-\hat{R}(g) \right) \int_{\left( \mathfrak{D} _{MDP},\mathfrak{D} _U \right) \in \mathcal{D} ^-(g)}{dF\left( \mathfrak{D} _{MDP},\mathfrak{D} _U \right)}
			\\&
			=\underset{\left( \mathfrak{D} _{MDP},\mathfrak{D} _U \right) \in \mathcal{D} ^-(g)}{sup}\left( f\left( \hat{R}_{MDP}(g) \right) +f\left( \hat{R}_U(g) \right) -\hat{R}_{MDP}(g)-\hat{R}_U(g) \right) \,\,Pr\left( \mathfrak{D} ^-\left( g \right) \right) 
			\\&
			\leq \underset{\left( \mathfrak{D} _{MDP},\mathfrak{D} _U \right) \in \mathcal{D} ^-(g)}{sup}\left( L_f\left| \hat{R}_{MDP}(g) \right|+L_f\left| \hat{R}_U(g) \right|+\left| \hat{R}_{MDP}(g) \right|+\left| \hat{R}_U(g) \right| \right) \,\,Pr\left( \mathfrak{D} ^-\left( g \right) \right) 
			\\&
			=\underset{\left( \mathfrak{D} _{MDP},\mathfrak{D} _U \right) \in \mathcal{D} ^-(g)}{sup}\left\{ \left( L_f+1 \right) \left| \hat{R}_{MDP}(g) \right|+\left( L_f+1 \right) \left| \hat{R}_U(g) \right| \right\} \,\,Pr\left( \mathfrak{D} ^-\left( g \right) \right) 
			\\&
			\leq \left[ \frac{\left( L_f+1 \right) \left( \pi _+\pi _- \right) MC_{\ell}}{\left( a\pi _--b\pi _+ \right)}+\left( L_f+1 \right) C_{\ell} \right] Pr\left( \mathfrak{D} ^-\left( g \right) \right) 
		\end{aligned} 
	\end{equation}
	
	The following inequality is directly obtained from Eq.(\ref{eq-J5}):
	\begin{equation}
		\begin{aligned}  
			\left| \tilde{R}_{MDPU}(g)-R(g) \right|&\leq \left| \tilde{R}_{MDPU}(g)-\mathbb{E} \left[ \tilde{R}_{MDPU}(g) \right] \right|+\left| \mathbb{E} \left[ \tilde{R}_{MDPU}(g) \right] -R(g) \right|
			\\&
			=\left| \tilde{R}_{MDPU}(g)-\mathbb{E} \left[ \tilde{R}_{MDPU}(g) \right] \right|+\left[ \frac{\left( L_f+1 \right) \left( \pi _+\pi _- \right) MC_{\ell}}{\left( a\pi _--b\pi _+ \right)}+\left( L_f+1 \right) C_{\ell} \right] Pr\left( \mathfrak{D} ^-\left( g \right) \right) 
		\end{aligned} 
	\end{equation}
	
	According to McDiarmid's inequality, there exits a constant $\varepsilon$ ($\varepsilon > 0$) for which the following inequality is valid:
	\begin{equation}
		\begin{aligned}  
			Pr\left\{ \left| \tilde{R}_{MDPU}(g)-\mathbb{E} \left[ \tilde{R}_{MDPU}(g) \right] \right|\geq \varepsilon \right\} \leq 2\exp \left( -\frac{2\varepsilon ^2}{n_{MDP}\left( \frac{\left( \pi _+\pi _- \right) C_{\ell}L_f}{M\left( a\pi _--b\pi _+ \right) n_{MDP}} \right) ^2+n_U\left( \frac{C_{\ell}L_f}{n_U} \right) ^2} \right) 
		\end{aligned} 
	\end{equation}
	
	then we have the following inequality with probability at least $1-\delta$:
	\begin{equation}
		\begin{aligned} 
			\left| \tilde{R}_{MDPU}(g)-\mathbb{E} \left[ \tilde{R}_{MDPU}(g) \right] \right|\leq C_{\ell}L_f\sqrt{\frac{1}{2}\ln \frac{2}{\delta}\left( \frac{\left( \pi _+\pi _- \right) ^2}{M^2\left( a\pi _--b\pi _+ \right) ^2n_{MDP}}+\frac{1}{n_U} \right)}
		\end{aligned} 
	\end{equation}

	Therefore, we have
	\begin{equation}
		\begin{aligned} 
			&\left| \tilde{R}_{MDPU}(g)-R(g) \right|
			\\
			\le& C_{\ell}L_f\sqrt{\frac{1}{2}\ln \frac{2}{\delta}\left( \frac{\left( \pi _+\pi _- \right) ^2}{M^2\left( a\pi _--b\pi _+ \right) ^2n_{MDP}}+\frac{1}{n_U} \right)}
			\\&
			+\left[ \frac{\left( L_f+1 \right) \left( \pi _+\pi _- \right) MC_{\ell}}{\left( a\pi _--b\pi _+ \right)}+\left( L_f+1 \right) C_{\ell} \right] Pr\left( \mathfrak{D} ^-\left( g \right) \right) 
			\\
			\le& C_{\ell}L_f\sqrt{\frac{1}{2}\ln \frac{2}{\delta}\left( \frac{\left( \pi _+\pi _- \right) ^2}{M^2\left( a\pi _--b\pi _+ \right) ^2n_{MDP}}+\frac{1}{n_U} \right)}
			\\&
			+\left[ \frac{\left( L_f+1 \right) \left( \pi _+\pi _- \right) MC_{\ell}}{\left( a\pi _--b\pi _+ \right)}+\left( L_f+1 \right) C_{\ell} \right] \left[ \exp \left( -\frac{2\zeta ^2M^2\left( a\pi _--b\pi _+ \right) ^2n_{MDP}}{\left( \pi _+\pi _- \right) ^2C_{\ell}^{2}} \right) +\exp \left( -\frac{2\eta ^2n_U}{C_{\ell}^{2}} \right) \right]  
		\end{aligned} 
	\end{equation}

\end{document}